\documentclass[a4paper,10pt]{article}
\usepackage[margin=1in]{geometry}
\usepackage[utf8]{inputenc} 
\usepackage[T1]{fontenc}
\usepackage{graphicx}
\usepackage{amsmath, amssymb}
\usepackage{algorithm}
\usepackage{algpseudocode}
\usepackage{float}
\usepackage{hyperref}
\usepackage{authblk} 
\usepackage{setspace}
\usepackage{caption}
\usepackage{xcolor}
\usepackage{xspace}
\usepackage{bm}
\usepackage{mathtools}   
\usepackage{booktabs}
\usepackage{tabularx}    
\usepackage{array}
\usepackage{algorithmicx}
\usepackage{amsthm}
\usepackage{etoc}        

\theoremstyle{definition}

\newcommand{\model}{\textsc{Newmark-$\beta$-DGN}\xspace}

\newcommand{\dgn}{\textsc{DGN}\xspace}

\newcommand{\eghn}{EGHN\xspace}
\newcommand{\egno}{EGNO\xspace}
\newcommand{\eghno}{EGHNO\xspace}
\newcommand{\gns}{GNS\xspace}
\newcommand{\mgn}{MGN\xspace}

\newcommand{\dt}{\Delta t}
\newcommand{\subdt}{\delta t}

\newcommand{\Kt}{\mathbf{K}}
\newcommand{\Dt}{\mathbf{D}}
\newcommand{\rev}[1]{\textcolor{black}{#1}}

\title{\textbf{Learning coarse-step dynamics and internal mechanical response with graph networks}}

\author[1]{Vinay Sharma}
\author[1,*]{Olga Fink}

\affil[1]{Intelligent Maintenance and Operations Systems, EPFL, Lausanne, Switzerland}
\affil[*]{Corresponding Author}

\date{}

\begin{document}
\etocdepthtag.toc{main}

\maketitle

\section*{Abstract}

Modern sensing records the motion of physical systems, but often leaves the forces and mechanical response governing that motion unobserved. Inferring these quantities from discretely sampled trajectories is especially difficult at coarse time scales, when mechanical response evolves between observations and interactions propagate across the system. Here we introduce \model, a graph neural network-based framework that combines two structures inspired by computational mechanics. First, a semi-implicit update inspired by the Newmark-$\beta$ method uses learned momentum fluxes and matrix-valued response operators to advance the state over each observed interval. Second, an operator-weighted virtual hub provides system-wide coupling through a sparse set of connections. The learned quantities thus determine the predicted motion and remain accessible for mechanical analysis. Across a deformable beam, human motion and protein dynamics, \model supports long-horizon prediction at time steps for which explicit learned simulators deteriorate. Without force, moment or constitutive relation supervision, forces inferred from walking kinematics track independently derived hip and knee joint moments, while response operators learned on the beam recover the relative spatial and directional structure of its finite-element stiffness tangent. \model therefore links coarse-step prediction to the inference of mechanical quantities that were never observed during training.

\section{Introduction}
\label{sec:intro}


Motion provides a direct window into the evolution of a physical system, but not necessarily into the mechanics that produce it. Structural behavior is governed by internal stresses and load transfer; biomechanical function by forces and joint moments; and interacting systems by forces transmitted across contacts and interfaces. Yet these quantities are often difficult or impossible to measure continuously, and different internal mechanical states can produce similar observable motion. Accurate trajectory prediction alone, therefore, does not establish that a model has recovered the mechanics underlying the observed dynamics. Bridging this gap requires models that infer unobserved mechanical quantities from kinematic observations while using those same quantities to generate the predicted evolution. In such a model, forces, stresses, and mechanical operators are not auxiliary explanations attached to a prediction, but explicit components of the learned dynamics itself.

Inferring these mechanical quantities becomes increasingly challenging as the observations become more widely spaced in time. At coarse temporal resolution, the observations no longer resolve wave propagation, contact, or vibration over their native timescales. They instead reveal only the net state change produced by these processes over the observation interval.
Over the same interval, a disturbance can travel through a larger part of the physical system. Predicting a single coarse transition, therefore, requires both a stable representation of the response accumulated over the interval and a mechanism for propagating that response across the corresponding spatial range.

Graph-based models offer a natural starting point for addressing these requirements in discretized mechanical systems. Material points, particles, or components become nodes, while their interactions become edges, allowing information to be propagated across the system through message passing. Graph Network-based Simulators (GNS)~\cite{gns} and MeshGraphNets (MGN)~\cite{mgn} use this organization to predict trajectories across changing geometries and connectivities. \(E(n)\)-Equivariant Graph Neural Networks (EGNNs) make these predictions consistent under changes of reference frame~\cite{REF_EGNN}, Equivariant Graph Hierarchy-based Neural Networks (EGHNs) shorten communication paths between distant parts of a system~\cite{eghn}, and the Equivariant Graph Neural Operator (EGNO) represents evolution over wider temporal intervals~\cite{egno}. These developments extend the spatial and temporal reach of graph-based prediction, but the quantities exchanged within these networks generally lack an explicit mechanical interpretation. They are optimized through training to predict the future state without requiring the learned interactions to correspond to the forces, moments, or stiffness and damping like response operators responsible for that evolution. 

A complementary line of work introduces mechanical meaning through either direct supervision or explicit physical structure specified a priori. Neural Equivariant Interatomic Potentials (NequIP)~\cite{REF_NEQUIP} and the Message-Passing Atomic Cluster Expansion (MACE)~\cite{REF_MACE} are directly supervised using interatomic energies and forces from quantum-mechanical calculations. 
This provides a direct mechanical interpretation of the learned internal quantities, but requires those quantities to be available as training labels.

When direct interaction labels are unavailable, different types of interactions can instead be inferred from observed trajectories. 
Neural Relational Inference (NRI)~\cite{REF_NRI} represents interactions as latent categorical edge types, while Collective Relational Inference (CRI)~\cite{REF_CRI} extends this construction to heterogeneous interactions. Both infer these latent relations through trajectory prediction rather than direct supervision of the interactions themselves. These relations characterize how components interact but do not themselves correspond to mechanically defined quantities such as forces or moments. A stronger mechanical structure is introduced by the Physics-Induced Graph Network for Particle Interaction (PIG'N'PI)~\cite{REF_PIGNPI} by representing the trajectory-inferred interactions as pairwise Newtonian forces. In these approaches, however, the inferred interactions either remain latent relational quantities or acquire mechanical meaning through an interaction form specified in advance.

The mechanical structure can also be provided through the physical equations governing trajectory evolution. Port-Hamiltonian neural networks represent conservative and dissipative effects through distinct  Hamiltonian and dissipation terms, with the port-Hamiltonian equations specifying how these terms contribute to the state evolution~\cite{REF_PORT_HNN}. The Information-preserving Graph Neural Simulator (IGNS) extends this port-Hamiltonian formulation to graph-based dynamics, using GNNs to parameterize the Hamiltonian and dissipative terms while retaining their roles within the port-Hamiltonian state evolution~\cite{REF_IGNS}. Mechanical information can also be supplied directly from physics-based simulations or models. Equi-Euler GraphNet~\cite{sharma2025equieuler} learns internal interaction forces using force labels generated by high-fidelity multiphysics simulations, whereas the Physics-encoded Time Integrator Graph Network (PeTIGN)~\cite{REF_PETIGN} uses mass, damping, and stiffness operators obtained from a finite-element model of the system as known inputs to a Newmark-based nodal update. These approaches therefore require mechanical quantities to be available independently of the observed trajectories, either as supervision or as prescribed model inputs.

Across these approaches, learned internal quantities acquire mechanical meaning from physical information or model structure beyond the observed trajectories themselves. Direct mechanical labels specify what these quantities represent, whereas assumed interaction forms or governing equations define the mechanical form they take and the role they play in the dynamics. Each of these requirements can be restrictive in real systems. Internal forces, moments, and load-transfer quantities are often difficult to measure directly because they arise at contacts, interfaces, or along structural load paths, while reconstructing them with physics-based models requires constitutive relations, mechanical parameters, boundary conditions, and system-specific calibration. Predefining the mechanical formulation not only constrains the learned dynamics to an assumed interaction or response structure, but also requires uncertain and difficult-to-estimate system-specific quantities, such as mass, damping, and stiffness,  may themselves be uncertain or unavailable. These limitations motivate learning mechanically interpretable internal quantities from observed kinematics without direct supervision or assuming any predefined form for those quantities, using general physical principles to define their mechanical role while using the same inferred quantities to generate the predicted state evolution.


Momentum conservation provides such a general physical principle. Dynami-CAL GraphNet (DGN) 
uses this universal law to assign a mechanical interpretation to interactions inferred from trajectories~\cite{dgn}.
Each edge represents a pairwise mechanical interaction between two components. DGN represents this interaction as a momentum exchange, aggregates the resulting momentum fluxes at each node, and uses them directly to advance the state. At coarse temporal resolution, however, the interaction represented over one observed state-transition is an effective exchange accumulated over unresolved bending, torsion, contact, or articulated motion, and its resultant force need not act along the line joining the interacting components. Such a non-central interaction cannot be described by linear-momentum exchange alone. Non-central forces generate a moment and thereby contribute to angular-momentum exchange, even when equal and opposite forces balance linear momentum. DGN therefore adopts a Cosserat-type representation in which each node carries a rotational degree of freedom and each edge carries an antisymmetric angular-momentum flux. This allows the orbital moment generated by the force and the intrinsic rotational contribution to be accounted for jointly in the angular-momentum balance. When rotation is not observed, the rotational state provides a latent representation of the angular-momentum exchange underlying the observed translational dynamics.

\begin{figure}[!tp]
  \centering
  \includegraphics[width=1.0\textwidth]
  {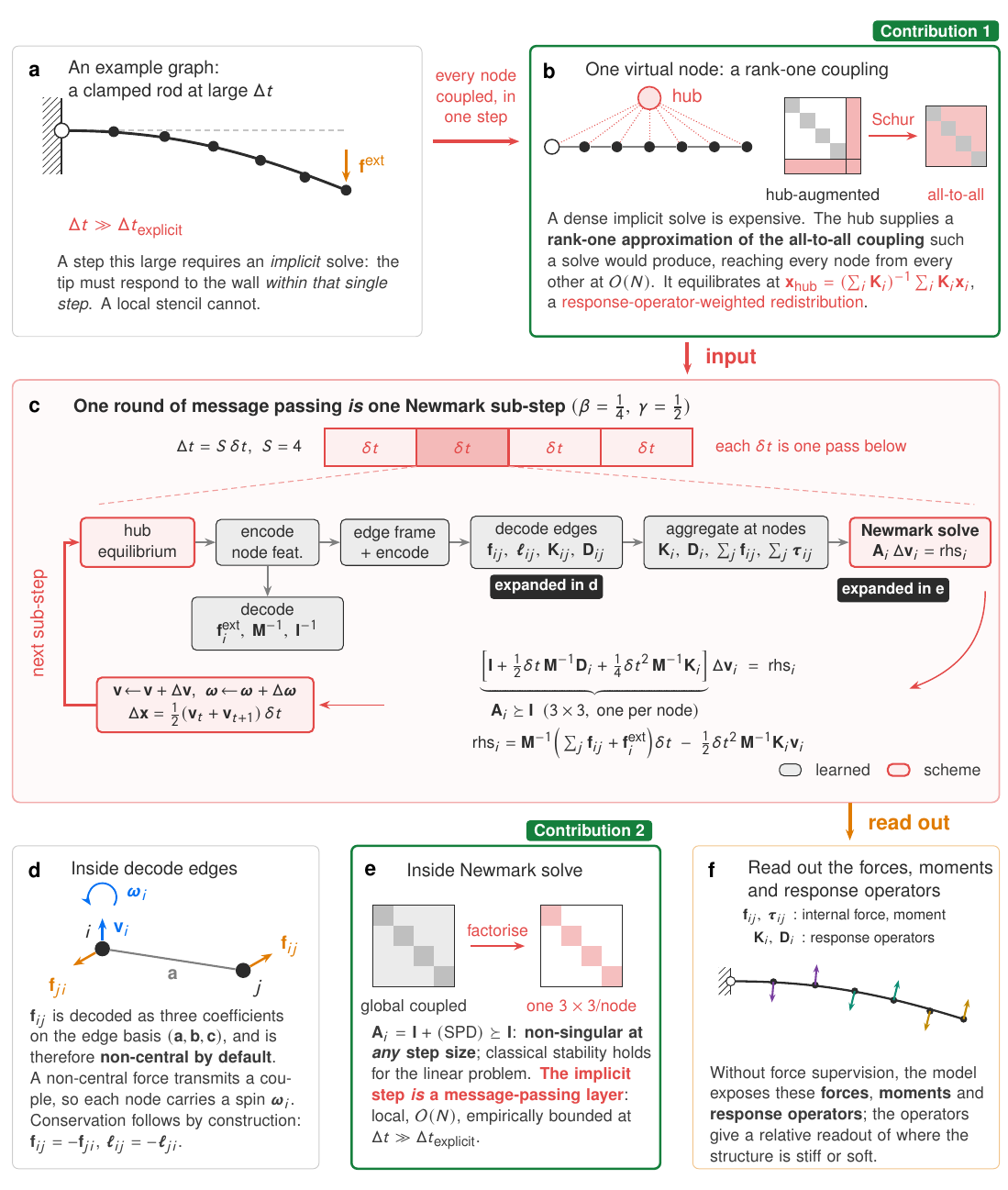}
  \caption{\textbf{Observation-learned mechanics within a coarse graph update.}
  \textbf{a}, Over a coarse observation interval, the loaded tip of a clamped rod
  must respond to the distant constraint within the same transition.
  \textbf{b}, Condensation of an implicit mechanical system produces dense
  coupling. Its rank-one approximation is represented by an operator-weighted
  virtual hub, giving a graph of diameter two with $\mathcal{O}(N)$ edges.
  \textbf{c}, Each message-passing round is one semi-implicit Newmark substep:
  edge quantities are decoded and aggregated, the state is advanced and the
  resulting state enters the next round.
  \textbf{d}, Physical edges exchange non-central linear- and angular-momentum
  fluxes antisymmetrically, whereas virtual hub edges satisfy collective zero-sum
  balance.
  \textbf{e}, Positive-definite learned response operators enter the local
  translational and rotational Newmark systems.
  \textbf{f}, The quantities generating the transition remain accessible as
  internal-force, torque and \rev{response-operator} readouts.}
  \label{fig:framework}
\end{figure}

DGN's interaction representation still leaves three limitations in the finite-time update, all of which become more pronounced as the observation interval increases. First, the interaction fluxes are evaluated from the current state and then used to advance it explicitly. Over a larger interval, the state can change substantially, and therefore the mechanical response can evolve during the transition rather than remain fixed at its initial value. Second, explicit integration of stiff interactions or increasingly fine spatial discretizations requires progressively smaller time steps, thereby increasing the number of integration steps needed to span a single observation interval. Third, mechanical influence is communicated only along edges connecting directly interacting components. As the observation interval grows, a disturbance can propagate across a larger part of the system within a single observed state-transition, requiring information to traverse an increasing number of physical interaction edges.
This third limitation is illustrated by the clamped rod  in Fig.~\ref{fig:framework}a. A disturbance introduced at the loaded tip propagates along the rod through successive local interactions. Once the observation interval is sufficiently long for the disturbance to reach the distant clamp, the state reached within that single transition must reflect the clamp’s influence on the system-wide mechanical response. Representing this finite-time coupling through local interactions alone would require information to traverse the full chain of physical edges within one observed transition. The update therefore must capture the system-wide mechanical coupling within single state-transition.

In this paper, we introduce \model to address these limitations through two coupled components: a semi-implicit nodal update for stable finite-time integration and an operator-weighted virtual hub that  approximates system-wide mechanical coupling within a single observed state transition. 
Both components are motivated by the structure of the conventional implicit average-acceleration Newmark scheme~\cite{newmark1959}.

In an explicit update,  mechanical forces are evaluated from the state at the beginning of the step and are then used to advance the system. An implicit Newmark scheme instead evaluates the mechanical response at the updated state, such that the forces depend on displacement and velocity increments that are themselves unknown~\cite{newmark1959}. In a discretized mechanical system, this makes the update inherently coupled: the response of each component depends not only on its own state increments but also on those of mechanically connected components. The increments therefore cannot be determined independently. Their equations must be assembled into a global matrix system and solved jointly. 

This global formulation is what allows the implicit formulation to address both the temporal and spatial limitations of the explicit update. Evaluating the response at the updated state allows the mechanics to evolve over the step and permits stable integration at step sizes that would be too large for the corresponding explicit scheme. The coupled equations are assembled into a global system matrix whose sparsity reflects the local connectivity of the underlying mechanical interactions. Although this matrix is sparse, its inverse is generally dense. Consequently, a force or constraint acting on one component can affect distant components within the same update. The global solve thereby converts local mechanical interactions into a system-wide finite-time response.  The global solve, however, is computationally expensive. Replacing it with independent nodal solves preserves efficiency but removes precisely the nonlocal coupling induced by the global solve. \model therefore separates these two roles: it retains the local finite-time response through independent semi-implicit nodal updates and introduces a separate mechanism to approximate the system-wide coupling omitted by those local solves.

To recover the system-wide coupling without solving the full global system, we consider a rank-one approximation of the dense coupling implied by the implicit solve. Specifically, we approximate the dense all-to-all coupling induced by the implicit solve with a rank-one operator. We show that, under this approximation, the long-range pairwise coupling can be represented through a single shared state. Each component contributes to this shared state through its learned response operator and responds relative to it, resulting in operator-weighted equilibrium. This induces a star topology in which long-range interactions are mediated through the shared state, while local interactions remain represented by direct physical connections between components. The hub, therefore, approximates system-wide mechanical coupling with \(O(N)\) virtual edges, without constructing or inverting a global matrix. The resulting augmented graph has an effective diameter of two. The operator-weighted hub is defined in Section~\ref{ssec:m:hub} (Eq.~\eqref{eq:m:hub}); its derivation, which eliminates the internal degrees of freedom of the implicit system and approximates the resulting dense coupling by a rank-one form, is given in Supplementary Information, Sections~9.2.1--9.2.2.

We implement this rank-one-motivated star construction in \model by augmenting the physical graph with a single virtual hub. The physical nodes represent the system components, and bidirectional physical edges represent their direct interactions, as illustrated for the clamped rod in Fig.~\ref{fig:framework}a. The hub is connected to all \(N\) physical nodes through \(\mathcal{O}(N)\) bi-directional virtual edges, giving the augmented graph a  diameter of two (Fig.~\ref{fig:framework}b). Unlike hierarchical or long-range graph architectures introduced primarily to shorten communication paths, such as EGHN~\cite{eghn}, this virtual connectivity is motivated by the rank-one approximation of the dense system-wide coupling implied by the implicit formulation. The resulting topology thus has a mechanical interpretation beyond merely facilitating message propagation.

\model does not assume that the true underlying system-wide coupling is exactly rank one. Instead, it retains the star topology revealed by the rank-one derivation while replacing the scalar couplings  with learned matrix-valued response operators. The hub in the star topology is neither a physical body nor an independently integrated mechanical degree of freedom. Rather,   it is a shared state whose position, velocity, and angular velocity are recomputed at every substep as operator-weighted equilibria of the physical-node states. 
The hub, therefore, provides a tractable approximation to the system-wide coupling that would otherwise arise through the global implicit solve.

\model advances each observed state transition through \(S\) semi-implicit sub time-steps, where \(S\) is a chosen hyperparameter and each substep corresponds to one complete round of message passing (Fig.~\ref{fig:framework}c--e). This recurrent update allows the mechanical response and the corresponding system-wide coupling to evolve within the observation interval. At the beginning of each sub-step, the hub state is recomputed from the current physical-node states using the aggregated nodal response operators decoded during the preceding round. For the first substep, these operators are initialized as  identity matrices, so the hub position, velocity, and angular velocity reduce to the corresponding arithmetic averages over the physical nodes.

The resulting physical-node and hub states form the augmented graph state for the current substep. Following the DGN construction, an edge-local reference frame is first constructed for each physical and virtual edge. Vector quantities associated with the edge and its incident nodes are projected onto this frame, yielding invariant scalars that are passed to the edge encoders. From the resulting edge representations, the decoders produce linear and angular-momentum fluxes. On each bidirectional physical edge, the antisymmetric DGN construction ensures that these momentum fluxes are equal and opposite in the two directions (Fig.~\ref{fig:framework}d). For the virtual-edges, the linear and angular-momentum fluxes are projected onto the subspace of zero collective sum. Because the hub carries no mass, inertia, or external load, this projection ensures that it redistributes internal forces and torques without introducing a net external momentum contribution to the system. \model additionally decodes symmetric positive-definite matrix-valued response operators: \(K_{ij}\) and \(D_{ij}\) for the translational channel, and \(K^{\mathrm{rot}}_{ij}\) and \(D^{\mathrm{rot}}_{ij}\) for the rotational channel.

The decoded edge quantities are then aggregated at each physical node. The aggregated linear and angular-momentum fluxes provide the internal force and torque contributions, whereas the edge response operators are summed to obtain the nodal response operators. Together with the learned inverse mass and inertia from the node decoder, these quantities define the translational and rotational semi-implicit Newmark updates in Section~\ref{ssec:m:solve} Eqs.~\eqref{eq:m:solve} and~\ref{eq:m:solverot}, respectively. 

The external force is either provided as an input when observed or decoded from the current state when unobserved. In Eq.~\eqref{eq:m:solve}, the aggregated internal force is combined with the external force to form the net force drive on the right-hand side, while in Eq.~\eqref{eq:m:solverot}, the aggregated torque provides the corresponding rotational drive. The nodal response operators enter the \(3\times3\) coefficient matrices on the left-hand sides of the respective updates, allowing each node to account semi-implicitly for the local evolution of the mechanical response. Solving these small nodal systems avoids the global matrix solve of conventional implicit integration. The operators are then carried forward for the next time substep, where they determine the operator-weighted hub state. The nodal update and hub construction are therefore coupled recurrently: the hub mediates the system-wide response used in the current update, while the newly decoded response operators determine the hub state used in the next.


Each unclamped physical node then solves separate \(3\times3\) systems for its translational and rotational increments. Fig.~\ref{fig:framework}e illustrates this replacement for the translational update. These local \(3\times3\) systems require neither assembly nor factorization of a global \(3N\times3N\) matrix and can be solved in parallel. The learned response operators on each edge are constrained to be symmetric positive definite, ensuring that each nodal solve is well posed for every substep. Because the independent nodal solves approximate rather than reproduce the fully assembled system, exact conservation after integration is not assumed; the resulting finite-substep conservation residuals are derived and quantified in Supplementary Information, Section~11.

The proposed architecture is trained solely from observed kinematics, together with boundary and load inputs when available. When boundary conditions are unobserved, their effects are learned from the observed motion; when external forces are unobserved, they are decoded from the current graph state. The model receives no supervision from internal forces, moments, or mechanical operators such as stiffness and damping. The decoded linear and angular momentum fluxes provide internal force and torque contributions for the translational and rotational updates; the translational update also incorporates the observed or decoded external force. The learned response operators determine how the translational and rotational states respond within each time substep. Because these quantities participate directly in the state update, they remain accessible after training as internal forces, moments, and response operators (Fig.~\ref{fig:framework}f). 
We evaluate mechanical quantities derived from the decoded interactions against independent references that were never provided during training. The position-response operator indicates where the structure is locally stiff, while the velocity-response operator indicates where it is locally dissipative. Both are learned coefficients of the coarse-step update, interpretable as relative mechanical indicators rather than uniquely identified material or constitutive properties. 

We evaluate the proposed framework on four systems. Three are trajectory prediction benchmarks: a clamped finite-element beam, human motion capture and protein dynamics. The fourth is a benchmark for human walking biomechanics that tests the recovery of internal mechanics. These four settings test coarse-step stability, prediction on sparse irregular graphs, generalization across loads, geometries, and spatial resolutions, and recovery of internal forces and moments from observed kinematics. Across the three trajectory-prediction benchmarks, \model maintains bounded autoregressive rollouts at coarse temporal resolution. On the beam, it remains accurate under extrapolation in load, geometry, and mesh resolution. On human motion and protein dynamics, it maintains bounded autoregressive predictions using the sparse hub-augmented graph, while requiring fewer edges than the dense distance-based or multi-hop interaction graphs used by the compared methods.

A staged ablation examines the contributions of the hub and semi-implicit Newmark update. The mean whole-body error decreases from \(3.00\%\) for the twelve-substep hub-free explicit \dgn baseline to \(1.26\%\) for the four-substep hub-augmented explicit update, and further to \(0.70\%\) with the semi-implicit response and \(0.54\%\) with the full Newmark update. These results support complementary roles for the two components: the hub communicates mechanical response across the system within an observed transition, while the semi-implicit update supports integration of stiff dynamics at a coarse temporal resolution. 

Beyond predicting trajectories, we ask whether the quantities that participate in the learned update recover internal mechanics   never observed during training. 
In the human biomechanics experiment, joint-moment estimates derived from the inferred forces track the reference hip and knee moments, reaching correlations of \(r=0.94\) and \(r=0.87\), respectively. In the beam experiment, the learned response operators recover the spatial and directional organization of the finite-element  tangent stiffness, while the predicted dynamics reproduce the  fundamental frequency and dominant vibration mode of the structure. These results indicate that the learned interactions and response operators used for prediction also provide accessible representations of the  mechanical processes underlying the observed motion.


These results indicate that the learned interactions and response operators do more than support accurate trajectory prediction: they retain mechanical structure that can be evaluated against independent references, despite receiving no supervision on internal forces, moments, or response operators. By coupling an operator-weighted virtual hub to semi-implicit nodal integration, \model advances dynamics at coarse observation intervals without a global implicit solve. The forces and response operators are part of that update itself. This makes it possible to learn from observed motion both how a mechanical system evolves and which internal mechanical responses govern the learned evolution.

\section{Results}\label{sec:results}

\subsection{Overview of experiments}
\label{sec:overview}

We evaluate \model on four systems: a clamped finite-element beam, human motion capture, a solvated protein, and human walking biomechanics. The four settings test coarse-step stability, spatial generalization, prediction on sparse irregular graphs, and recovery of internal mechanical quantities from kinematics alone. For the biomechanics case, we introduce a new graph-based benchmark curated from the instrumented-treadmill recordings of van der Zee et al.~\cite{vanderzee2022}. Each walking trial is converted into a sequence of graphs in which the motion-capture markers form the nodes and the anatomical connections form the edges. The source recordings contain synchronized kinematics, ground-reaction forces, and inverse-dynamics joint moments. Only the graph kinematics are provided to \model; the joint moments are excluded from training and used solely to evaluate whether the decoded internal forces recover independently derived mechanical structure. Graph construction, observed inputs, and data splits for all  four systems are specified in Supplementary Information, Sections~4.1--7.1.

Across all four systems, \model represents the observed components as physical nodes connected by their native interaction graph and augments this graph with a single virtual hub connected bidirectionally with virtual edges to every physical node. The physical nodes carry position and velocity together with case-specific attributes, while an edge-type attribute distinguishes physical from virtual edges.

We compare \model with six  graph-based simulation approaches spanning explicit mechanics, learned particle and mesh simulators, hierarchical equivariant models, and temporal neural operators. \dgn~\cite{dgn} is the closest comparison for the proposed integration scheme:  it shares the edge-local frames, antisymmetric momentum fluxes, and rotational channel of \model, but integrates the decoded fluxes explicitly. \gns~\cite{gns} uses an encode, process, and decode graph network in which particles are nodes and Cartesian relative displacements describe their interactions. \mgn~\cite{mgn} applies the same general architecture to simulation meshes, with messages passed along mesh edges and, where required, between points that are close in physical space. \eghn~\cite{eghn} uses equivariant hierarchical pooling, while \egno~\cite{egno} applies an equivariant temporal convolution in the Fourier domain. \eghno, introduced in the same work, combines the temporal operator of \egno with the hierarchical backbone of \eghn. On the clamped beam, we additionally compare against IGNS~\cite{REF_IGNS}, an explicit port-Hamiltonian graph integrator, to determine whether increasing the number of internal integration steps is sufficient for an explicit method spanning the same coarse observation interval. Supplementary Information, Section~2 documents the baseline objectives and shared adaptations, while Sections~4.2--6.2 give the benchmark-specific implementations and adaptations required for autoregressive evaluation.

To make the comparison controlled, all models are trained on identical trajectories, data splits, and use the same optimizer, early-stopping criterion, and rollout protocol. We retain their published training objectives except where an adaptation is required for the common kinematic task; Supplementary Information, Section~2 records which objectives are retained or adapted, and Section~3.3 defines the common autoregressive rollout. Performance is evaluated through full autoregressive rollouts, with each predicted state fed back as input to the next step. Errors in both the learned dynamics and the state update, therefore, accumulate over the complete prediction horizon. We measure error over all body nodes rather than at a single point, preventing locally accurate predictions from masking deformation elsewhere in the system. \egno predicts a window of future states, whereas the remaining models predict one state at a time. Because the published \eghn, \eghno and \egno architectures do not return velocity, we add an equivariant velocity head following the construction of each model's position decoder; Supplementary Table~2 specifies each added head and the representation it reads.

The clamped beam provides a controlled test of the proposed coarse-step integration. Its material parameters and reference time step are known; a temporally converged reference removes any dependence on the reference discretization. Mesh  refinement  then systematically increases $\omega_{\max}$, pushing the system beyond the frequency range encountered during training while progressively tightening the stability limit of explicit integration.  
Human motion capture and \rev{the protein transition} test the same architecture on sparse, irregular graphs (Sections~\ref{sec:mocap} and~\ref{sec:protein}). Finally, we test whether the quantities learned for prediction also recover mechanical quantities that are never provided during training. In human walking, inverse-dynamics joint moments provide an independent reference for joint moments assembled from the internal forces inferred from kinematics alone. In the beam, the learned response operators are compared with  the finite-element \rev{tangent and with the modal structure recovered from the rollout} (Section~\ref{sec:inference}). Conservation and computational cost are examined last (Section~\ref{sec:cost}).

\subsection{Stable beam elastodynamics under extrapolation}
\label{sec:beam}

\begin{figure}[!htbp]
  \centering
  \includegraphics[width=\textwidth]{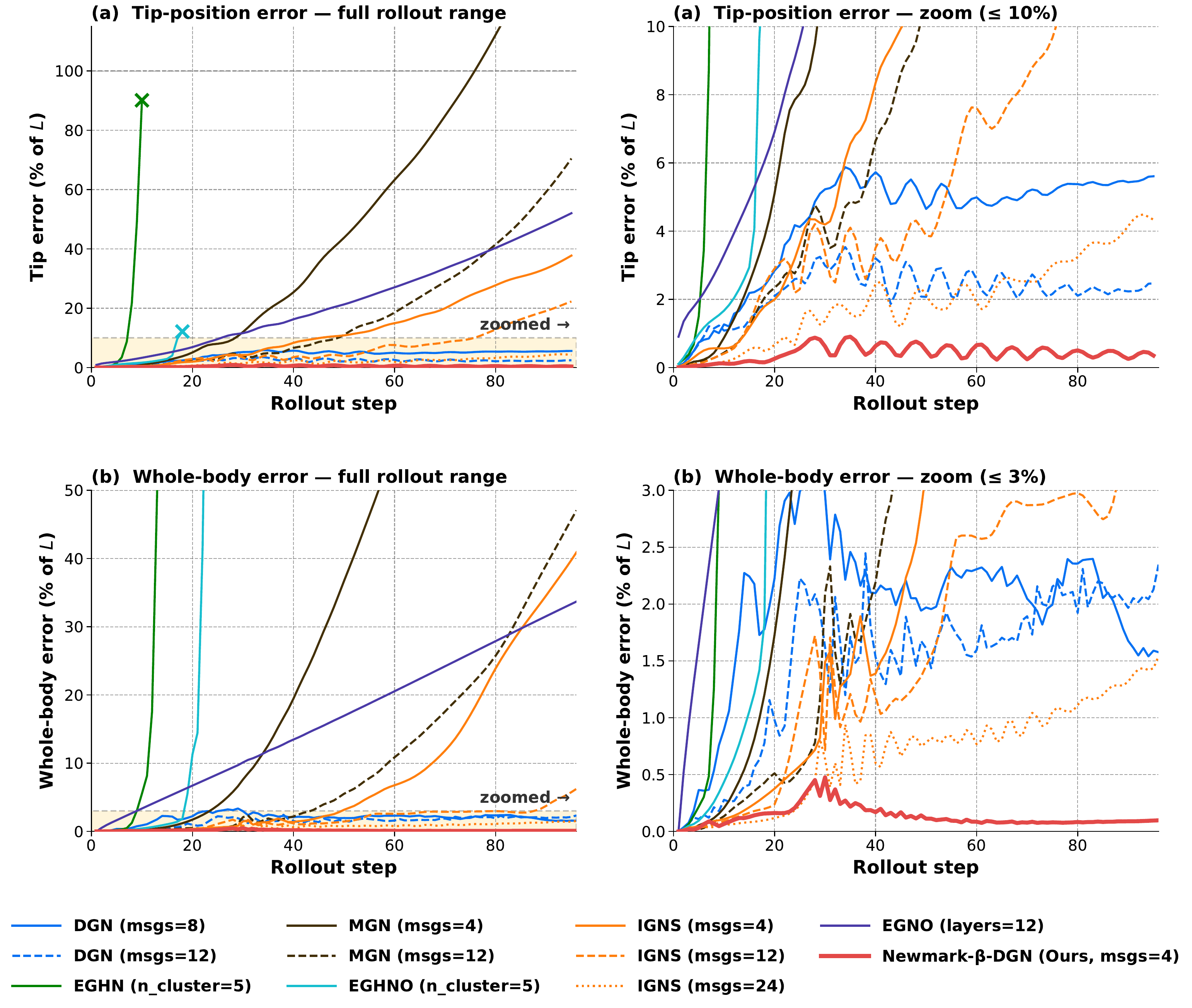}
    \caption{\rev{\textbf{\model remains bounded over the clamped-beam rollout.}
      Rollout position error against step, as a percentage of the beam length $L$, averaged over the 12 held-out test configurations. \textbf{a}, Tip-position error over the full rollout range (left) and zoomed to $\le 10\%$ (right). \textbf{b}, Whole-body error, the root-mean-square error over all body nodes, over the full range (left) and zoomed to $\le 3\%$ (right). \model (red) stays bounded and non-monotone, tracking the beam's oscillation, whereas every baseline grows or diverges; a cross marks the first non-finite value. Line colour denotes architecture and line style the number of message-passing steps or the cluster count. Curves are means over the 12 test configurations at a single training seed (seed 42); no across-seed dispersion is shown.}}
  \label{fig:beam:rollout}
\end{figure}

\begin{figure}[!htbp]
  \centering
  \includegraphics[width=0.85\textwidth]{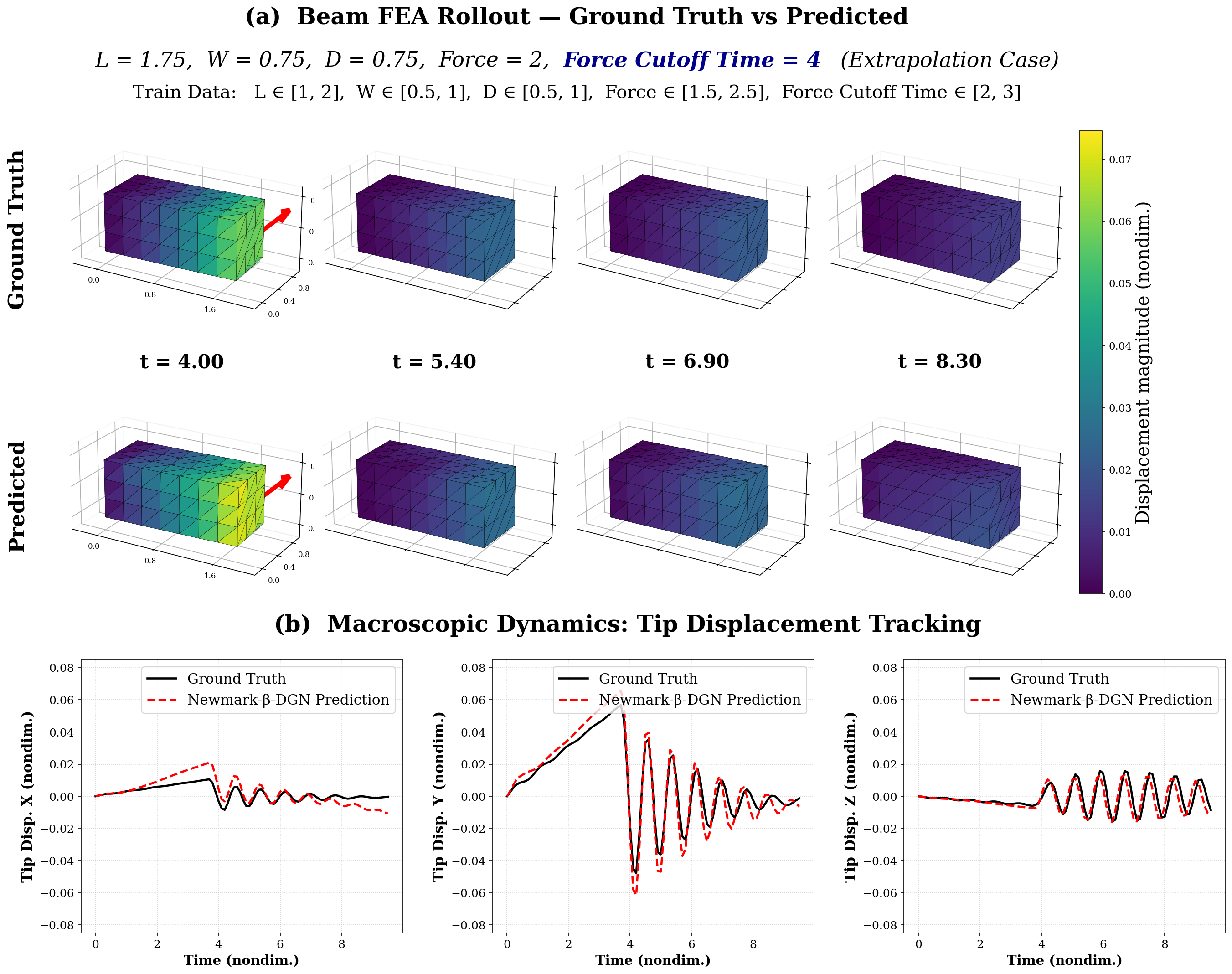}
  \caption{\textbf{Beam rollout under load-duration extrapolation.}
  \textbf{a}, Reference and predicted configurations at four instants of free vibration for
  a held-out case whose load is removed at $4\,$ unit-time, against cut-off times of $2$ to $3\,$ units in
  training. \textbf{b}, Components of the tip displacement over the same trajectory: the
  transverse response follows the phase and decaying envelope of the reference while a small
  axial drift accumulates. Single seed ($n=1$, seed 42); no dispersion is shown.}
  \label{fig:beam:extrap}
\end{figure}

\begin{figure}[!htbp]
  \centering
  \includegraphics[width=0.9\textwidth]{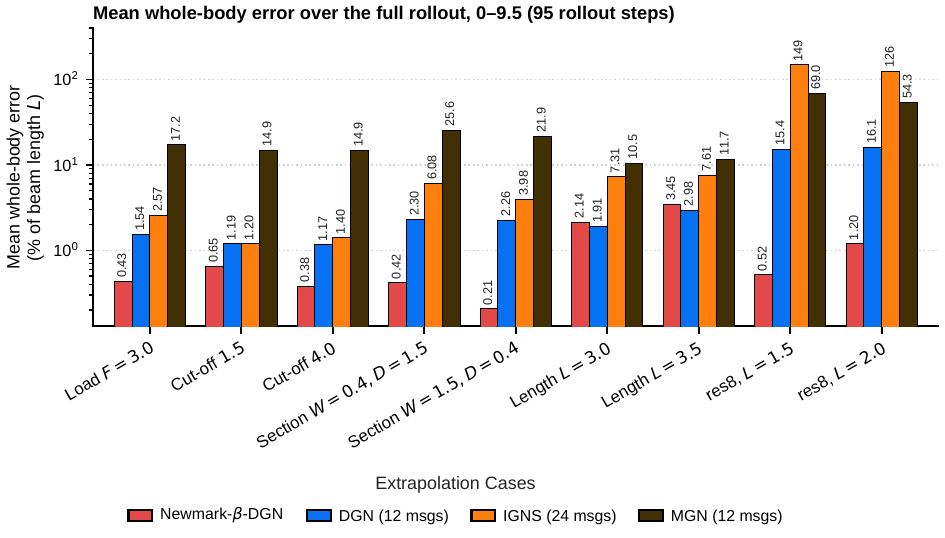}  
  \caption{\textbf{Whole-body rollout error on the nine extrapolation cases.}
  Each case extrapolates outside the training range in load, cut-off time, cross-section, length or mesh
  resolution.   Bars show the whole-body error averaged over the 95-step rollout ($0$--$9.5\, \text{time-units}$), expressed as a percentage of the beam length $L$; the y-axis is logarithmic. \model uses four Newmark sub-steps, \dgn and \mgn twelve message-passing steps, and IGNS twenty-four internal steps, six times \model's budget. Single seed ($n=1$, seed 42);}
  \label{fig:beam:extrapolation}
\end{figure}

\begin{figure}[!htbp]
  \centering
  \includegraphics[width=1.0\textwidth]{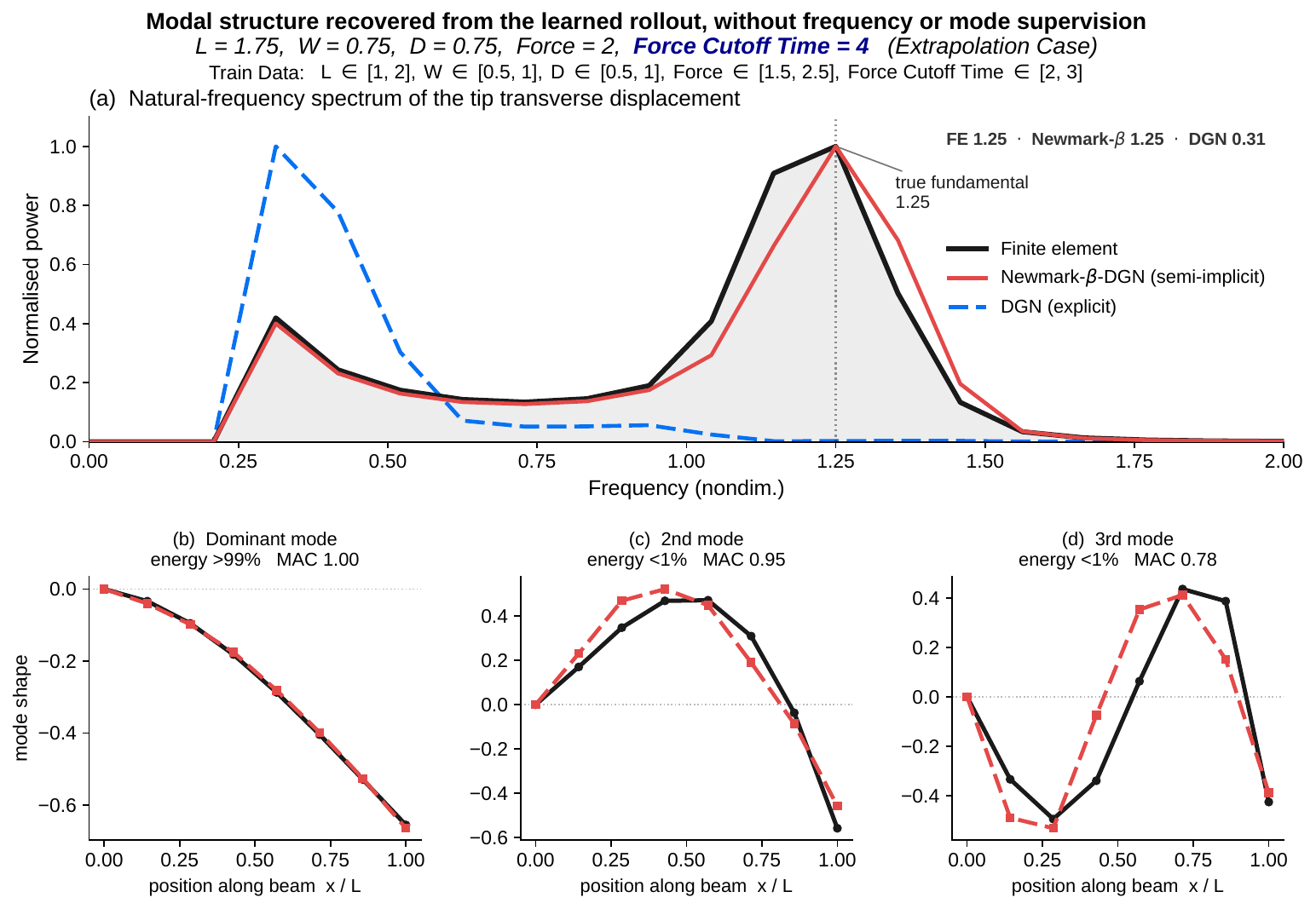}
  \caption{\rev{\textbf{Modal structure recovered from the rollout.}
  \textbf{a}, Natural-frequency spectrum of the tip transverse displacement for the finite-element
  reference (black), \model (red) and \dgn (blue dashed), with the fundamental frequencies annotated and
  the true fundamental marked. \textbf{b}--\textbf{d}, The first three proper-orthogonal mode shapes,
  finite element against \model, each labelled with its energy fraction and Modal Assurance Criterion.
  \model matches the finite-element fundamental and dominant mode; \dgn peaks at a spurious lower
  frequency. No frequency or mode is supervised. Single seed ($n=1$, seed 42).}}
  \label{fig:beam:modal}
\end{figure}

The beam case study considers a three-dimensional cantilever clamped at one end and driven by a transverse load at the other. The load is removed at a cut-off time $T_c$, after which the beam vibrates freely under Rayleigh damping. \rev{All quantities in this beam benchmark are nondimensional; geometric, load, material and temporal values are therefore reported without physical units.} Reference trajectories are generated with a finite-element solver~\cite{fenics,bleyer2018} on a tetrahedral mesh at $\dt=0.1\,$ for 100 steps. The in-distribution dataset comprises 108 trajectories spanning $L\in\{1.0,1.5,2.0\}$, $W,D\in\{0.5,1.0\}$, $F\in\{1.5,2.0,2.5\}$ and $T_c\in\{2.0,2.5,3.0\}\,$ at four elements per unit length, divided into 86 training, 10 validation and 12 test configurations. Nine additional configurations are held out entirely from these sets to test extrapolation beyond the training domain in operating conditions, geometry, and spatial discretizations. They include a 20\% increase beyond the maximum training load (\(F=3.0\) versus \(2.5\)); load removal outside the trained temporal range, occurring 25\% earlier and 33\% later than the respective training bounds (\(T_c=1.5\) and \(4.0\,\) versus \(2.0\!-\!3.0\,\)); cross-sections for which one dimension is 20\% below the trained minimum while the other is 50\% above the trained maximum, \((W,D)=(0.4,1.5)\) and \((1.5,0.4)\); beams that are 50\% and 75\% longer than the longest training beam (\(L=3.0\) and \(3.5\) versus \(2.0\)); and a twofold increase in mesh resolution (\(\mathrm{res}=8\) versus \(4\)) at \(L=1.5\) and \(2.0\). 
Supplementary Information, Section~4.1 specifies the finite-element model, material parameters, observed inputs and data splits.

The prescribed step is large relative to the mesh time scale. With non-dimensional $E=1000$ and $\rho=1$ the elastic wave speed is $c=\sqrt{E/\rho}=31.6$, so a wave crosses a unit-length beam in $0.032\,$  and one outer step spans roughly three such transits. \model does not advance this interval in a single step, but divides it into four equal semi-implicit Newmark substeps of \(\dt_s=\dt/4=0.025\,\). A complementary spectral estimate is obtained from the largest natural angular frequency of the discretized finite-element system, \(\omega_{\max}\), computed from the generalized eigenvalue problem \(K\phi=\omega^2M\phi\). For the corresponding explicit update, the critical step size is \(\delta t_{\mathrm{crit}}=2/\omega_{\max}\), giving approximately \(0.0045\,\) on the training mesh and \(0.0023\,\) on the refined mesh.  The outer step \(\Delta t=0.1\,\) exceeds the explicit stability limit $\dt\,\omega\le2$ by a factor of $22$ on the training mesh and $44$ on the refined mesh; even with four \model substeps, the corresponding factors remain \(5.5\) and \(11\). 
Supplementary Information, Section~4.7 derives this spectral estimate, and Supplementary Table~\rev{7} reports the frequencies, stability multiples and amplification factors. Both the finite-element solver and \model use the average-acceleration Newmark parameters $\gamma=\tfrac12$ and $\beta=\tfrac14$. 

The finite-element mesh defines the graph representation: each mesh vertex is represented as a node carrying its current position and velocity, together with a free/clamped boundary indicator and the applied load at that node, while bidirectional edges follow the mesh connectivity. \model augments this graph with one virtual hub connected to every mesh node through directed virtual edges, identified by an edge attribute of $-1$ to distinguish them from the mesh edges. Each node additionally carries an angular-velocity state, initialized to zero and updated latently across the Newmark substeps. The hub carries no external load, and no material properties or finite-element matrices are supplied to the model (Supplementary Table~3 summarizes the beam graph representation). 

Each outer time step comprises four Newmark substeps, each implemented as a single message-passing round with a linearized nodal solve. We roll the model out autoregressively for 95 time steps ($9.5\,\text{time-units}$), covering the free-vibration response until it is substantially attenuated by Rayleigh damping. Over this horizon, we evaluate both the loaded-tip trajectory and the deformation of the full mesh. Because a metric evaluated at a single point can remain small while the interior of the mesh over-deforms, we report the whole-body error, defined as the root-mean-square position error over all body nodes as a percentage of the beam length. Supplementary Information, Section~4.3 lists the model hyperparameters. All beam results use one trained seed ($n=1$, seed 42), so no dispersion is reported.



On the test cases, the error of \model relative to the finite element trajectory remains below approximately $1\%$ of the beam length throughout the 95-step autoregressive rollout. Figure~\ref{fig:beam:rollout} shows the evolution of the tip and whole-body errors, averaged over the test cases: for \model, shown in red, both errors are non-monotone, rising in the first half and falling afterwards, indicating that it tracks the beam's oscillation rather than accumulating the error in position. 

The same rollout behavior extends beyond the training domain. We define the whole-body error as the root-mean-square position error over all body nodes, expressed as a percentage of the beam length. Across the nine held-out configurations, its mean value over the 95-step rollout ranges from $0.21\%$ to $3.45\%$ (red bars in Fig.~\ref{fig:beam:extrapolation}). A representative load-duration extrapolation case is shown in detail in Fig.~\ref{fig:beam:extrap}. Panel~\textbf{a} shows the finite-element and predicted beam configurations at four instants of the rollout, colored by displacement magnitude, and panel~\textbf{b} shows the corresponding $x$, $y$ and $z$ components of the tip displacement. When the load is removed at $4\,$ time-units, one unit beyond the latest training cut-off time, the predicted tip displacement continues to follow the finite-element free-vibration response, with only a small axial drift developing. Supplementary Figure~2 shows the same rollout visualization for four additional held-out cases: a refined mesh at $L=1.5$ and $\mathrm{res}=8$, a longer beam with $L=3.0$, and the two cross-section extrapolations $(W,D)=(1.5,0.4)$ and $(0.4,1.5)$. Across all four cases, the predicted response remains close to the finite-element trajectory, although larger but still bounded deviations appear for the longer beam.

We next compare the \model with the baseline models introduced in Section~\ref{sec:overview}. All receive the same node features and mesh graph but without the virtual hub. Supplementary Information, Section~4.2 details their inputs and adaptations. Figure~\ref{fig:beam:rollout} shows that the baselines exhibit distinct forms of autoregressive rollout error growth across the same test cases. \eghn and \eghno reach non-finite values at steps 11 and 19, respectively. Their behavior shows that shortening spatial communication paths through hierarchical processing, even when combined with a temporal operator in \eghno, is not sufficient by itself to prevent rollout error amplification. The error of \egno remains finite, but its tip and whole-body errors grow steadily, showing that a temporal operator over the prediction interval likewise does not prevent error accumulation during autoregressive rollout. Increasing the message-passing depth of \mgn from 4 to 12 rounds slows this growth, yet the 12-round model still reaches approximately $70\%$ tip error and $61\%$ whole-body error by step 95, indicating that deeper local propagation improves the rollout without stabilizing it. For \dgn, increasing the number of explicit substeps from 8 to 12 reduces the final tip error from about $5.6\%$ to $2.2\%$, and the whole-body error from about $3.6\%$ to $1.8\%$. IGNS shows the same general dependence on temporal resolution: increasing its internal-step budget from 4 to 24 symplectic-Euler steps substantially reduces both errors, yet the 24-step variant still reaches about $4.3\%$ tip error and $2.6\%$ whole-body error by step 95. Thus, finer temporal subdivision improves both \dgn and IGNS, but their errors continue to accumulate over the rollout rather than exhibiting the low, non-monotone behavior of the proposed \model.

The extrapolation comparison in Fig.~\ref{fig:beam:extrapolation} shows the same performance gap between \model and the baselines. \dgn is the closest competitor, with a lower mean whole-body error than \model only for the two longest beams, $L=3.0$ and $L=3.5$; IGNS is less accurate throughout and \mgn has the largest errors. Mesh refinement is the sharpest test of stiffness: it lowers the critical explicit time step $\dt_{\mathrm{crit}}$ the most and is the only extrapolation that makes the mesh stiffer than any training case. On the refined mesh case, \model stays at $0.5\%$--$1.2\%$ whole-body error, against $15\%$--$16\%$ for \dgn, $54\%$--$69\%$ for \mgn and $126\%$--$149\%$ for IGNS, even at 24 internal steps, six times \model's budget. Adding internal steps for IGNS does not close this gap: a larger IGNS budget lowers its single-step validation loss without stabilizing the rollout (Supplementary Information, Section~4.7 and Supplementary Table~8). The reason is spectral: at the four-substep budget, each explicit symplectic-Euler substep multiplies any error in the stiffest vibration mode by $109$ on the training mesh and by $455$ on the refined mesh, so the error grows within a few substeps. In contrast, for the corresponding classical undamped linear mode, the average-acceleration Newmark update has unit amplification, so its amplitude is neither amplified nor numerically damped, although phase error remains; Supplementary Information, Section~10.4 derives this unit spectral radius and the associated phase error.

On the two longest beams in the extrapolation regime, $L=3.0$ and $L=3.5$, \dgn with twelve message-passing rounds attains a slight advantage compared to \model, for example, the whole body error of $1.9\%$ versus $2.1\%$ at $L=3.0$. \model couples distant nodes through a single operator-weighted hub, a rank-one approximation of the global coupling; for longer beams, whose bending response spans the full length, this coupling may require a higher-rank representation, whereas \dgn reaches the same range through progressive local message passing. Increasing beam length is therefore the regime in which the advantage of the hub-based approximation gets diminished.



To identify which components of \model sustain its rollout accuracy across both the test and extrapolation cases, we perform a staged ablation, reporting the whole-body error averaged over all 21 test and extrapolation configurations. Starting from \dgn, which is hub-free and uses twelve explicit substeps, the mean whole-body error is $3.00\%$. Adding the operator-weighted virtual hub with an explicit per-node update ($\mathbf{A}_i=\mathbf{I}$) and only four substeps reduces the error to $1.26\%$. Replacing this update with the semi-implicit Newmark solve lowers it further to $0.70\%$ with $\beta=0$ and to $0.54\%$ with $\beta=\tfrac14$ for the full \model. Supplementary Information, Section~4.5 and Supplementary Table~5 report this four-stage ablation. The operator-weighted hub provides the largest accuracy gain, with further improvements from the semi-implicit solve and stiffness term.

\rev{The ablation establishes how the hub and semi-implicit update reduce whole-body rollout error. We next ask whether the resulting trajectory also preserves the beam's modal structure, despite receiving no frequency or mode supervision (Fig.~\ref{fig:beam:modal}). From the tip-displacement spectrum of the $95$-step rollout \model reproduces the finite-element fundamental frequency, and a proper-orthogonal decomposition of the transverse motion recovers the dominant mode shape at a Modal Assurance Criterion of $1.00$. The explicit \dgn rollout instead peaks at a spurious lower frequency, so a bounded trajectory needs not carry the correct spectrum. This is the empirical counterpart of the unit amplification of the average-acceleration update: the period elongation of a direct integrator is a property of the scheme rather than of its one-step accuracy~\cite{REF_HUGHES,REF_BATHE,REF_HHT}. Supplementary Information, Section~4.8 reports the frequency and mode recovery for every configuration (Supplementary Table~9 and Supplementary Figure~3).}


\subsection{Human motion prediction benchmark}
\label{sec:mocap}

\begin{figure}[!htbp]
  \centering
  \includegraphics[width=0.8\textwidth]{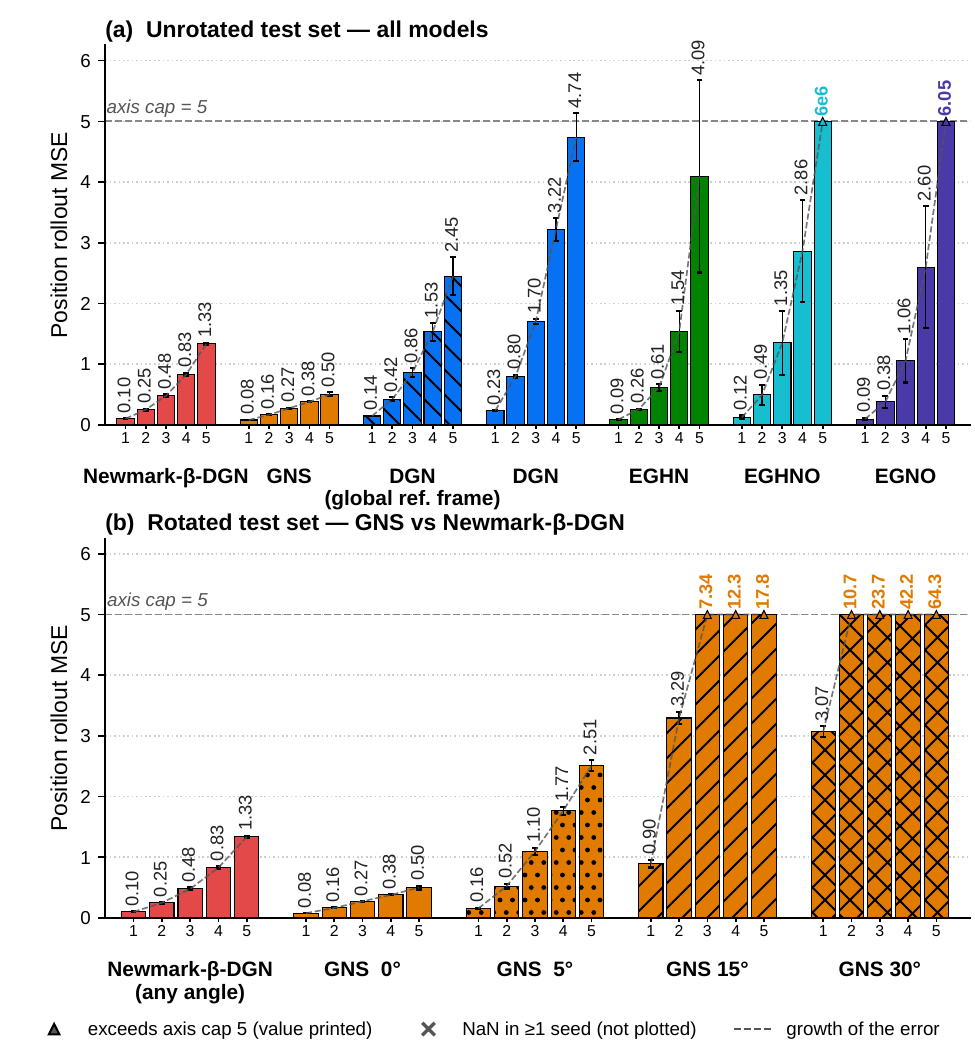}
  \caption{\textbf{Motion-capture rollout accuracy and rotation robustness.}
  \textbf{a}, Position rollout MSE on the unrotated test trajectories. Each group represents one model, and its five bars correspond to successive rollout steps of 30 frames each. \rev{\dgn~(global reference frame) uses \model's global reference frame in \dgn's external-force channel; both are rotation-equivariant.} \textbf{b}, Position rollout MSE after fixed rotations of the test trajectories about the vertical axis. The single \model group applies at every angle because its prediction is rotation-equivariant. Both panels are capped at \rev{$5$}; \rev{triangles} mark values beyond this limit, with the true value printed above. A step is marked as non-finite and omitted if any seed diverges. Bars show means over seeds $\{0,42,100\}$ and error bars show one standard deviation ($n=3$).}
  \label{fig:mocap}
\end{figure}

The beam tests coarse-step prediction on a discretized structure with a defined physical interaction graph. Articulated human motion presents a different setting: observations are sparse markers linked by a skeletal graph, and the model must predict coordinated movement across the body over an interval containing many recorded frames. We use this setting to test whether the virtual hub supports long-range coordination without a dense interaction graph. We evaluate \model on the walking sequences of subject 35 from the CMU Motion Capture Database~\cite{cmumocap} and follow the trial-level split and prediction interval of \eghn~\cite{eghn}. One rollout step predicts the marker configuration 30 frames ahead, and every model is trained from seeds $\{0,42,100\}$. The body is represented by 31 marker nodes and 60 directed skeletal edges. Each node carries its current position and a velocity computed as the backward difference of positions. \model adds a single virtual hub connected to all markers through 62 directed edges; these virtual edges are separated from skeletal edges by an edge attribute. While, the baselines add 70 two-hop edges in addition to the skeletal edges and separate them by an edge attribute. 
Supplementary Information, Section~5.1 defines the observed graph and data split, while Section~5.3 lists the training configurations.

Because the external forces acting on the walking body are unobserved, \model infers them from the current state. To express their directions without introducing a fixed Cartesian reference, it constructs a parameter-free spatial frame from the current marker positions and the two leading non-trivial eigenvectors of the body-graph Laplacian. The resulting frame rotates with the body, is invariant to translation, while the eigenvectors defining its construction depend only on the graph topology. We compare our model with \gns~\cite{gns}, \eghn~\cite{eghn}, \eghno~\cite{egno}, \egno~\cite{egno}, \dgn~\cite{dgn}, and \dgn (global reference frame), a  \dgn variant that uses the same graph-level equivariant reference frame as \model to represent the external force channel. Because the underlying \dgn interaction model remains unchanged, the \dgn~(global reference frame) variant isolates the effect of expressing the external force in the graph-level equivariant frame used by \model instead of \dgn's original reference frame. The original published configurations of \eghn, \eghno and \egno receive absolute height, while \dgn constructs it internally; neither \gns nor \model receives an absolute coordinate. The rotations used in the following experiment preserve height, so this difference in available information remains constant.

\rev{On unrotated trajectories, \model achieves the lowest step-five MSE among the equivariant models, at $1.335$ (Fig.~\ref{fig:mocap}a). \gns reaches a lower  error of $0.5006$ on these trajectories, but is not rotation-equivariant: although translation-invariant through its relative inputs, its encoder operates on Cartesian vector components and its decoder predicts Cartesian vectors directly. We therefore examine whether this advantage persists when the observation frame is rotated. A rotation of just $5^\circ$, which is small enough to arise from a slight misalignment of the motion-capture setup, raises the step-five error of \gns to $2.5097$, above the unrotated error of \model's $1.335$. At $15^\circ$ and $30^\circ$, the GNS error is  $35.6$ and $128$ times its unrotated value, respectively. Figure~\ref{fig:mocap}b shows the rotation sweep against the angle-invariant error of \model. Supplementary Table~13 reports the full rotation sweep for GNS, \model is rotation-equivariant so its error is unchanged at every angle, while Supplementary Information, Section~5.4 provides the unclipped rollout results and rotation analysis.}

\subsection{Protein dynamics prediction benchmark}
\label{sec:protein}
\rev{We next evaluate \model on a large collective conformational change of the protein adenylate kinase, from its closed to its open conformation. We use the trajectory dataset of Seyler and Beckstein~\cite{seyler2017adk}, accessed through MDAnalysis~\cite{mdanalysis2011,mdanalysis2016}, comprising 200 trajectories of the protein moving from the closed to the open state. Each trajectory contains the 855 backbone atoms over 90--106 frames, during which the two mobile domains move away from the rigid core. At each prediction step, the model forecasts the protein configuration 15 frames into the future. Repeating this autoregressively for five steps yields a total prediction horizon of 75 frames, covering most of the closed-to-open transition. We split the data by trajectory into 140 training, 30 validation and 30 test trajectories. Each node represents a backbone atom and carries its position, backward-difference velocity, and atom type. \model represents local interactions using the covalent-bond graph and adds a single virtual hub connected to all atoms for system-wide communication. \dgn uses the same covalent-bond graph but without the hub. In contrast, \eghn and \egno augment the covalent bonds with their published \(10\,\text{\AA}\) distance-based contact graph, producing a substantially denser interaction graph.
 Supplementary Information, Section~6.1 specifies the dataset and data split, while Section~6.3 lists the training configurations.}

\rev{All four models use the same trajectory-level data split and are trained with a single seed ($n=1$, seed $42$), retaining the checkpoint with the lowest validation loss. Their training objectives follow their respective formulations: \model predicts position and velocity increments, \dgn uses the same increment objective on the hub-free graph, \eghn predicts the end-state position and velocity together with its auxiliary link-prediction objective, and \egno retains its multi-frame prediction objective. Despite these differences in training objective, all models are evaluated identically using unscaled position MSE over the same autoregressive rollout. Because each result comes from a single trained seed, no across-seed dispersion is reported.}

\begin{figure}[tbp]
  \centering
  \includegraphics[width=\textwidth]{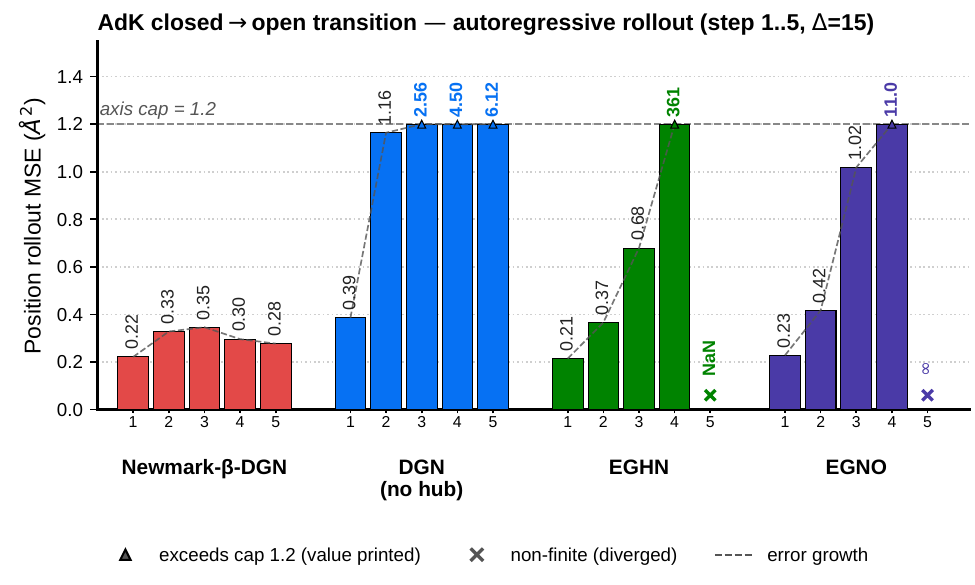}  
  \caption{\rev{\textbf{Autoregressive rollout on the adenylate-kinase closed--open transition.}
  Position mean squared error over five autoregressive steps of $15$ frames each, for \model and the three
  baselines from their minimum-validation checkpoints. \model stays bounded across the horizon, while \dgn diverges on the hub-free covalent graph and \eghn and \egno operating on dense contact-distance based graphs diverge: the ordinate is capped at $1.2\,$\AA$^2$,
  triangles mark values above the cap with the true value printed, and crosses mark a non-finite (diverged)
  step. One seed ($n = 1$, seed $42$). The full unclipped table and the mean-trajectory floor (the average-displacement baseline) are given in
  the Supplementary Information (Section~6.4).}}
  \label{fig:protein}
\end{figure}

\rev{\model remains bounded across the full horizon, rising from $0.221$ to a maximum of $0.346$ and settling near $0.28\,$\AA$^2$ (Fig.~\ref{fig:protein}). The \dgn remains finite at one step ($0.386$), but its rollout climbs monotonically to $6.12$; \eghn and \egno track \model to the third step and then diverge, reaching $361$ and $11.0$ at step four before becoming non-finite. The hub supplies the global coupling through $1710$ virtual edges, against the roughly $55\,610$ directed edges the contact-distance based baselines carry: providing better stability at a fraction of the graph density.}

\begin{figure}[tbp]
  \centering
  \includegraphics[width=\textwidth]{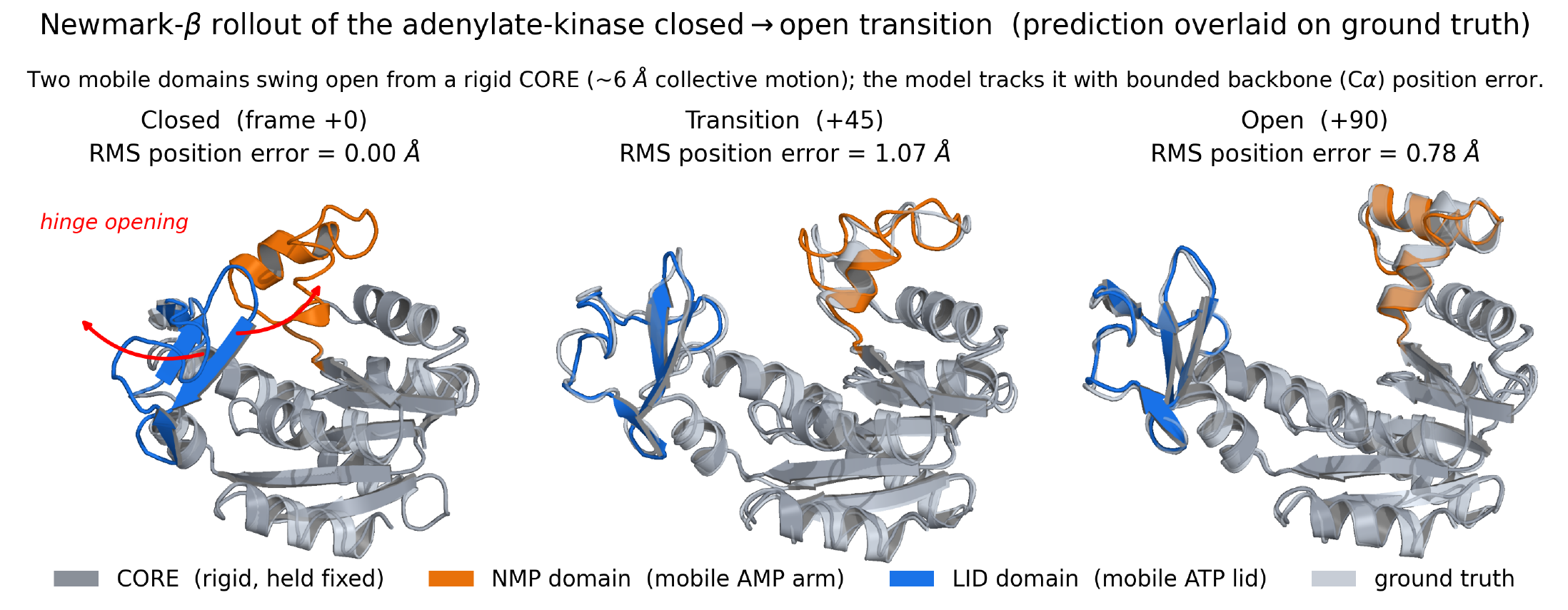}
  \caption{\rev{\textbf{Structural rollout of the closed--open transition.} \model's predicted backbone,
  coloured by domain (rigid core grey, NMP arm orange, LID blue), overlaid on the ground-truth conformation
  (light grey) at the closed start (frame $+0$), the transition midpoint ($+45$) and the open end ($+90$) of
  a held-out trajectory, superposed on the core. The two mobile domains swing away from the core through a
  collective motion of about $6\,$\AA{}, tracked with a root-mean-square C$\alpha$ position error of
  $1.07\,$\AA{} at the midpoint and $0.78\,$\AA{} at the open state.}}
  \label{fig:protein:cartoon}
\end{figure}

\rev{Figure~\ref{fig:protein:cartoon} renders the predicted structural transition for a held-out trajectory. For visualization, we extend the rollout to six autoregressive steps, reaching frame \(+90\). \model reproduces the collective opening in which the two mobile domains, the NMP arm and the LID, swing away from the core; the predicted backbone tracks the ground-truth conformation with a root-mean-square C$\alpha$ position error of $1.07\,$\AA{} at the midpoint of the transition and $0.78\,$\AA{} at the open state.}

\rev{To put this prediction accuracy in perspective, we find that the margin over a trivial predictor is small. A state-independent baseline that applies the training-averaged displacement at each step reaches $0.23$ to $0.37\,$\AA$^2$ across the horizon, and \model stays within $4$ to $10\%$ of it, so the benchmark separates the models by bounded rollout and graph sparsity rather than by a large accuracy margin (Supplementary Table~16). We use the closed-to-open transition trajectories rather than the established equilibrium benchmark because, at its 15-frame horizon, the equilibrium motion is largely decorrelated from its preceding trajectory and is dominated by fluctuations around the mean conformation. A simple linear restoring rule therefore outperforms all learned models, indicating that the benchmark mainly measures equilibrium fluctuation statistics rather than predictive collective dynamics (Supplementary Information, Section~6.5).}

Together, these two benchmarks, in addition to the finite-element beam case, demonstrate the applicability of \model across systems with markedly different structures and dynamics.

\subsection{Internal mechanics inferred from observations}
\label{sec:inference}

A model can reproduce observed motion while assigning mechanically implausible forces to the interactions that generate it. We therefore test the decoded quantities against mechanical references withheld from training. Human walking provides a stringent setting: marker trajectories describe the motion, while synchronized ground-reaction measurements and inverse-dynamics estimates of joint moments provide separate evidence about the forces and moments involved. To make this comparison, we introduce a graph-based benchmark using the instrumented-treadmill recordings of van der Zee et al.~\cite{vanderzee2022}. For subject p2, we convert 33 walking trials, containing 18,631 marker frames, into graph sequences sampled at $120\,$Hz. Each graph contains 37 marker nodes and 63 anatomical edges, augmented by a single virtual hub connected to all markers through bi-directional virtual edges. The measured joint moments and ground-reaction forces are retained as independent references and are not supplied to \model during training. Supplementary Information, Section~7.1 describes the graph construction, gait-condition split, deterministic evaluation windows and moment readout; Supplementary Table~17 lists the observed inputs.

The training and validation sets contain preferred walking over a range of speeds and cadence variants at $1.25\,\mathrm{m\,s^{-1}}$. We withhold every constant-step-length and constant-step-frequency trial, together with preferred walking at $2.0\,\mathrm{m\,s^{-1}}$. The first two groups change the gait strategy, whereas the last changes the speed within a familiar strategy. Each trial is evaluated using deterministic, non-overlapping windows that cover the recorded gait cycle rather than randomly selected starting frames. \model receives only marker positions and velocities obtained by finite differencing. Its loss contains the normalised errors in position and velocity increments and no force, moment or ground-reaction term.

The joint-moment readout uses the internal segment forces decoded at each step of a five-step autoregressive rollout. These are the same forces that \model uses to advance the markers. For a joint $J$, we sum the orbital moment of the forces acting on the markers distal to that joint (the markers on the limb segments below the joint),
\begin{equation}
\mathbf{M}_J(t)
=
-\sum_{i\in\mathcal{D}(J)}
\left(\mathbf{x}_i(t)-\mathbf{x}_J(t)\right)
\times
\mathbf{F}_i(t).
\label{eq:joint_moment}
\end{equation}
This produces a joint-moment vector throughout the rollout. We retain its flexion-extension component, which corresponds to sagittal-plane motion, and compute its root-mean-square value over the gait cycle. Applying the same reduction to the inverse-dynamics measurement gives one joint-demand value for each joint, leg and walking condition. We exclude the decoded spin contribution because the nodal spin is an unsupervised angular-momentum ledger rather than an identifiable joint torque. The readout is therefore obtained directly from the forces used by the update, without a separately trained inference head.

\begin{figure}[tbp]
  \centering
  \includegraphics[width=0.95\textwidth]{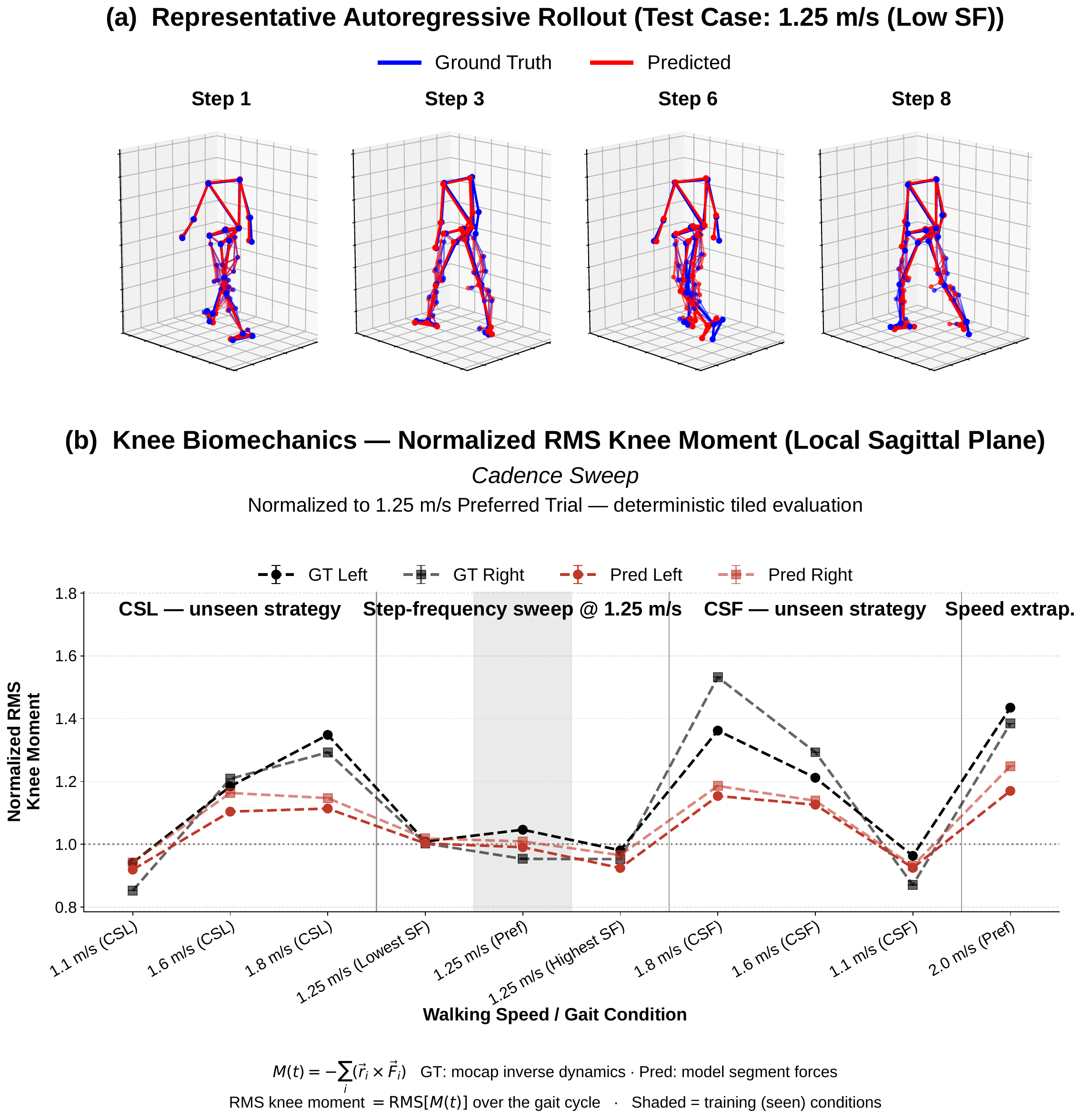}
  \caption{\textbf{Joint-moment demand inferred from marker kinematics alone.}
  \textbf{a}, Representative autoregressive rollout of the marker graph for subject p2 for 8 steps, showing the recorded and predicted configurations. The joint-moment readout is evaluated over the five-step rollout. \textbf{b}, Normalised root-mean-square sagittal knee moment inferred from the decoded internal forces using Eq.~\eqref{eq:joint_moment}, compared with the inverse-dynamics reference. Constant step length and constant step frequency are withheld gait strategies. Preferred walking at $2.0\,\mathrm{m\,s^{-1}}$ is a withheld speed within a familiar strategy. The shaded step-frequency sweep at $1.25\,\mathrm{m\,s^{-1}}$ contains training and validation conditions. Demand is normalised to the preferred $1.25\,\mathrm{m\,s^{-1}}$ trial at unit subject mass and is therefore a ratio rather than a calibrated torque. Error bars show variation across deterministic evaluation windows for one trained model and one subject ($n=1$), not variation across training seeds. Neither joint moments nor ground-reaction forces enter the training objective.}
  \label{fig:moments}
\end{figure}

Across five evaluation experiments and for both legs of the subject, the inferred demand profiles correlate with the inverse-dynamics measurements at $r=0.943$ for the hip and $r=0.869$ for the knee. Figure~\ref{fig:moments} shows the knee result across familiar and withheld gait conditions. Under the withheld constant-step-length strategy, the inferred demand rises with walking speed. Under the withheld constant-step-frequency strategy, it follows the measured rise to a maximum at $1.8\,\mathrm{m\,s^{-1}}$. The shaded step-frequency sweep at $1.25\,\mathrm{m\,s^{-1}}$ contains training and validation conditions and provides an interpolation control. There, the readout reproduces the measured minimum near the preferred cadence. Because no moment enters the objective, this non-monotonic response supports the mechanical interpretation of the readout, although it is not an extrapolation result.

At $1.8\,\mathrm{m\,s^{-1}}$ under constant step frequency, the inferred right-knee demand is $1.217$, compared with $1.532$ in the reference, an under-prediction of $21\%$. At the withheld $2.0\,\mathrm{m\,s^{-1}}$ preferred-speed condition, the corresponding difference is $5\%$. The model therefore extrapolates speed within the familiar preferred-walking strategy more accurately than the unseen constrained strategy.

The pooled hip, knee, and ankle results across the different experiments are reported in Supplementary Information, Section~7.4, and Supplementary Figure~\rev{6}. The ankle profile is not recovered: its pooled correlation is $r=0.492$, its amplitude is approximately one order of magnitude below the reference, and one profile is anticorrelated. This can be attributed to the fact that the ankle readout contains only two distal markers.

A separate in-distribution control using walking conditions seen during training is shown in Supplementary Figure~\rev{7}. The inferred hip and knee profiles again follow the inverse-dynamics reference across speed and cadence. At the knee, the correlation is $r=0.81$. The control is therefore no more accurate than the evaluation containing withheld conditions. \rev{We also assess the contribution of the angular-momentum channel to the physical fidelity of the inferred moments. Removing this channel leaves rollout accuracy essentially unchanged but degrades the joint-moment estimates (see ablation experiments detailed in Supplementary Information, Section~8.1 and Supplementary Figure~8.)}

The learned response operators provide a second mechanical readout. The per-node position- and velocity-response operators used by the Newmark solve align with the corresponding finite-element matrices in their dominant direction, with cosine similarities of approximately $0.85$ and $0.92$, and recover their spatial organisation, with correlations of approximately $0.63$ and $0.94$ between nodal traces. They do not recover the directional anisotropy or absolute scale. Moreover, the position-response operator is nearly orthogonal to the pointwise Jacobian of the decoded force, but predicts the change in the model's momentum response to a velocity perturbation with cosine similarity $0.99$ to $1.00$. Supplementary Information, Section~4.9 defines and reports these probes; Supplementary Figure~4 summarizes the comparison with the finite-element tangent. These quantities are therefore response operators of the learned update, not identified material tangents. \rev{They give a relative, interpretive readout of where the structure is stiff or soft, not a calibrated material tangent.}

Thus, \model returns more than a trajectory. On the walking biomechanics benchmark, the joint-moment readouts assembled from the internal forces used to advance the marker graph follow independently measured hip and knee demand across familiar and withheld gait conditions. \rev{On the beam, the same response operators recover the structure of the finite-element tangent, and the rollout reproduces the beam's fundamental frequency and dominant vibration mode (Section~\ref{sec:beam}).} None of these mechanical targets is supplied during training. The model therefore exposes aspects of the mechanics underlying its predictions, rather than providing trajectories alone.

\subsection{Conservation, scaling and computational cost}
\label{sec:cost}

\begin{figure}[htbp]
  \centering
  \includegraphics[width=0.95\textwidth]{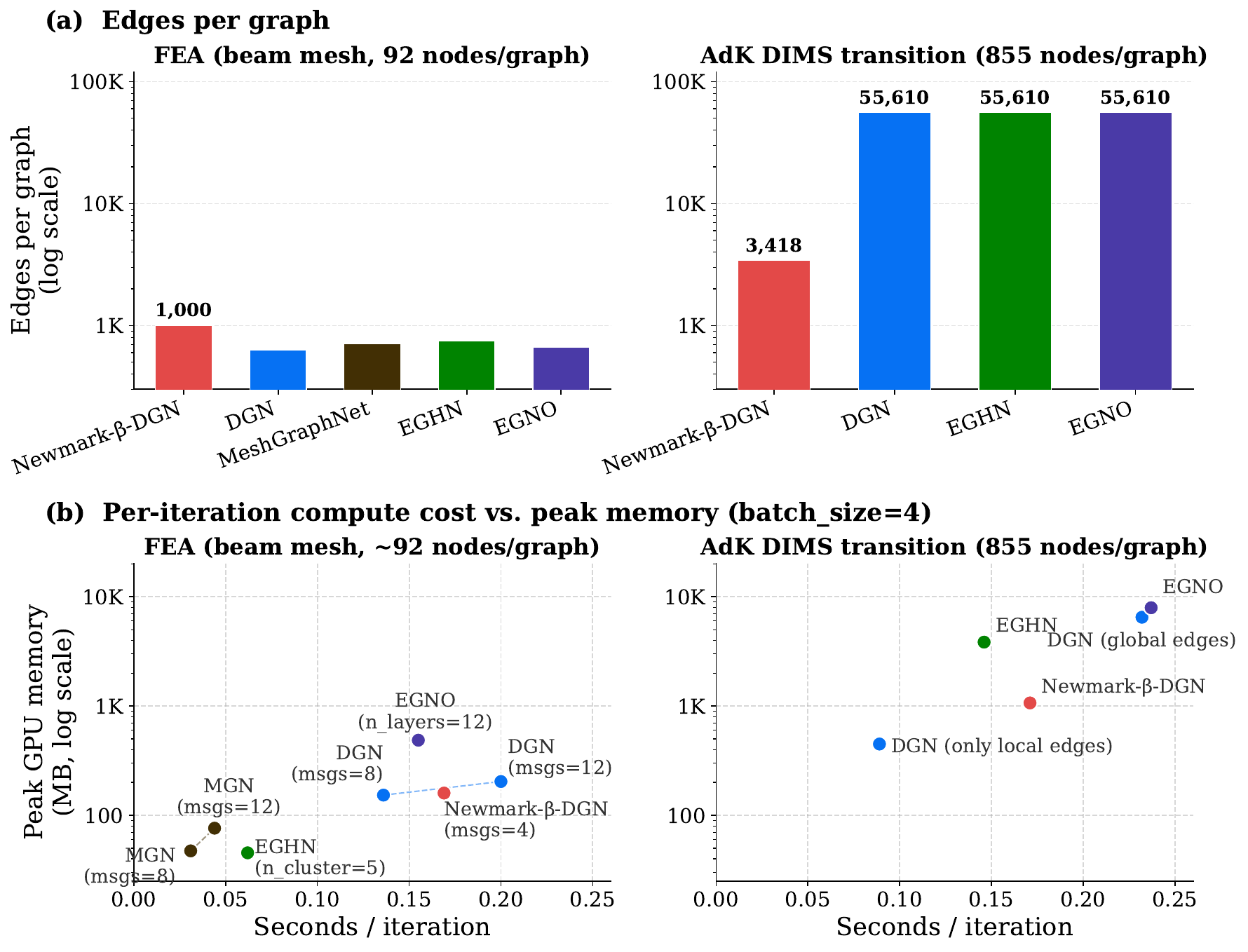}
  \caption{\textbf{Edge scaling and computational cost.}
  \textbf{a}, Directed edges per graph for the beam mesh (left) and the protein backbone (right).
  \model replaces a distance-based contact graph with $2N$ hub edges, which reduces the protein
  edge count by a factor of $16.3$. On the fixed beam mesh, no model uses a distance cutoff and
  no corresponding reduction arises.
  \textbf{b}, Peak memory against wall-clock time per training iteration, using a batch size of
  four and latent widths of $128$ for the protein and $64$ for the beam. Two protein
  configurations are exceptions: \egno and \dgn with global edges require $9622\,\mathrm{MB}$ and
  $10\,048\,\mathrm{MB}$, respectively, at width $128$ and therefore do not fit on the
  $11\,\mathrm{GB}$ device; their plotted points are measured at width $64$.
  Five forward-and-backward iterations are timed after one discarded warm-up iteration, with
  each model profiled in a separate process pinned to a single device and retaining its published
  edge topology. Single measurement per configuration.}
  \label{fig:cost}
\end{figure}

We examine two consequences of the proposed formulation separately: its momentum-conservation
properties and the computational cost of the virtual hub.

\paragraph{Momentum conservation.}
Before time integration, the physical edges conserve linear and angular momentum pairwise, as
inherited from \dgn~\cite{dgn}. For each interacting pair, the two directed edges carry
equal-and-opposite linear- and angular-momentum fluxes and use the same force-application point.
Their contributions therefore cancel exactly when summed over the pair
(Methods, Section~\ref{ssec:m:fluxes}; Supplementary Information, Section~11.1.1).

The virtual hub satisfies a different, collective balance. Its decoded linear- and
angular-momentum fluxes are projected so that the corresponding sums over all hub edges are zero
(Methods, Section~\ref{ssec:m:fluxes}, Equation~\eqref{eq:m:hubprojection};
Supplementary Information, Section~9.3.2). The force projection therefore guarantees that the
hub contributes no net linear momentum. For angular momentum, however, zero summed
angular-momentum flux is not sufficient to guarantee zero total angular contribution: the hub
forces act at different application points and can therefore generate a non-zero net moment about
a common origin. The hub thus enforces collective linear balance and a zero summed
angular-momentum flux, but not exact total angular-momentum balance
(Supplementary Information, Section~11.1.2).

A separate conservation error is introduced by the independent semi-implicit nodal solves. At
node \(i\), the total force drive \(\mathbf{b}_i\) is transformed by the local \(3\times3\)
coefficient matrix \(\mathbf{A}_i^{-1}\), and the velocity update contains an additional
position-response term proportional to \(\Kt_i\mathbf{v}_{i,t}\)
(Methods, Section~\ref{ssec:m:solve}). Before this solve, the physical-edge forces cancel
pairwise and the projected hub-edge forces sum to zero, so
\[
\sum_i \mathbf{b}_i=\mathbf{F}^{\mathrm{ext}},
\]
where reactions at prescribed degrees of freedom are included in
\(\mathbf{F}^{\mathrm{ext}}\).

Exact preservation of this balance would give
\(\sum_i\Delta\mathbf{p}_i=\subdt\,\mathbf{F}^{\mathrm{ext}}\).
Instead, the semi-implicit nodal update gives the residual
\begin{equation}
\mathbf{R}_{P}
=
\subdt\sum_i
\left(\mathbf{A}_i^{-1}-\mathbf{I}_3\right)\mathbf{b}_i
-
\frac{1}{2}\subdt^2
\sum_i
\mathbf{A}_i^{-1}\Kt_i\mathbf{v}_{i,t}.
\label{eq:momentum_residual}
\end{equation}
The two terms have distinct origins. The first appears because the balanced nodal drives
\(\mathbf{b}_i\) are transformed by node-dependent matrices \(\mathbf{A}_i^{-1}\), so their
sum is not generally preserved. The second is the contribution of the position-response term
\(\Kt_i\mathbf{v}_{i,t}\). In a fully assembled system, such internal responses are accompanied
by cross-node reactions that cancel when the coupled equations are summed. The independent
nodal solves omit these cross-node reactions, leaving a finite residual
(Supplementary Information, Section~11.2).

For bounded masses, states and response operators, both terms are
\(\mathcal{O}(\subdt^2)\). The absolute one-substep momentum residual is therefore second order
in the substep size, and its value relative to a non-zero external impulse is first order
(Supplementary Information, Section~11.3). The twelve-node synthetic test confirms this
behaviour. The complete update gives an observed relative-residual order of \(0.84\); retaining
the two contributions separately gives orders of \(0.97\) and \(0.81\), respectively; and
removing both reduces the residual to approximately \(1.7\times10^{-8}\), at numerical
round-off (Supplementary Information, Section~11.4 and Supplementary Table~20).
The residual also vanishes in the explicit-response limit
\(\Kt_i,\Dt_i\rightarrow0\), where the linear-momentum update reduces to the conserving explicit
force update of \dgn, apart from the trapezoidal position advance
(Supplementary Information, Section~11.3).

\paragraph{Computational cost.}
The virtual hub adds \(2N\) directed edges for \(N\) physical nodes, so its additional graph
cost scales linearly with system size. This is advantageous when the hub replaces a much denser
long-range interaction graph. On the protein benchmark, \model uses 1708 covalent-bond edges and
1710 hub edges, giving 3418 directed edges in total, whereas the distance-based baselines use
approximately \(55\,610\) directed edges, or \(16.3\) times as many
(Fig.~\ref{fig:cost}a). This reduction is reflected in memory use: at the common profiling
setting, \model requires \(1067\,\mathrm{MB}\) of peak memory, compared with
\(3843\,\mathrm{MB}\) for \eghn, while \egno and \dgn with global edges exceed the available
memory at the common latent width (Fig.~\ref{fig:cost}b).

This advantage is not expected when the underlying graph is already sparse. On the human
skeleton, the edge budgets are similar (122 versus 130 directed edges), while on the beam no
distance-based contact graph is used and the hub therefore provides no meaningful reduction in
edge count. Consistently, \mgn and \eghn are cheaper than \model per training iteration on the
beam benchmark. The computational benefit of the hub therefore arises specifically when
system-wide communication would otherwise require substantially denser connectivity.

\section{Discussion}\label{sec:discussion}

This study introduces \model, a mechanically structured  framework for learning coarse-step dynamics and examining the mechanical quantities that generate them from discretely sampled trajectories. Coarse observations create two coupled challenges: the mechanical response can evolve substantially within one observed transition, while mechanical influence can propagate across much of the system over the same interval. \model addresses the first through a semi-implicit update inspired by the average-acceleration Newmark method, in which conserved linear and angular-momentum fluxes, observed or decoded external forces, learned inverse mass and inertia, and matrix-valued response operators determine the finite-time state increment. The second is addressed through an operator-weighted virtual hub motivated by the star topology obtained from a rank-one approximation of implicit non-local coupling. Because these learned mechanical quantities are used to generate the predicted state update through a mechanically structured formulation, they remain accessible after training as mechanical readouts.

We test these two architectural components and their mechanical readouts across four settings. The finite-element beam (Section~\ref{sec:beam}) provides a controlled test of coarse-step rollout and extrapolation across loading, geometry, stiffness and mesh resolution. Human motion capture (Section~\ref{sec:mocap}) and protein dynamics (Section~\ref{sec:protein}) test the formulation on systems with different graph structures and scales, while walking biomechanics (Section~\ref{sec:inference}) tests whether quantities learned only through trajectory supervision retain independently verifiable mechanical meaning.

The beam results show why finite-time response and system-wide coupling must be considered together. \model maintains low, bounded error across the test and extrapolation cases and reproduces the fundamental vibration frequency and dominant mode without direct supervision. Increasing explicit temporal resolution does not reproduce this behavior: \dgn and IGNS improve with finer sub-time stepping, but their errors continue to accumulate. with the mean whole-body error decreasing from $3.00\%$ for twelve-substep \dgn to $0.54\%$ for \model. Hierarchical \eghn and \eghno shorten communication paths across the graph, yet both become non-finite during coarse-step rollout. These comparisons show that neither finer explicit integration nor broader communication alone is sufficient. The controlled ablation makes this separation explicit: adding the operator-weighted hub to the explicit update reduces the mean whole-body error from $3.00\%$ to $1.26\%$, introducing the semi-implicit solve reduces it further to $0.70\%$, and the full formulation reaches $0.54\%$. These results support a formulation in which finite-time response and system-wide coupling are incorporated jointly through structure inspired by computational mechanics.

The same formulation also applies beyond the continuum setting. On human motion, \model retains rotation-equivariant prediction, while on the 855-node protein system the hub provides system-wide connectivity with 16.3 times fewer edges than the distance-based interaction graphs used by the baselines. These results establish the reach of the formulation across different topologies and scales. The hub’s edge saving, however, is specific to settings in which long-range interactions would otherwise require a dense contact graph; it does not arise on a fixed mesh.

The stronger test is whether a model trained only on trajectories can infer forces and response operators that were never observed during training, and whether those inferred quantities agree with independent mechanical references. Joint-moment estimates assembled from internal forces inferred solely from walking kinematics track independently derived hip and knee demand trends, with correlations of $r=0.94$ and $r=0.87$, despite receiving no force or moment supervision. Removing the angular-momentum channel leaves rollout accuracy essentially unchanged while degrading the inferred joint moments, further separating predictive and mechanical fidelity. On the beam, the learned response operators recover relative directional and spatial organization of the finite-element response, but not its anisotropy or absolute scale. These results show that mechanically informative quantities can emerge from trajectory supervision when they participate directly in generating the predicted dynamics.

The beam results also define the present limitations and directions for extension. A single rank-one hub becomes less effective for the longest beams, motivating higher-rank or multiple-hub representations that retain more of the non-local interaction. Independent nodal solves avoid a global implicit system but do not guarantee exact finite-step momentum conservation; future work could therefore consider more strongly coupled updates that improve update-level conservation while preserving coarse-step efficiency. Observed external interactions or limited calibration data could additionally anchor the scale of the learned mechanical quantities, extending the present relative readouts towards absolute forces and moments. 

Overall, \model establishes a mechanically structured approach to coarse-step learning in which conserved interactions, finite-time response, and system-wide coupling are learned jointly within the state transition, linking accurate dynamical prediction with access to internal mechanics underlying the observed motion.
By making inferred forces and response operators accessible from trajectory data, \model opens a route to studying mechanical behavior when direct measurements are unavailable. 
These inferred mechanical quantities can reveal how mechanical demand is distributed across the system and how it changes with operating conditions, paving the way for unsupervised virtual sensing for condition monitoring applications. With its demonstrated generalization across unseen loads, operating conditions, and configurations, \model can also support what-if scenario evaluation by showing how mechanical demand redistributes under changed loads, configurations, or operating conditions.

\section{Methods}
\label{sec:method}
\subsection{Graph representation and model inputs}
\label{ssec:m:graph}

\model takes as input a graph representation of the observed system state. We denote the physical graph by
$\mathcal{G}=(\mathcal{V},\mathcal{E})$, where the $N$ nodes in $\mathcal{V}$ represent the discretised system components and the edges in $\mathcal{E}$ represent their direct physical connections. Each physical connection is represented by two directed edges, $i\!\to\!j$ and $j\!\to\!i$. The physical nodes correspond to finite-element vertices for the beam, anatomical markers for the human systems, and backbone atoms for the protein; the corresponding physical connections follow the mesh connectivity, anatomical skeleton, and covalent bonds, respectively.

Each physical node $i$ carries its position $\mathbf{r}_i$ and velocity $\mathbf{v}_i$. Velocities are obtained by causal finite differencing when they are not directly observed. The model additionally carries a spin $\bm{\omega}_i$ as a latent angular-momentum state, initialized to zero at the beginning of each prediction interval. Additional node and edge inputs are case-specific and are given in Supplementary Information, Sections~4.1--7.1.

The physical graph is augmented with one virtual hub connected to every physical node through bi-directional virtual edges, adding $2N$ virtual edges (Fig.~\ref{fig:framework}b). Physical and virtual edges are distinguished by an edge-type attribute. The hub is not a physical body, carries no external load, and is not advanced as an independent degree of freedom. Its position, velocity, and spin are recomputed during the model update as described in Section~\ref{ssec:m:hub}.

\subsection{Model update overview}
\label{ssec:m:overview}

Given the augmented graph at the current observation time, \model predicts the physical-node state after an interval $\dt$. The interval is divided into $S$ substeps of size
\begin{equation}
    \subdt=\frac{\dt}{S},
    \label{eq:m:substep}
\end{equation}
with one complete message-passing round performed at each substep (Fig.~\ref{fig:framework}c). We use $S=4$ throughout this work.

Each substep follows the same sequence. The hub state is first recomputed from the current physical-node states using the nodal response operators retained from the preceding substep; for the first substep, these operators are initialized as the identity. The resulting augmented graph state is then encoded using the equivariant representation of Section~\ref{ssec:m:encode}. From this representation, the edge decoders produce the linear- and angular-momentum fluxes and the translational and rotational response operators, while the node decoder produces the inverse mass and inverse inertia and, when the external force is unobserved, its components in the global frame, as described in Section~\ref{ssec:m:decode}. The virtual-edge fluxes are collectively projected and the edge response operators are aggregated at the physical nodes. Together with the observed or decoded external force, these quantities enter the translational and rotational semi-implicit Newmark updates of Section~\ref{ssec:m:solve}. The updated physical state and the newly aggregated response operators are then carried to the following substep.

After $S$ substeps, the updated physical-node positions and velocities define the prediction at the next observation time. Section~\ref{ssec:m:training} describes how this transition is learned from kinematic supervision and how the mechanical quantities used to generate it remain accessible after training.

\subsection{Equivariant interaction encoding and processing}
\label{ssec:m:encode}

We retain the edge-local scalarisation--vectorisation construction of \dgn~\cite{dgn}. Each directed edge $ij$ is equipped with an orthonormal frame
$\mathbf{R}_{ij}=[\mathbf{a}_{ij}|\mathbf{b}_{ij}|\mathbf{c}_{ij}]$
whose first axis lies along the edge,
\begin{equation}
    \mathbf{a}_{ij}
    =
    \frac{\mathbf{r}_j-\mathbf{r}_i}
    {\lVert\mathbf{r}_j-\mathbf{r}_i\rVert}.
    \label{eq:m:localframe}
\end{equation}
The remaining axes follow the construction of \dgn and make the complete frame rotation-equivariant, translation-invariant, and antisymmetric under interchange of the two connected nodes,
$\mathbf{R}_{ji}=-\mathbf{R}_{ij}$.

The construction of the node embedding depends on whether the external force is observed. When the external force is observed, as for the beam, no global reference frame is required for force inference. The case-specific scalar node features are passed directly through the node encoder to obtain the invariant node embedding $\mathbf{h}_i$, while the observed force is provided separately to the mechanical update.

When the external force is unobserved, as for the human and protein systems, we construct a second, graph-level global frame $\mathbf{R}^{\mathrm{glob}}$. This frame is used both to express the vector-valued node features as invariant scalars and to reconstruct the decoded external force. From the physical-node graph, excluding the virtual hub, we form the combinatorial Laplacian
$\mathbf{L}=\mathbf{D}-\mathbf{A}$, where $\mathbf{A}$ is the adjacency matrix of the physical graph and $\mathbf{D}$ is its diagonal degree matrix, with $D_{ii}=\sum_j A_{ij}$. Its eigenvectors satisfy
\begin{equation}
    \mathbf{L}\bm{\phi}_k
    =
    \lambda_k\bm{\phi}_k,
    \qquad
    0=\lambda_0\leq\lambda_1\leq\cdots,
    \label{eq:m:phi}
\end{equation}
where $\bm{\phi}_k$ is the $k$th eigenvector and $\lambda_k$ its corresponding eigenvalue. The eigenvalue $\lambda_0=0$ corresponds to the constant mode of the connected physical graph.

We retain the first two non-trivial eigenvectors, $\bm{\phi}_1$ and $\bm{\phi}_2$. Because $-\bm{\phi}_k$ is an equally valid eigenvector, its sign is fixed by locating the component with the largest magnitude and multiplying the entire eigenvector by $-1$ when that component is negative.

At each substep, the eigenvectors are combined with the current physical-node positions to form two three-dimensional anchor vectors,
\begin{equation}
    \mathbf{q}_k
    =
    \sum_{i\in\mathcal{V}}
    \phi_k(i)\,\mathbf{r}_i,
    \qquad
    k\in\{1,2\},
    \label{eq:m:anchors}
\end{equation}
where $\mathbf{r}_i\in\mathbb{R}^3$ is the current position of node $i$ and $\phi_k(i)$ is the component of $\bm{\phi}_k$ associated with that node. Since the non-trivial eigenvectors are orthogonal to the constant mode,
\begin{equation}
    \sum_{i\in\mathcal{V}}\phi_k(i)
    =
    0,
    \qquad
    k\in\{1,2\},
    \label{eq:m:phi_zero_sum}
\end{equation}
translating every node by a common vector $\mathbf{c}$ leaves the anchors unchanged,
\begin{equation}
    \sum_{i\in\mathcal{V}}
    \phi_k(i)(\mathbf{r}_i+\mathbf{c})
    =
    \mathbf{q}_k
    +
    \mathbf{c}
    \sum_{i\in\mathcal{V}}\phi_k(i)
    =
    \mathbf{q}_k.
    \label{eq:m:anchor_translation}
\end{equation}

The anchors are orthonormalised to define the frame axes,
\begin{equation}
    \mathbf{e}_1
    =
    \frac{\mathbf{q}_1}{\lVert\mathbf{q}_1\rVert},
    \qquad
    \mathbf{e}_2
    =
    \frac{
    \mathbf{q}_2-
    (\mathbf{q}_2\!\cdot\!\mathbf{e}_1)\mathbf{e}_1
    }{
    \left\|
    \mathbf{q}_2-
    (\mathbf{q}_2\!\cdot\!\mathbf{e}_1)\mathbf{e}_1
    \right\|
    },
    \qquad
    \mathbf{e}_3
    =
    \mathbf{e}_1\times\mathbf{e}_2,
    \label{eq:m:gs}
\end{equation}
where $\cdot$ denotes the Euclidean inner product and $\times$ the three-dimensional cross product. The resulting global frame is
\begin{equation}
    \mathbf{R}^{\mathrm{glob}}
    =
    [\,\mathbf{e}_1\,|\,\mathbf{e}_2\,|\,\mathbf{e}_3\,].
    \label{eq:m:global_frame}
\end{equation}
The frame is translation-invariant by Equation~\eqref{eq:m:anchor_translation}; under a global rotation, the node positions, anchors, and frame axes rotate together, making the construction rotation-equivariant.

For systems using this frame, the node vector features, including velocity and hub-relative position
$\mathbf{r}^{\mathrm{rel}}_i=\mathbf{r}_i-\mathbf{r}_H$,
are projected onto $\mathbf{R}^{\mathrm{glob}}$ to obtain invariant scalar components. These are combined with the scalar node features and passed through the node encoder to obtain $\mathbf{h}_i$.

Given the node embeddings, each directed edge forms its interaction embedding. The vector features of the sender and receiver are projected onto $\mathbf{R}_{ij}$ and $-\mathbf{R}_{ij}$, respectively, to obtain invariant scalar edge features. Following \dgn~\cite{dgn}, these scalars are combined with the symmetric node representation $\mathbf{h}_i+\mathbf{h}_j$, the edge length, and the edge type to obtain the invariant interaction embedding $\bm{\epsilon}_{ij}$.

The recurrent edge processor of \dgn, including its edge-level latent memory, is retained across the $S$ substeps. The first substep initializes the edge latent representation, while subsequent substeps combine the current interaction with the latent state retained from the preceding round. We refer to \cite{dgn} for the complete edge-frame construction, scalarization and recurrent processing.

\subsection{Decoded interactions and response operators}
\label{ssec:m:decode}
\subsubsection{Edge-wise Interaction Fluxes}
\label{ssec:m:fluxes}
From the edge interaction embeddings and the node embeddings, \model decodes the quantities required by the semi-implicit update. On each edge, in addition to the linear and angular-momentum fluxes of \dgn~\cite{dgn}, we decode translational and rotational response operators. From each node embedding, a node decoder produces the node's inverse mass and inverse inertia and, when the external force is unobserved, its external force components in the global frame of Section~\ref{ssec:m:encode}.

For each edge, invariant scalar coefficients are decoded from the invariant interaction embedding $\bm{\epsilon}_{ij}$ and combined with the edge-local basis to reconstruct the linear-momentum flux $\mathbf{f}_{ij}$ and the angular-momentum flux $\mathbf{A}_{ij}$. By the antisymmetric edge-frame construction and the coefficients shared between the two traversals,
\begin{equation}
    \mathbf{f}_{ij}=-\mathbf{f}_{ji},
    \qquad
    \mathbf{A}_{ij}=-\mathbf{A}_{ji}.
    \label{eq:m:fluxbalance}
\end{equation}
The two directions of each physical connection also share a common reference point,
$\mathbf{r}^{0}_{ij}=\mathbf{r}^{0}_{ji}$,
about which the angular-momentum exchange is balanced. The spin torque delivered to node $i$ is obtained by removing the orbital contribution of the force,
\begin{equation}
    \bm{\tau}_{ij}
    =
    \mathbf{A}_{ij}
    -
    (\mathbf{r}_i-\mathbf{r}^{0}_{ij})\times\mathbf{f}_{ij}.
    \label{eq:m:torque}
\end{equation}
Together with the antisymmetric fluxes, these relations give pairwise linear and angular-momentum balance on physical interactions before time integration, as inherited from \dgn~\cite{dgn} (Supplementary Information, Section~11.1.1).

The virtual edges connecting the hub to the physical nodes decode the same fluxes as the physical edges. For these edges, the decoded force and angular-momentum fluxes are projected onto the subspace of zero collective sum, so that the hub only redistributes internal force and torque contributions among the physical nodes,
\begin{equation}
    \widetilde{\mathbf{f}}_{Hj}
    =
    \mathbf{f}_{Hj}
    -
    \frac{1}{N}\sum_k \mathbf{f}_{Hk},
    \qquad
    \widetilde{\mathbf{A}}_{Hj}
    =
    \mathbf{A}_{Hj}
    -
    \frac{1}{N}\sum_k \mathbf{A}_{Hk},
    \label{eq:m:hubprojection}
\end{equation}
with
$\sum_j\widetilde{\mathbf{f}}_{Hj}=\mathbf{0}$ and
$\sum_j\widetilde{\mathbf{A}}_{Hj}=\mathbf{0}$.
Unlike the pairwise balance on physical edges, this balance is collective over all virtual edges. Because the virtual edges have distinct reference points, the zero-sum projection does not by itself guarantee zero total moment about a common origin; the corresponding residual couple is derived in Supplementary Information, Sections~9.3.2 and~11.1.2.

\subsubsection{Edge-wise Response Operators}
\label{ssec:m:tangents}

Alongside the momentum fluxes, each edge interaction embedding is also decoded into four state-dependent matrix-valued response operators that represent the finite-time response to position and velocity increments within the corresponding semi-implicit substep: the translational position- and velocity-response operators $\Kt_{ij}$ and $\Dt_{ij}$, and their rotational counterparts $\Kt^{\mathrm{rot}}_{ij}$ and $\Dt^{\mathrm{rot}}_{ij}$. These operators are learned coefficients of the update rather than derivatives of the decoded force or identified material stiffness and damping matrices.

Each response operator is parameterized independently by six invariant scalars. For a generic operator
\begin{equation}
    \mathbf{W}_{ij}
    \in
    \left\{
        \Kt_{ij},
        \Dt_{ij},
        \Kt^{\mathrm{rot}}_{ij},
        \Dt^{\mathrm{rot}}_{ij}
    \right\},
\end{equation}
the six scalars define a lower-triangular factor
\begin{equation}
    \mathbf{L}^{(\mathbf{W})}_{ij}
    =
    \begin{bmatrix}
        d_0 & 0 & 0\\
        s_1 & d_1 & 0\\
        s_3 & s_4 & d_2
    \end{bmatrix},
    \qquad
    (d_0,d_1,d_2)
    =
    \operatorname{softplus}(s_0,s_2,s_5)+10^{-4},
    \label{eq:m:response_factor}
\end{equation}
and the operator is reconstructed in global coordinates as
\begin{equation}
    \mathbf{W}_{ij}
    =
    \mathbf{R}_{ij}
    \mathbf{L}^{(\mathbf{W})}_{ij}
    \mathbf{L}^{(\mathbf{W})\top}_{ij}
    \mathbf{R}_{ij}^{\top}.
    \label{eq:m:response_operator}
\end{equation}
The positive diagonal of $\mathbf{L}^{(\mathbf{W})}_{ij}$ makes each operator symmetric positive definite, while its construction in the edge-local frame makes it rotation-equivariant. Because the decoder coefficients are shared between the two traversals and $\mathbf{R}_{ji}=-\mathbf{R}_{ij}$, the quadratic construction is unchanged under edge reversal,
\begin{equation}
    \mathbf{W}_{ji}
    =
    \mathbf{W}_{ij}.
    \label{eq:m:response_symmetry}
\end{equation}

The same construction is used on physical and virtual edges; unlike the momentum fluxes, the virtual-edge response operators are not subjected to the zero-sum projection of Equation~\eqref{eq:m:hubprojection}. The edge operators are summed over all edges incident on each physical node,
\begin{equation}
\begin{aligned}
    \Kt_i
    &=\sum_{j\in\mathcal{N}(i)}\Kt_{ij},
    &\qquad
    \Dt_i
    &=\sum_{j\in\mathcal{N}(i)}\Dt_{ij},\\
    \Kt^{\mathrm{rot}}_i
    &=\sum_{j\in\mathcal{N}(i)}\Kt^{\mathrm{rot}}_{ij},
    &
    \Dt^{\mathrm{rot}}_i
    &=\sum_{j\in\mathcal{N}(i)}\Dt^{\mathrm{rot}}_{ij},
\end{aligned}
\label{eq:m:operatoraggregate}
\end{equation}
where $\mathcal{N}(i)$ contains the physical neighbours of node $i$ and the virtual hub. These nodal operators enter both the operator-weighted hub of Section~\ref{ssec:m:hub} and the semi-implicit update of Section~\ref{ssec:m:solve}.

\subsubsection{Node-wise inverse mass, inverse inertia, and external force}
\label{ssec:m:nodedecode}

In addition to the edge-wise quantities, the node embedding $\mathbf{h}_i$ is decoded into the node-wise quantities required by the semi-implicit update. The node decoder produces positive scalar inverse mass and inverse inertia,
\begin{equation}
    \mathbf{M}^{-1}_i
    =
    m_i^{-1}\mathbf{I}_3,
    \qquad
    \mathbf{I}^{-1}_i
    =
    I_i^{-1}\mathbf{I}_3,
    \qquad
    m_i^{-1}>0,\quad I_i^{-1}>0,
    \label{eq:m:mass_inertia}
\end{equation}
where $\mathbf{I}_3$ is the $3\times3$ identity matrix. Their isotropic form preserves rotation equivariance. The inverse mass enters the translational update, while the inverse inertia enters the rotational update in Section~\ref{ssec:m:solve}.

The treatment of the external force depends on whether it is observed. When it is unobserved, the node decoder maps the invariant embedding $\mathbf{h}_i$ to three invariant components
$\mathbf{s}_i\in\mathbb{R}^3$,
which are reconstructed in global coordinates using the frame of Section~\ref{ssec:m:encode},
\begin{equation}
    \mathbf{f}^{\mathrm{ext}}_i
    =
    \mathbf{R}^{\mathrm{glob}}\mathbf{s}_i.
    \label{eq:m:external_force}
\end{equation}
Because $\mathbf{s}_i$ is invariant and $\mathbf{R}^{\mathrm{glob}}$ rotates with the system, the decoded force is rotation-equivariant and translation-invariant. When the external force is observed, no force decoder is used and the observed force is provided directly to the semi-implicit update.

\subsection{Operator-weighted virtual hub}
\label{ssec:m:hub}

The response operators also set the virtual-hub state at the start of each substep. At the first substep, the nodal operators are initialized as the identity, so the hub position, velocity, and spin reduce to arithmetic averages of the corresponding physical-node states. At each subsequent substep, the hub is recomputed from the updated physical-node states using the operators aggregated in the preceding substep; it therefore carries no independently propagated state.

The hub state is
\begin{equation}
\begin{aligned}
    \mathbf{r}_H
    &=\left(\sum_i\Kt_i\right)^{-1}\sum_i\Kt_i\mathbf{r}_i,\\
    \mathbf{v}_H
    &=\left(\sum_i\Dt_i\right)^{-1}\sum_i\Dt_i\mathbf{v}_i,\\
    \bm{\omega}_H
    &=\left(\sum_i\Dt^{\mathrm{rot}}_i\right)^{-1}\sum_i\Dt^{\mathrm{rot}}_i\bm{\omega}_i .
\end{aligned}
\label{eq:m:hub}
\end{equation}
Thus, the hub position, velocity, and spin are weighted by the corresponding nodal response operators. Because the operators are symmetric positive definite, their sums are positive definite and the inverses in Equation~\eqref{eq:m:hub} are well defined. With identity operators at the first substep, these weighted states reduce to arithmetic averages.

The rank-one motivation for this operator-weighted star construction is derived in Supplementary Information, Sections~9.2.1--9.2.2.

\subsection{Semi-implicit nodal Newmark update}
\label{ssec:m:solve}

At each substep, the quantities defined in Sections~\ref{ssec:m:fluxes}--\ref{ssec:m:nodedecode} are combined at each physical node. Using the same incident-edge set $\mathcal{N}(i)$ as in Equation~\eqref{eq:m:operatoraggregate}, the translational force drive and rotational torque drive are
\begin{equation}
    \mathbf{b}_i
    =
    \mathbf{f}^{\mathrm{ext}}_i
    +
    \sum_{j\in\mathcal{N}(i)}
    \mathbf{f}_{ij},
    \qquad
    \bm{\tau}_i
    =
    \sum_{j\in\mathcal{N}(i)}
    \bm{\tau}_{ij}.
    \label{eq:m:drives}
\end{equation}
Here $\mathbf{f}^{\mathrm{ext}}_i$ is the observed or decoded external force of Section~\ref{ssec:m:nodedecode}, while $\mathbf{f}_{ij}$ and $\bm{\tau}_{ij}$ are the interaction force and spin-torque contributions of Section~\ref{ssec:m:fluxes}. For a virtual edge, the projected force and angular-momentum flux of Equation~\eqref{eq:m:hubprojection} are used to form these contributions.

Let $t$ and $t+1$ denote the beginning and end of the current substep, and define
\begin{equation}
    \Delta\mathbf{v}_i
    =
    \mathbf{v}_{i,t+1}-\mathbf{v}_{i,t},
    \qquad
    \Delta\mathbf{x}_i
    =
    \mathbf{r}_{i,t+1}-\mathbf{r}_{i,t}.
    \label{eq:m:increments}
\end{equation}
The aggregated position- and velocity-response operators of Equation~\eqref{eq:m:operatoraggregate} define the change in translational mechanical response over the substep as
\begin{equation}
    \Delta\mathbf{f}^{\mathrm{resp}}_i
    =
    -\Dt_i\Delta\mathbf{v}_i
    -
    \Kt_i\Delta\mathbf{x}_i,
    \label{eq:m:response}
\end{equation}
where $\Delta\mathbf{f}^{\mathrm{resp}}_i$ denotes the response associated with the state increment. Equation~\eqref{eq:m:response} specifies how the learned operators enter the update; it does not identify them as derivatives of the decoded force.

Using the learned inverse mass $\mathbf{M}^{-1}_i$ of Equation~\eqref{eq:m:mass_inertia}, the trapezoidal response balance is
\begin{equation}
    \Delta\mathbf{v}_i
    =
    \mathbf{M}^{-1}_i\mathbf{b}_i\,\subdt
    +
    \frac{1}{2}\subdt\,
    \mathbf{M}^{-1}_i
    \Delta\mathbf{f}^{\mathrm{resp}}_i .
    \label{eq:m:response_balance}
\end{equation}
For the average-acceleration Newmark parameters
$\gamma=\tfrac{1}{2}$ and $\beta=\tfrac{1}{4}$, the corresponding position increment is
\begin{equation}
    \Delta\mathbf{x}_i
    =
    \frac{\subdt}{2}
    \left(
        \mathbf{v}_{i,t}
        +
        \mathbf{v}_{i,t+1}
    \right)
    =
    \subdt
    \left(
        \mathbf{v}_{i,t}
        +
        \frac{1}{2}\Delta\mathbf{v}_i
    \right).
    \label{eq:m:dx}
\end{equation}
Substituting Equations~\eqref{eq:m:response} and~\eqref{eq:m:dx} into Equation~\eqref{eq:m:response_balance} and collecting the unknown velocity increment gives the translational system
\begin{equation}
\boxed{
    \left[
        \mathbf{I}_3
        +
        \frac{1}{2}\subdt\,
        \mathbf{M}^{-1}_i\Dt_i
        +
        \frac{1}{4}\subdt^2\,
        \mathbf{M}^{-1}_i\Kt_i
    \right]
    \Delta\mathbf{v}_i
    =
    \mathbf{M}^{-1}_i\mathbf{b}_i\,\subdt
    -
    \frac{1}{2}\subdt^2
    \mathbf{M}^{-1}_i\Kt_i\mathbf{v}_{i,t}.
}
\label{eq:m:solve}
\end{equation}
Here $\mathbf{I}_3$ is the $3\times3$ identity matrix. The translational state is then advanced as
$\mathbf{v}_{i,t+1}
=\mathbf{v}_{i,t}+\Delta\mathbf{v}_i$
and
$\mathbf{r}_{i,t+1}
=\mathbf{r}_{i,t}+\Delta\mathbf{x}_i$.

The rotational channel follows the same construction. Defining
$\Delta\bm{\omega}_i
=
\bm{\omega}_{i,t+1}-\bm{\omega}_{i,t}$,
and using the inverse inertia $\mathbf{I}^{-1}_i$ from Equation~\eqref{eq:m:mass_inertia}, the aggregated torque $\bm{\tau}_i$ from Equation~\eqref{eq:m:drives}, and the rotational response operators from Equation~\eqref{eq:m:operatoraggregate}, gives
\begin{equation}
\boxed{
    \left[
        \mathbf{I}_3
        +
        \frac{1}{2}\subdt\,
        \mathbf{I}^{-1}_i\Dt^{\mathrm{rot}}_i
        +
        \frac{1}{4}\subdt^2\,
        \mathbf{I}^{-1}_i\Kt^{\mathrm{rot}}_i
    \right]
    \Delta\bm{\omega}_i
    =
    \mathbf{I}^{-1}_i\bm{\tau}_i\,\subdt
    -
    \frac{1}{2}\subdt^2
    \mathbf{I}^{-1}_i
    \Kt^{\mathrm{rot}}_i
    \bm{\omega}_{i,t}.
}
\label{eq:m:solverot}
\end{equation}
The spin is then advanced as
$\bm{\omega}_{i,t+1}
=
\bm{\omega}_{i,t}
+
\Delta\bm{\omega}_i$.
The spin remains a latent angular-momentum state and receives no direct supervision.

Equations~\eqref{eq:m:solve} and~\eqref{eq:m:solverot} are solved independently at every free physical node as two $3\times3$ systems; the translational solve is illustrated in Fig.~\ref{fig:framework}e. Because the aggregated response operators are symmetric positive definite and the learned inverse mass and inverse inertia are positive isotropic operators, the coefficient matrices are symmetric positive definite with minimum eigenvalue at least one. Each local solve is therefore nonsingular for every $\subdt$ (Supplementary Information, Section~10.3).

We refer to the update as \emph{semi-implicit} because the unknown state increments enter the mechanical response being solved at each node, while the decoded drives and response operators are evaluated from the current graph state and held fixed during that solve. Unlike a classical global implicit Newmark step, \model performs one local solve per substep, without an inner Newton iteration or an assembled $3N\times3N$ system. The connection to the average-acceleration Newmark method and the scope of its classical stability properties are given in Supplementary Information, Sections~10.2--10.4.

The physical-edge interactions satisfy the pairwise balance of Section~\ref{ssec:m:fluxes}, while the virtual-edge interactions satisfy the collective balance of Equation~\eqref{eq:m:hubprojection}, before integration. Because these balanced drives are subsequently processed by different nodal response matrices, the independent local solves do not preserve total linear momentum exactly at finite $\subdt$. Supplementary Information, Section~11 derives the resulting one-substep residual and shows that its absolute magnitude is $\mathcal{O}(\subdt^2)$ under the stated boundedness assumptions.

\subsection{Training and mechanical readouts}
\label{ssec:m:training}

\model is trained solely from observed kinematic transitions, with boundary and load information included only when available. Before encoding, vector inputs are normalized using statistics computed from the training partition. The target position and velocity increments are likewise standardized using training-set statistics.

For one observed transition, the training objective is
\begin{equation}
    \mathcal{L}
    =
    \frac{1}{|\mathcal{V}|}
    \sum_{i\in\mathcal{V}}
    \left(
        \left\|
        \widehat{\Delta\mathbf{x}}_i
        -
        \Delta\mathbf{x}_i
        \right\|^2
        +
        \left\|
        \widehat{\Delta\mathbf{v}}_i
        -
        \Delta\mathbf{v}_i
        \right\|^2
    \right),
    \label{eq:m:loss}
\end{equation}
where the increments in Equation~\eqref{eq:m:loss} denote their standardised values and $\mathcal{V}$ contains only physical nodes. The hub is excluded from the objective.

No internal force, moment, stress, ground reaction force, stiffness or damping operator, finite-element matrix, or constitutive residual enters the loss. The latent spin, inverse mass, inverse inertia, momentum fluxes, and response operators are therefore learned through their role in generating the observed state transition. Training uses Adam with early stopping on the validation objective, and the checkpoint with the lowest validation loss is retained. The learning rate, weight decay, batch size, training budget, and other case-specific settings are reported in Supplementary Information, Sections~4.3--7.3; the settings shared across all four implementations are listed in Supplementary Table~1.

The quantities used to generate the transition remain accessible after training as mechanical readouts (Fig.~\ref{fig:framework}f). The decoded physical-edge interactions provide internal-force and spin-torque contributions. For the walking-biomechanics experiment, the moment about a joint centre is assembled from the decoded internal forces acting on the distal markers,
\begin{equation}
    \mathbf{M}
    =
    -\sum_{i\in\mathrm{distal}}
    \mathbf{r}_i\times\mathbf{F}_i,
    \label{eq:m:jointmoment}
\end{equation}
where $\mathbf{r}_i$ is measured from the joint center. These inferred moments are compared only at evaluation time with the inverse-dynamics references withheld from training.

The learned position- and velocity-response operators are also retained directly. They are interpreted as relative indicators of the position- and velocity-dependent response encoded by the coarse-step update, rather than as identified material or constitutive tangents. Their relation to the finite-element tangent structure is evaluated independently in Supplementary Information, Section~4.9.

\subsection{Datasets and evaluation protocol}
\label{ssec:m:data}

The clamped-beam trajectories are generated with FEniCS~\cite{fenics}, adapting the elastodynamics formulation of Bleyer~\cite{bleyer2018}. The reference trajectories use the average-acceleration Newmark parameters
$\gamma=\tfrac12$ and $\beta=\tfrac14$; Supplementary Information, Section~4.6 additionally compares the trained model with a temporally converged reference.

The human-motion benchmark uses the walking sequences of subject~35 from the CMU motion-capture database~\cite{cmumocap}, following the prediction span and data split used by \eghn~\cite{eghn}. The protein benchmark uses the driven closed--open transition ensemble of adenylate kinase protein~\cite{seyler2017adk}. The walking-biomechanics benchmark is constructed from the instrumented-treadmill recordings of van der Zee et al.~\cite{vanderzee2022}. Dataset construction, graph inputs, prediction spans, train--validation--test splits, and case-specific metrics are given in Supplementary Information, Sections~4--7.

Evaluation is autoregressive: each predicted state becomes the input to the next prediction interval. Errors are evaluated without clipping. The beam uses whole-body position error relative to beam length together with the deformation measure defined in Supplementary Information, Section~4.5; human motion and protein dynamics use position mean-squared error; and the walking-biomechanics experiment evaluates the inferred joint-moment demand against the withheld inverse-dynamics reference. The common preprocessing and rollout protocol is given in Supplementary Information, Section~3.

\subsection{Baseline implementations}
\label{ssec:m:baselines}

We compare \model with \dgn~\cite{dgn}, \gns~\cite{gns}, \mgn~\cite{mgn}, \eghn~\cite{eghn}, \egno~\cite{egno}, and \eghno, and additionally with IGNS~\cite{REF_IGNS} on the clamped beam. Comparisons are controlled within each benchmark: the models use the same trajectories, data partitions, and observed physical state, while retaining their stated architectures and training objectives unless an adaptation is required for autoregressive evaluation.

\dgn is the closest controlled comparison for the proposed update because it uses the same edge-local conserved interaction representation and rotational channel but integrates the decoded interactions explicitly. For \eghn, \eghno, and \egno, equivariant velocity heads are added because their published architectures do not return the velocity required as input to the next rollout step. IGNS is evaluated only on the beam using its released port-Hamiltonian architecture and symplectic-Euler integrator; stress supervision from its published solid-mechanics experiment is removed so that it receives the same kinematic supervision as the other beam models. The baseline objectives and shared comparison protocol are documented in Supplementary Information, Sections~2--3, while case-specific adaptations, hyperparameters, and evaluation settings are reported in Sections~4.2--7.3.

\section*{Acknowledgments}
This research was funded by the Swiss National Science Foundation (SNSF) Grant Number $200021-200461$.

\bibliographystyle{unsrt}
\bibliography{references}

\clearpage
\etocdepthtag.toc{si}
\setcounter{section}{0}
\setcounter{figure}{0}
\setcounter{table}{0}
\setcounter{equation}{0}
\renewcommand{\theHsection}{SI.\arabic{section}}
\renewcommand{\theHfigure}{SI.\arabic{figure}}
\renewcommand{\theHtable}{SI.\arabic{table}}
\renewcommand{\theHequation}{SI.\arabic{equation}}
\renewcommand{\figurename}{Supplementary Figure}
\renewcommand{\tablename}{Supplementary Table}
\renewcommand{\dgn}{\textsc{Dynami-CAL GraphNet}\xspace}

\begin{center}
  {\LARGE \textbf{Supplementary Information}}\\[0.6em]
  {\Large \textbf{Learning coarse-step dynamics and internal mechanical response with graph networks}}\\[1.2em]
  {\large Vinay Sharma$^{1}$ and Olga Fink$^{1}$}\\[0.4em]
  {\small $^{1}$Intelligent Maintenance and Operations Systems, EPFL, Lausanne, Switzerland}
\end{center}
\vspace{1em}

\etocstandardlines
\etocstandarddisplaystyle
\etocsettagdepth{main}{none}
\etocsettagdepth{si}{subsection}
\tableofcontents
\newpage

\section{Proposed Model: Training and Shared Configuration}
\label{si:model-setup}

This section specifies the objective and implementation settings used by \model in all four benchmarks. Dataset construction, graph inputs, prediction spans and case-specific hyperparameters are given in the corresponding case sections.

\subsection{Model state and update}
\label{si:model-setup:update}

\model represents each system as a graph of physical nodes and edges augmented by one virtual hub. Over one observed transition, it performs $S$ semi-implicit Newmark substeps. At each substep, the network decodes edge-resolved exchanges of linear and angular momentum together with symmetric positive-definite translational and rotational \rev{response} operators. The edge quantities are aggregated at the physical nodes and used in the Newmark update. The operator-weighted hub supplies the shared graph state required by this update without constructing a dense inverse.

The model receives no prescribed mass, damping or stiffness matrix. Its inverse mass, inverse inertia, fluxes and \rev{response} operators are learned through the state transition. Whether an external load is observed or inferred is specified for each benchmark.

\subsection{Learning objective}
\label{si:model-setup:objective}

For every system, \model is trained to predict the normalised increments of position and velocity over one prediction span,
\begin{equation}
  \mathcal{L} = \frac{1}{|\mathcal{V}|}\sum_{i \in \mathcal{V}}
  \left(
    \left\lVert \hat{\Delta \mathbf{x}}_i-\Delta \mathbf{x}_i \right\rVert^2
    +
    \left\lVert \hat{\Delta \mathbf{v}}_i-\Delta \mathbf{v}_i \right\rVert^2
  \right),
  \label{eq:si:model:loss}
\end{equation}
where each increment is standardised using statistics computed from the training split. The set $\mathcal{V}$ contains only physical nodes; the hub is excluded from the loss.

Equation~\eqref{eq:si:model:loss} is the complete training objective. No force, moment, stress or ground-reaction-force labels are used. The \rev{response} operators $\Kt_{ij}$ and $\Dt_{ij}$, their rotational counterparts, and the angular-velocity channel are also not supervised directly.

\subsection{Optimisation and shared model settings}
\label{si:model-setup:hparams}

In every case, \model is trained with the Adam optimiser and early stopping on the validation value of Equation~\eqref{eq:si:model:loss}. We retain the checkpoint with the lowest validation loss. Table~\ref{tab:si:model:shared} lists the settings shared by all four implementations; each case table reports only the quantities that differ.

\begin{table}[h]
\centering
\caption{\textbf{Settings shared by \model across the four benchmarks.}}
\label{tab:si:model:shared}
\small
\begin{tabular}{ll}
\toprule
Setting & Value \\
\midrule
Optimiser & Adam \\
MLP layers per block & 2 \\
Newmark parameters $\gamma,\beta$ & $\tfrac{1}{2},\tfrac{1}{4}$ (fixed) \\
Newmark substeps $S$ & 4 \\
Message-passing rounds per prediction span & 4 (one per substep) \\
\rev{Response} operators $\Kt_{ij},\Dt_{ij},\Kt^{\mathrm{rot}}_{ij},\Dt^{\mathrm{rot}}_{ij}$ & per-edge SPD matrices, Cholesky parameterisation \\
Independent parameters per tangent operator & 6 \\
Inverse mass $\mathbf{M}^{-1}$ and inverse inertia $\mathbf{I}^{-1}$ & learned isotropic scalars per node \\
Hub augmentation & one hub and $2N$ directed virtual edges \\
Virtual-edge attribute & \texttt{edge\_attr}$=-1$ \\
Loss & Equation~\eqref{eq:si:model:loss}, physical nodes only \\
Checkpoint selection & minimum validation loss \\
\bottomrule
\end{tabular}
\end{table}
\section{Baseline Implementations and Adaptations}
\label{si:baselines}

The baselines retain the training objectives of their original publications unless an adaptation is stated explicitly. \dgn~\cite{dgn} uses a position--velocity increment loss. The published \gns~\cite{gns} formulation predicts normalised acceleration and adds Gaussian noise to the input velocities; our implementation instead uses Equation~\eqref{eq:si:model:loss} while retaining this noise injection. \mgn~\cite{mgn} is trained on normalised acceleration. \eghn~\cite{eghn} augments a position--velocity mean-squared error with a linkage-prediction term weighted by $\lambda_{\mathrm{link}}$. \egno~\cite{egno} uses a position--velocity loss over the predicted trajectory, and \eghno~\cite{egno} adds the same linkage term. IGNS~\cite{REF_IGNS} is trained in its published form by matching trajectories over a multi-step window; the adaptation required for the beam comparison is specified in Section~\ref{si:beam:baselines}.

Section~\ref{si:evaluation} defines the data splits, checkpoint selection and rollout rules shared within each comparison. The graph, inputs, capacity settings and case-specific adaptations are reported in the case sections rather than here.

\subsection{Velocity heads for autoregressive rollout}
\label{si:baselines:velhead}

Autoregressive rollout requires each model to provide every state variable used as input at the next step. \model and \dgn provide updated positions and velocities, while \gns and \mgn predict the acceleration from which both are advanced using their respective published updates. The published \eghn, \egno and \eghno architectures predict positions but not the velocities required by our rollout state. We therefore add one velocity head to each architecture. The added head follows the construction of the corresponding position decoder and reads the representation already produced by the backbone. The message-passing backbone, pooling, temporal convolution and training objective remain unchanged. Table~\ref{tab:si:velhead} summarises the three heads.

\begin{table}[h]
\centering
\caption{\textbf{Velocity heads added to the position-only baselines.} Each head uses the same network class, hidden width and activation as the corresponding position decoder. $\mathbf{l}_{nf}$ and $\mathbf{nf}$ are the pooled and unpooled equivariant vectors of the hierarchical block; $\mathbf{l}_X$ and $\mathbf{l}_V$ are the pooled coordinate and velocity; $h$ and $\mathbf{l}_H$ are the low- and high-level invariant features; and $h_{\mathrm{out}}$ is the hidden state after temporal convolution.}
\label{tab:si:velhead}
\small
\begin{tabular}{llll}
\toprule
Model & Head & Equivariant vector inputs & Invariant scalar inputs \\
\midrule
\eghn  & \texttt{EquivariantScalarNet} or \texttt{EGMN} & $[\mathbf{l}_{nf},\mathbf{x}-\mathbf{l}_X,\mathbf{v}-\mathbf{l}_V,\mathbf{nf}]$ & $\mathrm{cat}(h,\mathbf{l}_H)$ \\
\eghno & \texttt{EquivariantScalarNet} or \texttt{EGMN} & $[\mathbf{l}_{nf},\mathbf{x}_{0r}-\mathbf{l}_X,\mathbf{v}_{0r}-\mathbf{l}_V,\mathbf{nf}]$ & $\mathrm{cat}(h_{\mathrm{low}},\mathbf{l}_H)$ \\
\egno  & \texttt{EquivariantScalarNet} & $[\mathbf{v}_{\mathrm{egnn}},\mathbf{x}_{\mathrm{out}}-\mathbf{x}_{\mathrm{init}}]$ & $h_{\mathrm{out}}$ \\
\bottomrule
\end{tabular}
\end{table}

For \eghn and \eghno, the velocity and position heads read the same equivariant vectors and invariant scalars but have independent parameters. For \egno, the velocity head reads the equivariant backbone velocity, displacement from the initial position and invariant hidden state at each predicted time. Each head follows the equivariant scalarisation--vectorisation construction of the corresponding position decoder and is therefore rotation-equivariant and translation-invariant.

The position and velocity heads are trained jointly but do not share weights. In the \texttt{EGMN} implementation, the velocity decoder receives a shallow copy of its vector-input list because the decoder appends intermediate outputs to that list at every layer.

Velocity is decoded independently because it is defined at one-frame spacing, whereas a position increment spans the complete prediction interval. Their ratio would therefore be a chord velocity over the prediction span, not the one-frame velocity consumed by the next rollout step.

\subsection{Scope of the adaptations}
\label{si:baselines:adaptations}

Only adaptations required to place a published architecture under the stated comparison protocol are introduced. Baseline processor, pooling and temporal-operator components are otherwise retained. A capacity count has different meanings across architectures: it denotes Newmark substeps for \model, processor depth for \dgn and \mgn, temporal or graph layers for \egno and IGNS, and clusters for \eghn and \eghno. These quantities are therefore reported without treating them as equivalent. Published results that were not reproduced are identified explicitly in the relevant case.

\section{Data Processing and Evaluation Protocol}
\label{si:evaluation}

This section defines the conventions shared by all models within a benchmark. It does not impose a common architecture or training objective. The objective of \model is given in Section~\ref{si:model-setup:objective}, and the objectives and adaptations of the baselines are given in Section~\ref{si:baselines}.

\subsection{Data splits and preprocessing}
\label{si:evaluation:data}

Within each benchmark, all models use the same trajectories and training, validation and test partitions. Normalisation statistics are computed from the training partition only. The construction of each partition and the inputs supplied to each architecture are stated in the corresponding case section.

Across all four systems, velocities are computed from recorded positions using the causal one-frame backward difference
\begin{equation}
  \mathbf{v}[t]=\mathbf{x}[t]-\mathbf{x}[t-1].
  \label{eq:si:evaluation:velocity}
\end{equation}
The same position and velocity records are used by every model evaluated within a case.

\subsection{Optimisation and checkpoint selection}
\label{si:evaluation:optimisation}

All trained models use the Adam optimiser and early stopping on their respective validation objectives. We evaluate the checkpoint attaining the lowest validation loss. Learning rates, weight decay, batch size, stopping patience and training budget are reported for each case.

\subsection{Autoregressive rollout}
\label{si:evaluation:rollout}

Rollouts are autoregressive: each call advances the state by one prediction span, and the predicted state becomes the input to the next call. Errors are evaluated separately at each rollout step. An aggregate over a complete rollout is reported only where it is defined explicitly in the corresponding case.

Errors are not clipped. If any seed produces a non-finite value at a given step, that step is reported as non-finite rather than averaged over the remaining seeds and is marked rather than plotted.

\subsection{Comparison protocol}
\label{si:evaluation:comparison}

Comparisons are controlled within each benchmark, not across benchmarks. Models use the same trajectories, partitions and observed physical state, while retaining their stated objectives and update rules. Any additional input required by a published architecture, any withheld mechanical label and every implementation adaptation are identified in the corresponding case section. Seed sets, error definitions and whether a result is reproduced or quoted from the original publication are also stated there.

\section{Elastodynamics of a Clamped Beam}
\label{si:beam}

\subsection{Case description}
\label{si:beam:desc}

The system is a three-dimensional cantilever beam of length $L$ and rectangular cross-section $W \times D$, clamped at one end. A transverse load of magnitude $F$ is applied from rest and removed at a cut-off time $T_c$, after which the beam vibrates freely and its oscillation decays through Rayleigh damping. Reference trajectories are produced with the finite-element solver FEniCS~\cite{fenics}, adapting the elastodynamics demonstration of Bleyer~\cite{bleyer2018}, on an unstructured tetrahedral mesh whose density is set by a resolution parameter \texttt{res}, the number of elements per unit length.

The material is linear elastic and nondimensional, with Young's modulus $E = 1000$, Poisson ratio $\nu = 0.3$ and density $\rho = 1$. Damping is of Rayleigh type, $\mathbf{C} = \eta_m \mathbf{M} + \eta_k \mathbf{K}$ with $\eta_m = \eta_k = 0.01$. Each trajectory spans $T = 10\,$ in $100$ steps, so one observed transition is $\dt = 0.1\,$. The finite-element trajectories are advanced with the average-acceleration Newmark scheme, $\gamma = \tfrac{1}{2}$ and $\beta = \tfrac{1}{4}$. These parameters are fixed, not fitted. Section~\ref{si:beam:converged} separately evaluates the trained model against a temporally converged reference to ensure that its accuracy is not an artefact of sharing the coarse reference discretisation.

The elastic wave speed is $c = \sqrt{E/\rho} = 31.6$, so a wave crosses the base beam in $L/c = 0.032\,$ and one outer step spans roughly three wave transits. Measured against the largest natural frequency of the model graph, the outer step exceeds the explicit stability limit $\subdt\,\omega \le 2$ by more than an order of magnitude at the outer step, and by 5.5–11× even when resolved into the four Newmark sub-steps; the spectral analysis, with the exact factors, is given in Section~\ref{si:beam:stiffness}. A coarser element-size estimate, $\subdt_{\mathrm{crit}} \approx h/c$, gives the same order of magnitude.

\textbf{Observed inputs and graph.} Every model receives the current nodal positions and velocities, the mesh graph, the free/clamped boundary mask and the current applied load. No material property, constitutive parameter, finite-element matrix, stress, internal force or moment is supplied. Nodes are the mesh vertices and physical edges follow the mesh connectivity. \model augments this graph with one hub joined to all $N$ physical nodes by $2N$ directed virtual edges, identified by an edge attribute of $-1$; the baselines retain their native hub-free topology. Graphs are preprocessed once and are not rebuilt during training.

\begin{table}[h]
\centering
\caption{\textbf{Beam representation supplied to \model.} $N$ is the number of mesh vertices and depends on the configuration. Baselines receive the same observed physical state on the hub-free graph.}
\label{tab:si:beam:rep}
\small
\begin{tabular}{ll}
\toprule
Property & Value \\
\midrule
Nodes & mesh vertices \\
Physical edges & mesh connectivity \\
Hub edges & $2N$ virtual edges to one hub \\
Node features & position, velocity, free/clamped type, applied load \\
Edge features & scalar attribute ($-1$ on hub edges) \\
Edge-local frame & yes (\dgn) \\
Global frame & no \\
External force & given per node (applied load) \\
Seeds & $\{42\}$, $n = 1$ \\
\bottomrule
\end{tabular}
\end{table}

\textbf{Splits.} The training distribution is the Cartesian product $L \in \{1.0, 1.5, 2.0\}$, $W \in \{0.5, 1.0\}$, $D \in \{0.5, 1.0\}$, $F \in \{1.5, 2.0, 2.5\}$, $T_c \in \{2.0, 2.5, 3.0\}\,$ at $\texttt{res} = 4$, giving $108$ configurations and one trajectory each, divided into training, validation and test partitions. Nine further configurations are held out entirely and never seen in any form; they vary the load amplitude, the cut-off time, the cross-section, the beam length and the mesh resolution, one axis at a time.

\textbf{Metric.} At each rollout step, the whole-body error is the root-mean-square position error over the physical nodes, expressed as a percentage of the beam length $L$. Where one value summarises a trajectory, we report its mean over the $95$ autoregressive steps. The deformation ratio, predicted over reference total displacement, measures over- or under-deformation. Section~\ref{si:beam:wholebody} explains why this whole-body measure is preferred to a tip-only error. All results use one seed, $42$, so no dispersion is reported.

\subsection{Evaluated baselines}
\label{si:beam:baselines}

We compare \model with \dgn, \mgn, \eghn, \eghno, \egno and IGNS. All receive the observed inputs listed above, and none receives the finite-element mass, damping or stiffness matrices. The baselines use the physical mesh graph without the hub. \eghn, \eghno and \egno use the velocity heads of Section~\ref{si:baselines:velhead}. \dgn is evaluated with eight and twelve message-passing rounds, and \mgn with four, eight and twelve processor rounds, to test whether additional local propagation closes the gap to the hub-mediated update. \mgn has \rev{no rotational channel} and is included only in the rollout comparison. No global frame is used because the applied load is observed.

IGNS is evaluated only in this case. Its published solid-mechanics experiment uses multi-step trajectory matching and stress supervision. Here it receives no stress target and is trained on Equation~\eqref{eq:si:model:loss}, so that the comparison tests the integration mechanism under the same one-step kinematic supervision as \model. We use the released port-Hamiltonian architecture and the symplectic-Euler integrator specified in the paper. Because the input edge features are standardised and may be signed, the released positive-distance adjacency weighting is replaced by symmetric degree normalisation; distance remains in the edge features. All other inputs, normalisation, optimisation and training budgets match the hub-free beam protocol. The number of internal integration steps is swept over $\{1,4,12,24\}$; four matches the substep budget of \model.

For \mgn, we corrected the output normalisation before evaluation; without this correction, its initial loss was of order $10^5$ and rollout drift began immediately. All reported \mgn results use the corrected implementation.

\subsection{Hyperparameters}
\label{si:beam:hparams}

Settings for \model not listed in Table~\ref{tab:si:beam} follow Table~\ref{tab:si:model:shared}.

\begin{table}[h]
\centering
\caption{\textbf{Beam elastodynamics: hyperparameters.} Capacity parameters have architecture-specific meanings, as explained in Section~\ref{si:baselines:adaptations}.}
\label{tab:si:beam}
\small
\begin{tabular}{lccccc}
\toprule
Model & latent $nf$ & capacity parameter & batch & learning rate & weight decay \\
\midrule
\model         & 64 & $S = 4$ sub-steps        & 32 & $5\times10^{-4}$ & $10^{-10}$ \\
\dgn           & 64 & $8$ and $12$ messages     & 32 & $5\times10^{-4}$ & $10^{-10}$ \\
\mgn           & 64 & $4$, $8$, $12$ messages   & 64 & $5\times10^{-4}$ & $10^{-10}$ \\
\eghn          & 64 & $15$ clusters             & 32 & $5\times10^{-4}$ & $10^{-10}$ \\
\eghno         & 64 & $5$ clusters              & 32 & $5\times10^{-4}$ & $10^{-10}$ \\
\egno          & 64 & $12$ layers               & 32 & $5\times10^{-4}$ & $10^{-10}$ \\
\rev{IGNS}     & \rev{published} & \rev{$1$, $4$, $12$, $24$ internal steps} & \rev{$64$} & \rev{$5\times10^{-4}$} & \rev{$10^{-10}$} \\
\bottomrule
\end{tabular}
\end{table}

All models train for at most $2000$ epochs with early stopping, evaluated every second epoch, and the checkpoint minimising the validation loss is retained. Seed $42$ throughout.

\subsection{Additional results}
\label{si:beam:additional}

\begin{figure}[H]
  \centering
  \includegraphics[width=0.52\textwidth]{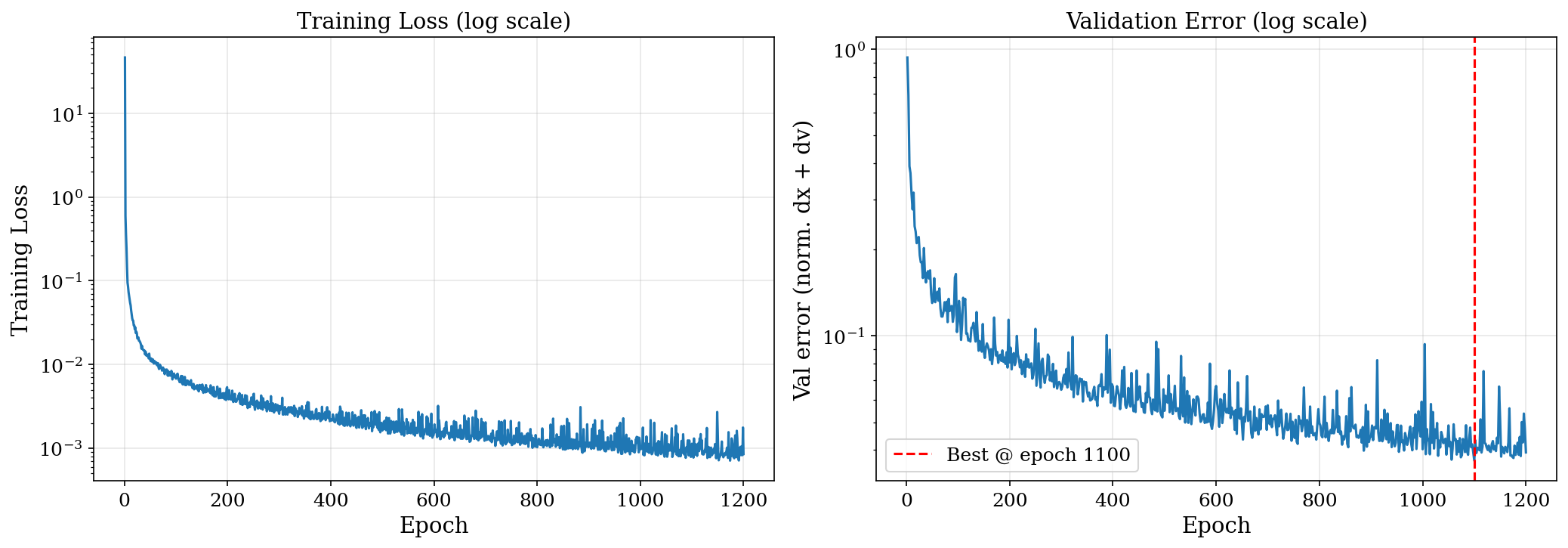}
  \caption{\textbf{Training and validation loss for \model on the beam.} Both terms of
  Equation~\eqref{eq:si:model:loss} are shown. Early stopping selects the checkpoint at the minimum of
  the validation curve. Seed 42.}
  \label{fig:si:beam:training}
\end{figure}

\begin{figure}[H]
  \centering
  \includegraphics[width=0.44\textwidth]{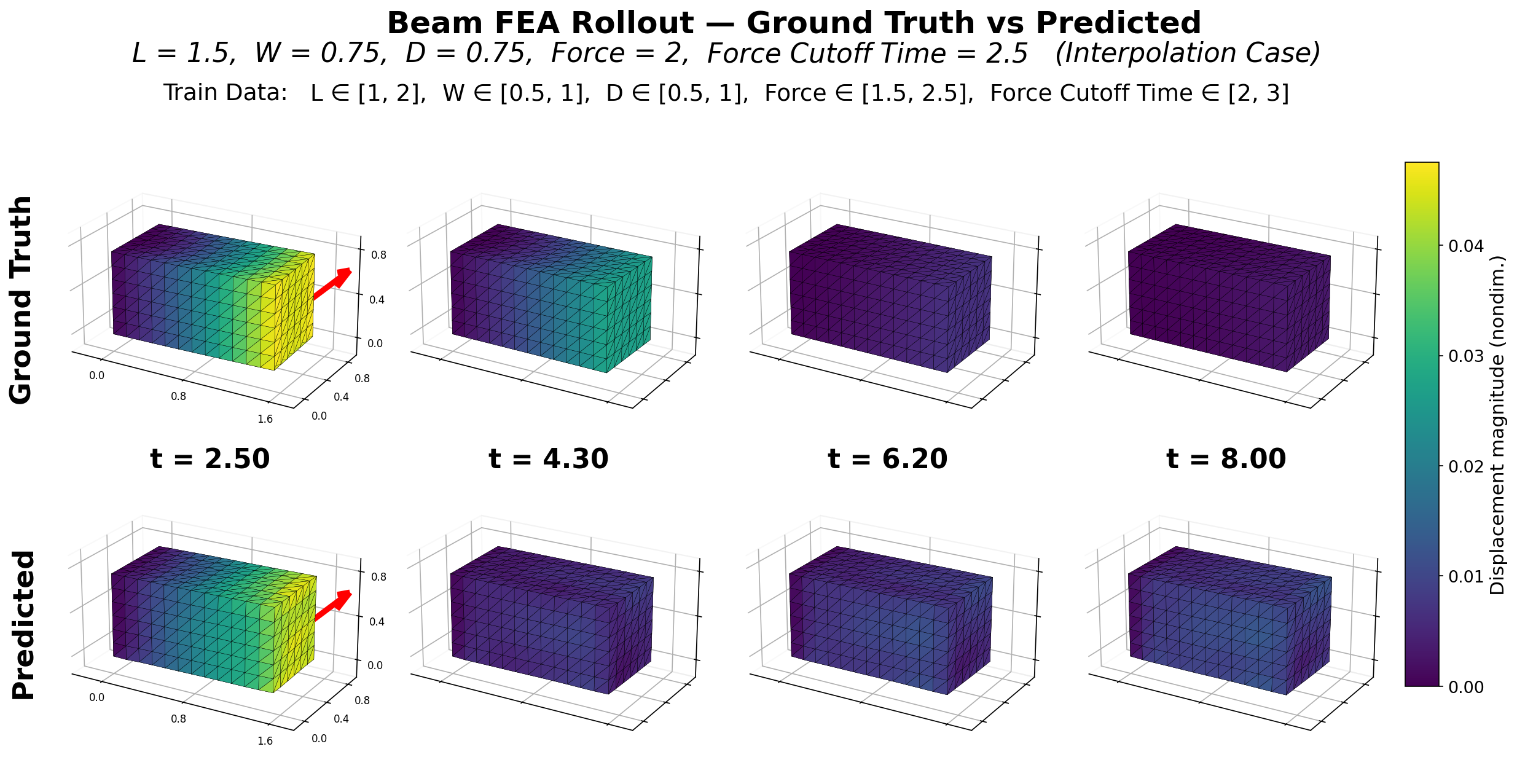}
  \hfill
  \includegraphics[width=0.44\textwidth]{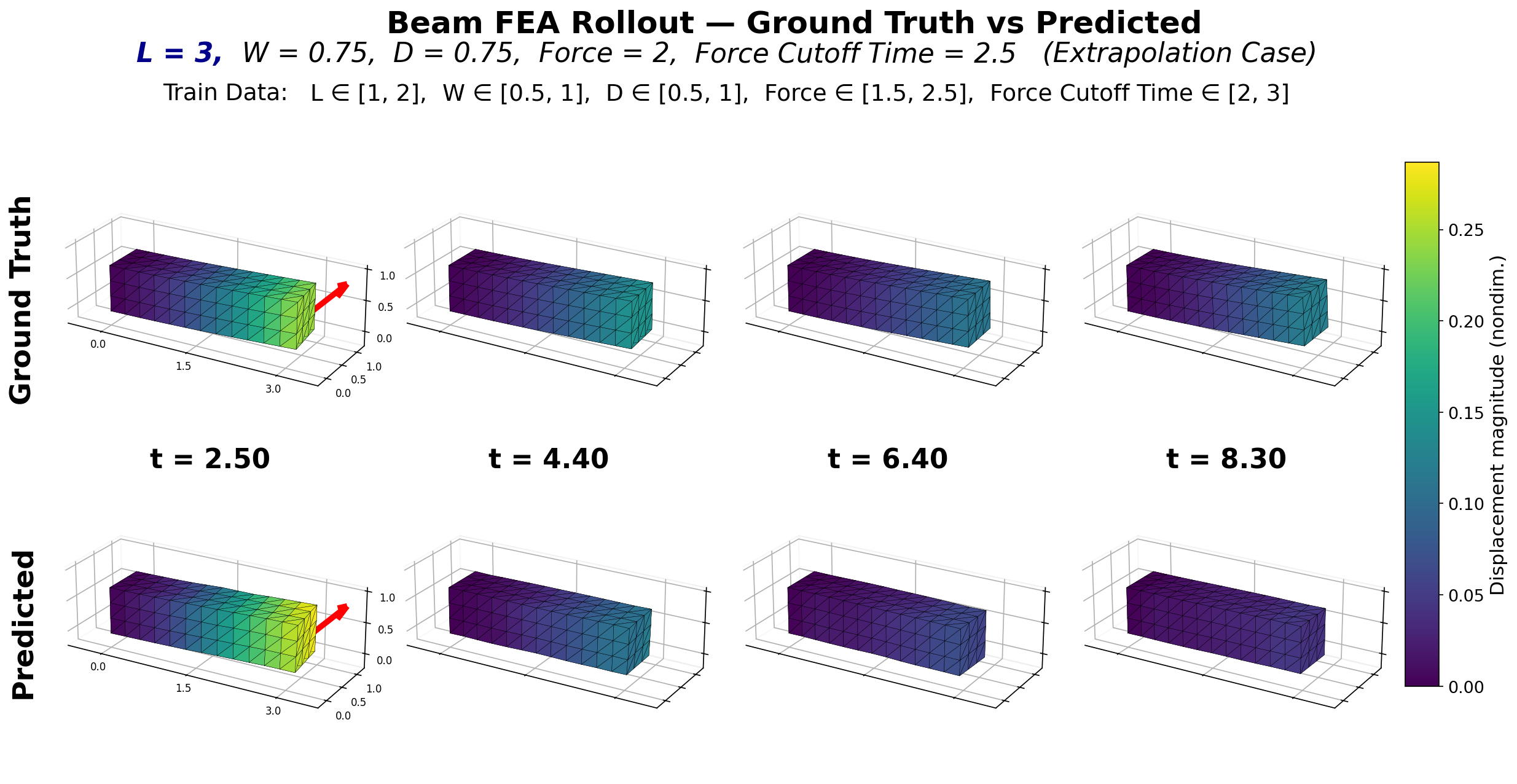}
  \\[4pt]
  \includegraphics[width=0.44\textwidth]{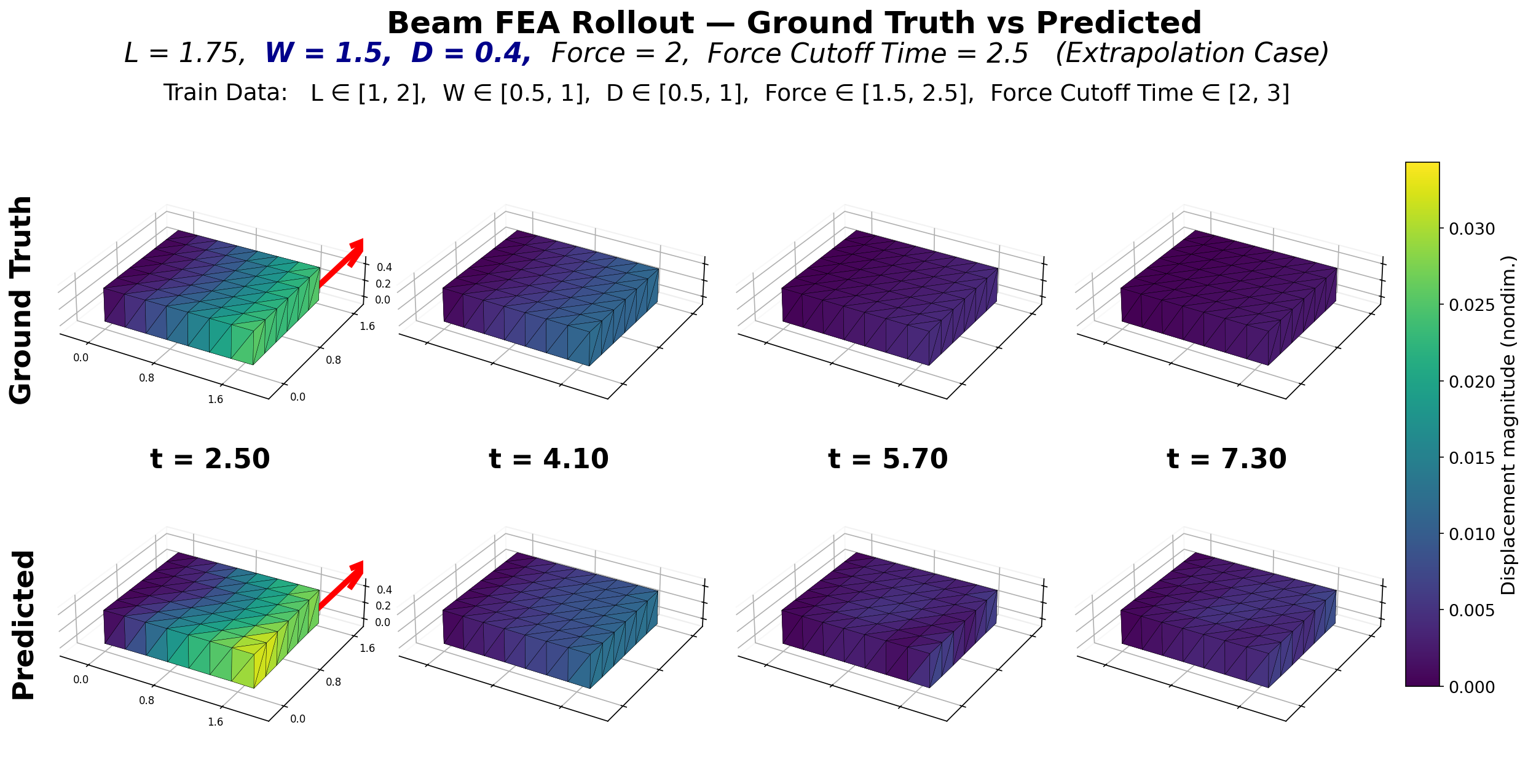}
  \hfill
  \includegraphics[width=0.44\textwidth]{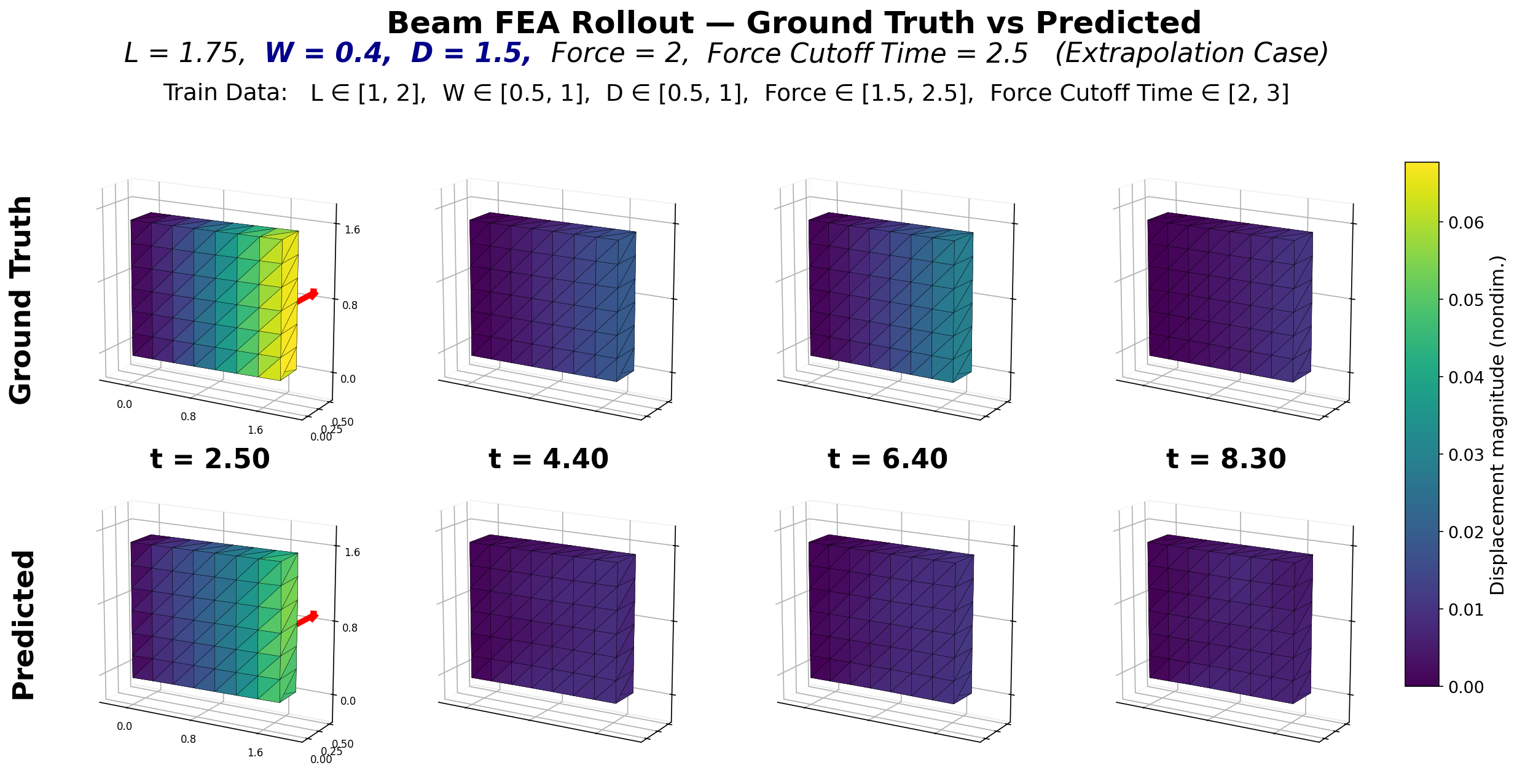}
  \caption{\textbf{Rollout of \model on four held-out configurations.} Top left, the refined mesh
  ($\texttt{res} = 8$), the only extrapolation that leaves the training envelope of the stiffness
  spectral radius. Top right, the beam extrapolated to three times the training length, where the
  rank-one hub begins to be an inadequate surrogate for the long-range coupling. Bottom, the two
  cross-sections outside the training product, $W = 1.5$, $D = 0.4$ on the left and $W = 0.4$,
  $D = 1.5$ on the right, which exchange the bending stiffness of the two transverse axes. The
  remaining five configurations behave comparably and are not reproduced here.}
  \label{fig:si:beam:rollout}
\end{figure}


\subsection{Whole-body error and the role of the semi-implicit solve}
\label{si:beam:wholebody}

The metric of a single point like tip can under-report the errors since a bounded tip motion can coexist with an interior that over-deforms: a mesh that swells or buckles between the clamp and the tip can still return a small tip error. We therefore also report the \emph{whole-body} error, the root-mean-square position error over all body nodes as a percentage of the beam length, together with a deformation ratio, the predicted total displacement divided by the reference total displacement, whose departure from unity measures over- or under-deformation. Both are accumulated over the same $95$ steps and evaluated on the same seed.

Table~\ref{tab:si:beam:decomp} decomposes the model into the ingredients that distinguish it from \dgn: the operator-weighted hub, the semi-implicit solve, and the stiffness term $\beta$ of the Newmark left-hand side. Adding the hub to \dgn---with a purely explicit per-node update ($\mathbf{A}_i=\mathbf{I}$) and only four sub-steps in place of twelve---reduces the mean whole-body error from \rev{$3.00\%$} to \rev{$1.26\%$}, a factor of \rev{$2.4$}: the hub is the dominant ingredient, improving accuracy even as the sub-step count is reduced. Restoring the semi-implicit solve then refines the result, to \rev{$0.70\%$} at $\beta=0$ and \rev{$0.54\%$} with the stiffness term, further factors of \rev{$1.8$} and \rev{$1.3$}. The gap to \dgn is widest on the refined mesh, where \dgn reaches $15$--$16\%$ whole-body error (for each of the two refined-mesh configurations, all nodes, mean over 95 steps) and over-deforms the mesh by factors of $29$--$85$, against $0.5$--$1.2\%$ and a deformation ratio near two for \model.

\begin{table}[h]
\centering
\caption{\rev{\textbf{Whole-body error decomposition (mean over $21$ configurations, \% of $L$).} The hub
is the dominant ingredient; the semi-implicit solve and the stiffness term $\beta$ are refinements. Cells
are cumulative: each row adds one ingredient to the row above. The first hub row also reduces the
sub-step count from $12$ to $4$, so its factor bundles the two changes. Seed 42.}}
\label{tab:si:beam:decomp}
\small
\begin{tabular}{lccc}
\toprule
Configuration & Sub-steps & Whole-body error (\%) & factor \\
\midrule
\dgn\ (hub off, explicit)                          & \rev{$12$} & \rev{$3.00$} & --- \\
\quad $+$ operator-weighted hub (explicit, $\mathbf{A}_i=\mathbf{I}$) & \rev{$4$}  & \rev{$1.26$} & \rev{$2.4\times$} \\
\quad $+$ semi-implicit solve ($\beta=0$)          & \rev{$4$}  & \rev{$0.70$} & \rev{$1.8\times$} \\
\quad $+$ stiffness term $\beta$ (\model)           & \rev{$4$}  & \rev{$0.54$} & \rev{$1.3\times$} \\
\bottomrule
\end{tabular}
\end{table}

\subsection{Accuracy against a temporally converged reference}
\label{si:beam:converged}

Because the finite-element reference uses the same Newmark parameters and the same step as \model, its accuracy could in principle reflect a shared discretisation rather than the physics. To rule this out, we regenerate the reference with the same solver at a $100$ times finer step, $\subdt = 10^{-3}\,$, and subsample it back to the $0.1\,$ grid; halving this step again changes the whole-body trajectory by $0.005\%$, so $\subdt = 10^{-3}\,$ is the converged solution. The mesh is deterministic, so the fine and coarse body nodes coincide exactly. Table~\ref{tab:si:beam:converged} evaluates the unchanged seed-42 checkpoint against this converged reference on eight representative configurations spanning the training distribution, both refined meshes, a longer beam and a cut-off-time extrapolation.

\model\ reproduces the converged solution to between $0.19\%$ and $2.06\%$ whole-body error, essentially identical to its error against the coarse reference (means $0.72\%$ versus $0.67\%$). The reference's own coarse-step discretisation error is only $0.09$--$0.58\%$, so the comparison is not integrator-matching; the model is accurate to the physics, not to a co-discretised reference. On the stiffest and longest configurations the model is marginally closer to the converged solution than the single-step coarse reference is, consistent with its four sub-steps of $\subdt = 0.025\,$ resolving the interval more finely than a single $0.1\,$ step.

\begin{table}[h]
\centering
\caption{\textbf{Whole-body error against a temporally converged reference}
($\subdt=10^{-3}\,$, mean over 95 steps, \% of $L$).
The first two error columns compare \model with the indicated reference;
the final column reports the coarse-reference discretisation error. Seed 42.}
\label{tab:si:beam:converged}
\small
\begin{tabular*}{\linewidth}{@{\extracolsep{\fill}}lccc@{}}
\toprule
Configuration
& \shortstack{Model error\\coarse reference}
& \shortstack{Model error\\converged reference}
& \shortstack{Reference\\discretisation error} \\
\midrule
$L1.0\ W0.5\ D1.0$ res4 (in-distribution) & $0.14$ & $0.19$ & $0.09$ \\
$L1.5\ W1.0\ D1.0$ res4 (in-distribution) & $0.08$ & $0.24$ & $0.19$ \\
$L2.0\ W0.5\ D0.5$ res4 (in-distribution) & $0.70$ & $0.70$ & $0.49$ \\
$L1.75\ T_c4$ res4 (extrapolation)      & $0.38$ & $0.51$ & $0.38$ \\
$L1.75\ W1.5\ D0.4$ res4 (cross-section)  & $0.21$ & $0.26$ & $0.14$ \\
$L1.5\ W0.75\ D0.75$ \textbf{res8}        & $0.52$ & $0.52$ & $0.35$ \\
$L2.0\ W0.75\ D0.75$ \textbf{res8}        & $1.21$ & $1.24$ & $0.58$ \\
$L3.0\ W0.75\ D0.75$ res4 (long)          & $2.14$ & $2.06$ & $0.47$ \\
\midrule
Mean                                       & $0.67$ & $0.72$ & $0.34$ \\
\bottomrule
\end{tabular*}
\end{table}

\subsection{Spectral radius and the IGNS baseline}
\label{si:beam:stiffness}

The outer step of $\dt = 0.1\,$ lies far beyond the reach of any explicit stability limit. Linearizing the reference dynamics about its trajectory gives a largest natural frequency $\omega_{\max} \approx 442$ on the training mesh and $\approx 876$ on the refined mesh; at the outer step $\dt = 0.1\,$ the product $\dt\,\omega_{\max}$ is $44$ and $88$, that is $22$ and $44$ times the explicit stability limit $\dt\,\omega \le 2$, and even resolved into the four sub-steps of \model it remains $5.5$ and $11$ times that limit. The corresponding symplectic-Euler update therefore amplifies the stiffest mode by a factor of $109$ per sub-step on the training mesh and $455$ on the refined mesh, whereas the corresponding classical undamped average-acceleration Newmark update has unit amplification (Table~\ref{tab:si:beam:spectral}, and Section~\ref{si:newmark}, \rev{Equation~\eqref{eq:si:linear-stability}}). Mesh refinement is the decisive axis because it is the only held-out change that raises $\omega_{\max}$ beyond the training range.

\begin{table}[h]
\centering
\caption{\textbf{Spectral stability of the beam operator.} $\omega_{\max}$ is the largest natural
frequency of the linearised system; the explicit limit is $\dt\,\omega \le 2$. The Newmark
amplification is unity at every step size; the symplectic-Euler amplification is per sub-step
($S=4$).}
\label{tab:si:beam:spectral}
\small
\begin{tabular}{lcc}
\toprule
Quantity & training mesh (res4) & refined mesh (res8) \\
\midrule
$\omega_{\max}$                              & $442$   & $876$ \\
$\dt\,\omega_{\max}$ (outer step)            & $44$    & $88$ \\
\quad multiple of explicit limit            & $22\times$ & $44\times$ \\
symplectic-Euler amplification / sub-step   & $109\times$ & $455\times$ \\
Newmark amplification / step                 & $1.0$   & $1.0$ \\
\bottomrule
\end{tabular}
\end{table}

This prediction is borne out by IGNS, evaluated under the protocol of Section~\ref{si:beam:baselines} as the method-specific analogue of the identity-matrix ablation. At its native budget of roughly one internal step per frame the rollout diverges to $10^{6}$--$10^{7}\%$ on every configuration. At four internal steps, the substep budget of \model, it fails on all configurations. Even at $24$ internal steps, six times that budget, it reaches $231$--$338\%$ whole-body error on the two refined meshes, against $0.5$--$1.5\%$ for \model, and its per-case error on the refined meshes does not decrease as the budget grows (Table~\ref{tab:si:beam:igns}). Its single-step validation loss meanwhile improves monotonically with the budget, from $0.21$ at one step to $0.008$ at $24$, so accurate fitting of one transition does not imply a stable repeated application. Increasing the number of explicit internal steps therefore does not reproduce the coarse-step behaviour of the semi-implicit response.

\begin{table}[h]
\centering
\caption{\textbf{IGNS rollout error under an increasing internal-step budget} (whole-body, final
step, \% of $L$). Four internal steps matches the sub-step budget of \model; the refined-mesh
error does not fall as the budget grows. \model\ (four sub-steps) is shown for reference.}
\label{tab:si:beam:igns}
\small
\begin{tabular}{lccc}
\toprule
Internal steps & compliant res4 & long beams & refined res8 \\
\midrule
$1$ (native)          & $10^{6}$--$10^{7}$ & $10^{6}$ & $10^{7}$ \\
$4$ ($=$ \model)      & $11$--$50$ & $14$--$20$ & $78$--$123$ \\
$12$                  & $5$--$15$  & $61$--$78$ & $100$--$122$ \\
$24$ ($6\times$)      & $1$--$7$   & $23$--$25$ & $231$--$338$ \\
\midrule
\model\ ($S=4$)       & $<1$ & $\sim 2$ & $0.5$--$1.5$ \\
\bottomrule
\end{tabular}
\end{table}

The trained one-step maps confirm this mechanism empirically. Off the reference trajectory the amplification factor of \model is near unity, between $0.97$ and $1.08$ across configurations, a few percent short of the physical Rayleigh contraction rather than expansive, so its rollout stays bounded and non-monotone rather than growing. \dgn's map is likewise near-neutral ($0.999$--$1.006$, including the refined mesh), which is why it drifts rather than diverges; \eghn's map instead amplifies, by $1.1$--$1.6\times$ on the training mesh and by roughly $11\times$ on the refined mesh, which is why it reaches a non-finite value within tens of steps. \model is accurate on the trajectory and near-neutral off it, and that is what keeps its rollout bounded.

Consistent with the absence of numerical dissipation in the linear analysis, the model's free-vibration ring-down matches the reference's physical Rayleigh decay to within about ten percent in-distribution (decay ratios $0.92$--$1.09$) and shifts the vibration frequency by under one percent, rather than damping more strongly as an over-dissipative scheme would. On the refined mesh this breaks down: the model over-damps by roughly $2.5\times$ and mis-predicts the frequency by $-27\%$, a genuine extrapolation limitation of the learned operators rather than of the integrator.

\subsection{\rev{Modal structure recovered from the rollout}}
\label{si:beam:modal}

\rev{The spectral analysis of Section~\ref{si:beam:stiffness} concerns the integrator alone: the average-acceleration Newmark update neither amplifies nor damps a linear mode, whereas the explicit update amplifies the stiff modes it cannot resolve. A model can fit each single transition accurately and still reproduce the wrong frequency, because frequency error accumulates only over the rollout. We therefore read the free-vibration content directly from the autoregressive trajectory, with no frequency or mode supervision.}

\begin{figure}[htbp]
  \centering
  \includegraphics[width=\textwidth]{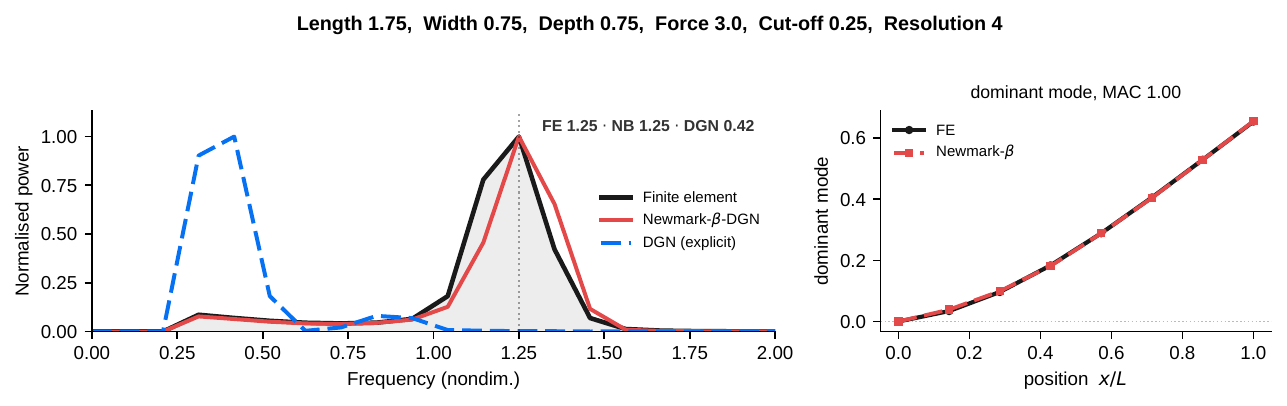}\\[4pt]
  \includegraphics[width=\textwidth]{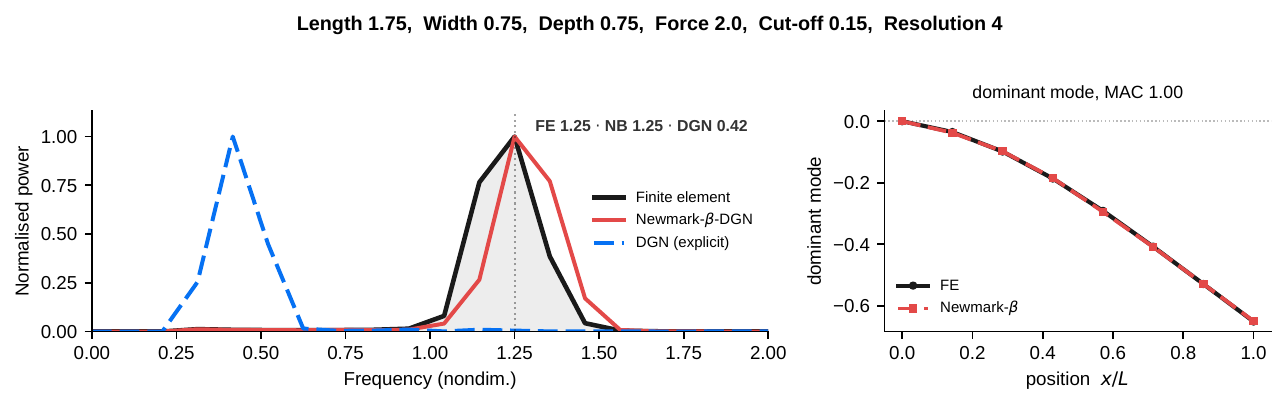}\\[4pt]
  \includegraphics[width=\textwidth]{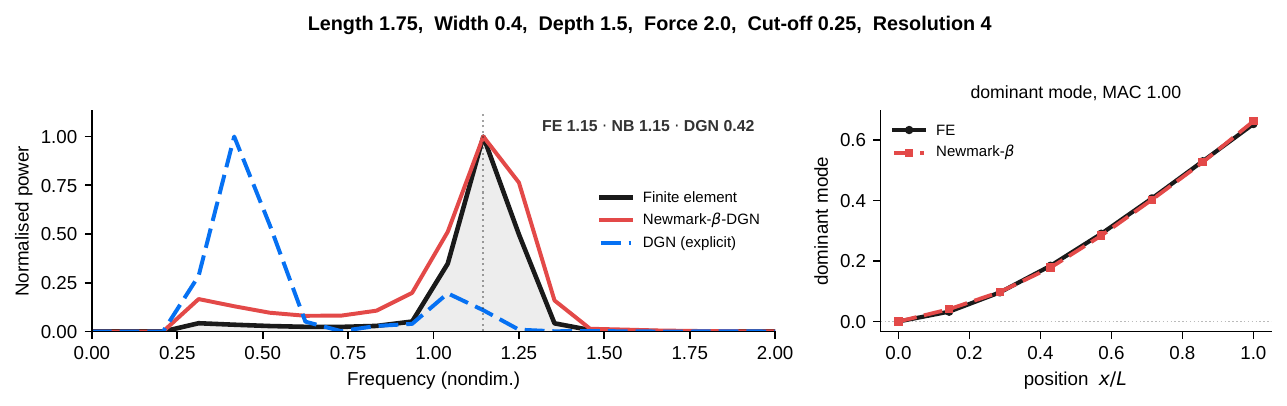}
  \caption{\rev{\textbf{Fundamental frequency and dominant mode recovered from the rollout,} on three
  held-out configurations (top to bottom: a load-amplitude, a load-duration and a cross-section
  extrapolation, identified in each title). Left, the natural-frequency spectrum of the tip transverse
  displacement, finite element (black), \model\ (red), \dgn\ (blue dashed), with the fundamental
  frequencies annotated; right, the dominant proper-orthogonal mode shape, finite element against \model,
  with its MAC. \model\ matches the finite-element fundamental and mode; \dgn\ peaks at a spurious lower
  frequency. No frequency or mode is supervised.}}
  \label{fig:si:beam:modal}
\end{figure}

\rev{For each held-out configuration, the tip transverse displacement over the $95$ rollout steps is detrended to remove the slow loading envelope and its power spectrum is taken; the peak is the fundamental frequency. Independently, a proper-orthogonal decomposition of the transverse displacement along the beam returns the dominant spatial mode shape, compared with the finite-element mode by the Modal Assurance Criterion (MAC), which equals one for identical shapes. Figure~\ref{fig:si:beam:modal} shows both for three representative configurations and Table~\ref{tab:si:beam:modal} reports them for every configuration.}

\begin{table}[h]
\centering
\caption{\rev{\textbf{Natural frequency and dominant mode recovered from the rollout, every
configuration.} FE is the finite-element fundamental; the \model\ value is read from its rollout tip
spectrum; MAC compares the dominant \model\ mode shape with the finite-element mode. A frequency counts as
recovered when it falls in the same spectral bin ($\approx 0.1\,$Hz at $\texttt{res}=4$). Seed 42.}}
\label{tab:si:beam:modal}
\rev{\footnotesize
\setlength{\tabcolsep}{3pt}
\begin{tabular}{@{}llccp{3.2cm}@{}}
\toprule
Configuration & Extrapolation axis & FE $\to$ \model\ (Hz) & MAC & Modal structure recovered \\
\midrule
$L1.0\,W0.5\,D1.0\,F2.0\,T_c3.0$              & in-distribution      & $2.29 \to 2.29$ & $1.00$ & yes \\
$L1.75\,W0.75\,D0.75\,F3.0\,T_c2.5$          & load amplitude       & $1.25 \to 1.25$ & $1.00$ & yes \\
$L1.75\,W0.75\,D0.75\,F2.0\,T_c1.5$          & load duration        & $1.25 \to 1.25$ & $1.00$ & yes \\
$L1.75\,W0.75\,D0.75\,F2.0\,T_c4.0$           & load duration        & $1.25 \to 1.25$ & $1.00$ & yes \\
$L1.75\,W0.4\,D1.5\,F2.0\,T_c2.5$            & cross-section        & $1.15 \to 1.15$ & $1.00$ & yes \\
$L1.75\,W1.5\,D0.4\,F2.0\,T_c2.5$            & cross-section        & $1.67 \to 1.56$ & $1.00$ & mode; frequency within one bin \\
$L3.0\,W0.75\,D0.75\,F2.0\,T_c2.5$           & length ($3\times$)   & $0.52 \to 0.62$ & $1.00$ & mode; frequency within one bin \\
$L3.5\,W0.75\,D0.75\,F2.0\,T_c2.5$           & length ($3.5\times$) & $0.31 \to 0.52$ & $1.00$ & mode; frequency over-predicted \\
$L1.5\,W0.75\,D0.75\,F2.0\,T_c2.5\ \mathbf{res8}$ & mesh refinement & $1.46 \to 0.94$ & $0.97$ & mode; frequency under-predicted \\
$L2.0\,W0.75\,D0.75\,F2.0\,T_c2.5\ \mathbf{res8}$ & mesh refinement & $0.94 \to 0.52$ & $0.98$ & mode; frequency under-predicted \\
\bottomrule
\end{tabular}}
\end{table}

\rev{The dominant mode shape is recovered on every configuration, at $\mathrm{MAC}\ge 0.97$. The fundamental
frequency of \model\ matches the finite-element value to the spectral resolution of the record on the
loading and thin-cross-section extrapolations, and departs from it only on the length and mesh-refinement
extrapolations, the axes that raise $\omega_{\max}$ beyond the training range (Section~\ref{si:beam:stiffness}).
The explicit \dgn\ rollout never recovers the fundamental, peaking at a spurious $0.3$--$0.4\,$Hz on every
configuration, so a bounded rollout does not imply a correct spectrum. This recovery is the empirical
counterpart of the integrator's unit amplification; it is inherited from the linear analysis and is not
proved for the learned nonlinear rollout.}

\subsection{Learned operators and the finite-element tangent}
\label{si:beam:operators}

We ask how much of the true finite-element tangent the decoded operators recover, since a readout is trustworthy only to the extent that its content is identifiable from the observed motion. At each free body node, the position-response operator $\Kt_i = \sum_j \Kt_{ij}$ and velocity-response operator $\Dt_i$, the $3\times3$ symmetric positive-definite blocks used by the learned update, are compared with the corresponding finite-element matrices: the stiffness $\mathbf{K}_{ii}$, which is state-independent for the linear beam, and the Rayleigh damping $\mathbf{C}_{ii} = \eta_k\mathbf{K}_{ii} + \eta_m\mathbf{M}_{ii}$. The comparison is made at three stages of the trajectory (loaded, free vibration and rest) for three held-out configurations.

\begin{figure}[h!]
  \centering
  \includegraphics[width=0.86\textwidth]{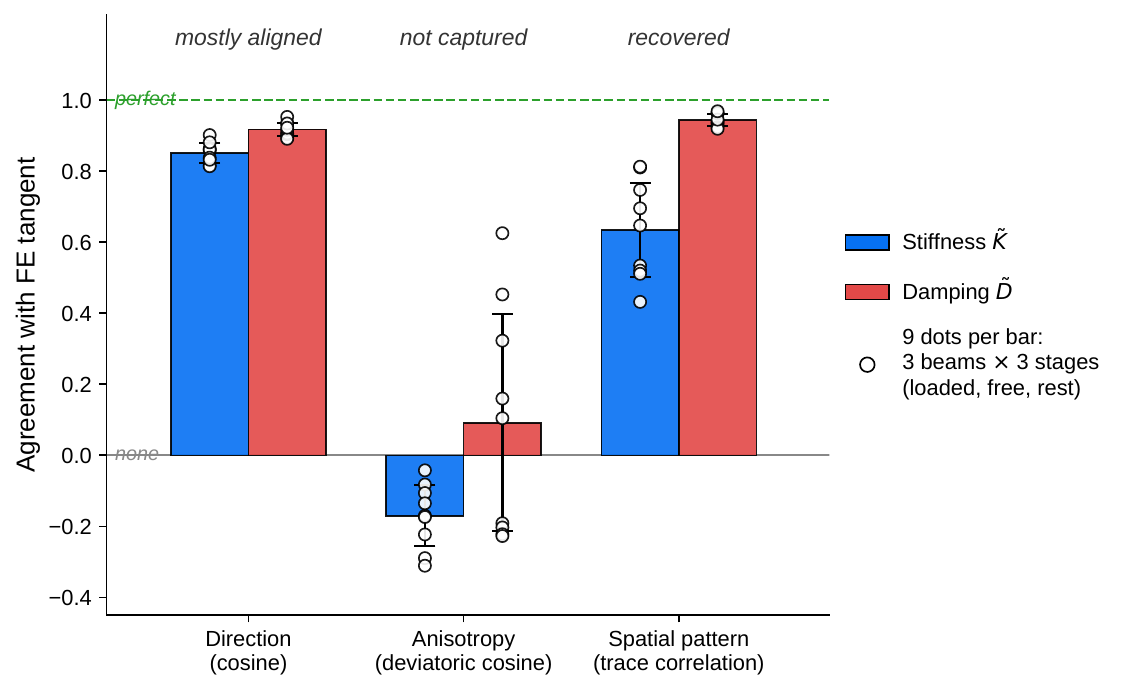}
  \caption{\textbf{The learned response operators recover part of the finite-element tangent
  structure, but not its anisotropy or magnitude.} Agreement of the decoded per-node
  position-response operator $\Kt_i$ (blue) and velocity-response operator $\Dt_i$ (red) with the corresponding finite-element matrices under three
  scale-separated measures: \emph{direction} is the Frobenius cosine between the operators;
  \emph{anisotropy} is the cosine between their trace-free (deviatoric) parts; \emph{spatial pattern}
  is the correlation of their traces across nodes. Each dot is one of three held-out configurations at
  one of three trajectory stages (loaded, free vibration, rest); bars are the mean and error bars one
  standard deviation. The operators align in direction (cosine $\approx0.85$ for stiffness, $0.92$ for
  damping) and reproduce where the material is stiff or soft (trace correlation $\approx0.63$ for
  stiffness, $0.94$ for damping), but do not recover the directional anisotropy (deviatoric cosine
  $\approx0$) and are off in absolute scale by $2$--$9\times$ for stiffness and roughly $10^{3}\times$
  for damping. They are therefore operational operators, usable as relative readouts, rather than
  identified material tangents.}
  \label{fig:si:tangent}
\end{figure}

A symmetric operator splits orthogonally, in the Frobenius inner product $\langle\mathbf{A},\mathbf{B}\rangle_F = \sum_{ab}A_{ab}B_{ab}$, into an isotropic part $\tfrac13(\operatorname{tr}\mathbf{M})\mathbf{I}$ and a trace-free deviatoric part, so that $\lVert\mathbf{M}\rVert_F^2 = \lVert\mathbf{M}_{\mathrm{iso}}\rVert_F^2 + \lVert\mathbf{M}_{\mathrm{dev}}\rVert_F^2$; physically the isotropic part is the overall stiffness magnitude, which is direction-independent, and the deviatoric part is the anisotropy, that is, which directions are stiffer than others. We separate the two with three measures (Fig.~\ref{fig:si:tangent}). The \emph{direction} is the Frobenius cosine between the decoded and exact operators, $\cos_i = \langle\Kt_i,\mathbf{K}_{ii}\rangle_F / (\lVert\Kt_i\rVert_F\,\lVert\mathbf{K}_{ii}\rVert_F)$: it is scale-invariant, equals one when the two are proportional, and because it is dominated by the large isotropic part it reports the mean direction and not the anisotropy. The \emph{anisotropy} is the same cosine taken on the deviatoric parts $\operatorname{dev}(\mathbf{M}) = \mathbf{M} - \tfrac13(\operatorname{tr}\mathbf{M})\mathbf{I}$, which isolates the direction-dependent content. The \emph{spatial pattern} is the Pearson correlation across nodes between the traces $\operatorname{tr}\Kt_i$ and $\operatorname{tr}\mathbf{K}_{ii}$; the trace is the sum of the eigenvalues, a per-node stiffness magnitude, so this correlation tests whether the model places stiffness where the finite-element tangent does, independently of any global scale. A fourth quantity, the ratio of the median traces, measures the absolute scale.

The decoded operators align with the corresponding finite-element matrices in direction (cosine $\approx0.85$ for stiffness and $0.92$ for damping) and reproduce their spatial pattern (trace correlation $\approx0.63$ and $0.94$, respectively), but they do not recover the directional anisotropy (the deviatoric cosine is approximately zero) or the absolute scale, which differs by factors of two to nine from the stiffness and by roughly $10^{3}$ from the damping. The position-response operator also changes across the three stages although the true stiffness is constant, so it adapts to the state rather than reproducing a fixed material tangent. Two further probes locate the operators mechanically. The position-response direction is nearly orthogonal to the pointwise Jacobian of the decoded internal force (cosine $-0.05$ to $+0.09$), confirming that it is not the derivative of the force channel. It nevertheless predicts the change in the model's momentum response to a velocity perturbation with a cosine of $0.99$ to $1.00$. The operators therefore describe the response produced by the learned update, not an identified constitutive tangent.

The recovered content therefore determines what the operators can be used for. They recover the direction and the spatial organisation of the finite-element tangent, which a \emph{relative} readout requires, but not its anisotropy or its absolute scale, which a calibrated material tangent would need. \rev{They are therefore usable as a relative, interpretive readout of where the structure is stiff or soft; they cannot be read as a calibrated material or constitutive tangent, as the derivative of the decoded force, or as a predictor of an absolute stress or of the response to a change of material.}


\section{Human Motion Capture}
\label{si:mocap}

\subsection{Case description}
\label{si:mocap:desc}

We use the walking sequences of subject 35 of the CMU motion capture database~\cite{cmumocap} and follow the prediction span and data-split protocol introduced by \eghn~\cite{eghn}, so that our numbers are directly comparable with the values published for that benchmark. The task is to predict the marker configuration one prediction span, $\texttt{delta\_frame} = 30$ raw frames, ahead of the observed state.

\textbf{Observed inputs and graph.} All models receive marker positions, the causal velocities of Equation~\eqref{eq:si:evaluation:velocity} and the anatomical graph. The $31$ markers are connected by $60$ directed skeletal edges. \model augments this graph with $62$ directed virtual edges to one hub, giving $122$ edges in total. The baseline graph contains two-hop skeletal shortcuts and has $130$ directed edges. The edge budgets are therefore similar, while the topologies differ.

The external forcing is not observed. \model represents it using a parameter-free spectral frame formed from the two leading non-trivial eigenvectors of the body-subgraph Laplacian. The frame is rotation-equivariant, translation-invariant and determined by graph topology rather than pose.

\begin{table}[h]
\centering
\caption{\textbf{Human-motion representation supplied to \model.} Baseline-specific features are described in Section~\ref{si:mocap:baselines}.}
\label{tab:si:mocap:rep}
\small
\begin{tabular}{ll}
\toprule
Property & Value \\
\midrule
Nodes & 31 markers \\
Physical edges & 60 skeletal connections \\
Hub edges & 62 virtual edges to one hub (122 directed in total) \\
Node features & position, finite-difference velocity \\
Edge features & scalar attribute ($-1$ on hub edges) \\
Edge-local frame & yes (\dgn) \\
Global frame & yes (spectral, from the body-subgraph Laplacian) \\
External force & inferred; SO(3)-equivariant, translation-invariant \\
Seeds & $\{0, 42, 100\}$, $n = 3$ \\
\bottomrule
\end{tabular}
\end{table}

\textbf{Splits and seeds.} Training, validation and test partitions follow the reference protocol. Every model is trained from three random seeds, $\{0, 42, 100\}$, and results are reported as the mean and standard deviation over those seeds, $n = 3$. This is the only case in this work for which a dispersion measure is available.

\textbf{Rotation evaluation.} To assess whether a model's predictions are covariant with a rotation of the observation frame, we rotate the entire test trajectory about the vertical walking axis by a fixed angle of $5^\circ$, $15^\circ$ or $30^\circ$ and re-run the identical rollout evaluation. The $0^\circ$ condition reproduces the unrotated row exactly and serves as a control.

\subsection{Evaluated baselines}
\label{si:mocap:baselines}

We compare \model with \gns, \eghn, \eghno, \egno, \dgn and \dgn~(\rev{global reference frame}), \rev{which gives \dgn the global reference frame used by \model}. \eghn, \eghno and \egno use the velocity heads of Section~\ref{si:baselines:velhead}. The corrected rollout variant of \eghno is used: its velocity is decoded from the hierarchical representation rather than from the low-level block alone.

The architectures retain their published input representations. \gns receives the current and previous velocities at each node; its edges carry relative position, relative velocity and their norms. These inputs are translation-invariant, but its Cartesian multilayer perceptrons are not rotation-equivariant. \eghn, \eghno and \egno receive absolute height as an invariant scalar. \dgn constructs a normalised height internally at every message-passing round. \model and \gns receive no absolute height.

\rev{\dgn~(global reference frame) gives \dgn's external-force channel the same parameter-free spectral frame that \model uses; \dgn is itself rotation-equivariant, and the original \dgn is retained to separate the effect of this frame.} Because the rotation test is performed about the vertical axis, height is unchanged by the applied transformation.

\subsection{Hyperparameters}
\label{si:mocap:hparams}

Settings for \model not listed in Table~\ref{tab:si:mocap} follow Table~\ref{tab:si:model:shared}.

\begin{table}[h]
\centering
\caption{\textbf{Human motion-capture hyperparameters.}
Capacity parameters have architecture-specific meanings, as explained in
Section~\ref{si:baselines:adaptations}.}
\label{tab:si:mocap}

\begingroup
\footnotesize
\setlength{\tabcolsep}{3pt}
\renewcommand{\arraystretch}{1.1}

\begin{tabularx}{\linewidth}{
  @{}
  >{\raggedright\arraybackslash}X
  c
  >{\raggedright\arraybackslash}p{0.17\linewidth}
  c
  c
  c
  @{}
}
\toprule
Model
& \shortstack{Latent\\$n_f$}
& Capacity parameter
& Batch
& \shortstack{Learning\\rate}
& \shortstack{Weight\\decay} \\
\midrule
\model
& 64 & $S=4$ substeps & 12
& $4\times10^{-4}$ & $10^{-6}$ \\

\dgn
& 64 & $S=4$ substeps & 12
& $4\times10^{-4}$ & $10^{-6}$ \\

\shortstack[l]{\dgn\\(\rev{global reference frame})}
& 64 & $S=4$ substeps & 12
& $4\times10^{-4}$ & $10^{-6}$ \\

\gns
& 64 & $S=4$ substeps & 12
& $4\times10^{-4}$ & $10^{-6}$ \\

\eghn
& 64 & 5 clusters & 12
& $4\times10^{-4}$ & $10^{-10}$ \\

\eghno
& 64 & 5 clusters & 12
& $4\times10^{-4}$ & $10^{-6}$ \\

\egno
& 128 & 6 layers & 12
& $5\times10^{-4}$ & $10^{-6}$ \\
\bottomrule
\end{tabularx}

\endgroup
\end{table}

All models train for at most $10\,000$ epochs with early stopping. Prediction span $\texttt{delta\_frame} = 30$ frames for every model.

\subsection{Additional results}
\label{si:mocap:additional}

Table~\ref{tab:si:mocap:seeds} reports the per-step rollout error for the seven models, unclipped. \rev{Two behaviours visible in the table are omitted from the main text. First, the equivariant baselines \eghn, \eghno and \egno stay finite on the canonical split but grow large and unstable by the five-step horizon, reaching mean squared errors of roughly four to six, with \eghno carried to a large finite value by a single seed; their equivariant construction keeps them bounded but does not control the coarse-step growth. Second, \model, whose learned response operators enter a semi-implicit solve, is the most accurate of the equivariant models at every step, whereas the non-equivariant \gns is more accurate still on the unrotated set but loses that advantage under the rotation of Table~\ref{tab:si:mocap:rot}.}

\begin{table}[h]
\centering
\caption{\rev{\textbf{Per-step rollout error on the human-walk benchmark.} Unscaled position mean
squared error, mean $\pm$ standard deviation over seeds $\{0,42,100\}$, $n = 3$, on the canonical
split. Values are not clipped. Every model stays finite across the five-step rollout on this split;
\eghno reaches a large finite value at step five, driven by seed~0 ($1.7\times10^{7}$) while its
other seeds remain near $10$.}}
\label{tab:si:mocap:seeds}
\setlength{\tabcolsep}{4.5pt}
\resizebox{\textwidth}{!}{%
\begin{tabular}{lccccc}
\toprule
Method & 1 step & 2 steps & 3 steps & 4 steps & 5 steps \\
\midrule
\model (implicit)             & $0.0999 \pm 0.0069$ & $0.2491 \pm 0.0153$ & $0.4773 \pm 0.0246$ & $0.8288 \pm 0.0282$ & $1.3349 \pm 0.0195$ \\
\gns                          & $0.0788 \pm 0.0041$ & $0.1639 \pm 0.0079$ & $0.2735 \pm 0.0059$ & $0.3785 \pm 0.0079$ & $0.5006 \pm 0.0324$ \\
\dgn~(global reference frame) & $0.1448 \pm 0.0036$ & $0.4210 \pm 0.0306$ & $0.8622 \pm 0.0672$ & $1.5319 \pm 0.1439$ & $2.4480 \pm 0.3098$ \\
\dgn                          & $0.2346 \pm 0.0086$ & $0.7960 \pm 0.0288$ & $1.7010 \pm 0.0459$ & $3.2218 \pm 0.1868$ & $4.7374 \pm 0.3951$ \\
\eghn                         & $0.0872 \pm 0.0088$ & $0.2577 \pm 0.0087$ & $0.6128 \pm 0.0535$ & $1.5396 \pm 0.3390$ & $4.0918 \pm 1.5819$ \\
\eghno                        & $0.1208 \pm 0.0328$ & $0.4931 \pm 0.1694$ & $1.3513 \pm 0.5260$ & $2.8605 \pm 0.8386$ & $5.83\times10^{6}$ \\
\egno                         & $0.0940 \pm 0.0211$ & $0.3786 \pm 0.1015$ & $1.0551 \pm 0.3579$ & $2.5967 \pm 1.0046$ & $6.0528 \pm 2.7320$ \\
\bottomrule
\end{tabular}}
\end{table}

\begin{table}[h]
\centering
\caption{\textbf{Degradation of \gns under a rotation of the test set.} Per-step rollout error, mean
$\pm$ standard deviation over seeds $\{0,42,100\}$, $n = 3$. The $0^\circ$ row reproduces the \gns
row of Table~\ref{tab:si:mocap:seeds}. \model is rotation-equivariant by construction and its error
is unchanged at every angle, so it is not tabulated per angle.}
\label{tab:si:mocap:rot}
\small
\begin{tabular}{lccccc}
\toprule
Rotation & 1 step & 2 steps & 3 steps & 4 steps & 5 steps \\
\midrule
$0^\circ$  & $0.0788 \pm 0.0041$ & $0.1639 \pm 0.0079$ & $0.2735 \pm 0.0059$ & $0.3785 \pm 0.0079$ & $0.5006 \pm 0.0324$ \\
$5^\circ$  & $0.1566 \pm 0.0142$ & $0.5243 \pm 0.0393$ & $1.0978 \pm 0.0591$ & $1.7665 \pm 0.0663$ & $2.5097 \pm 0.0941$ \\
$15^\circ$ & $0.8958 \pm 0.0655$ & $3.2939 \pm 0.1031$ & $7.3419 \pm 0.2866$ & $12.3016 \pm 0.3948$ & $17.8291 \pm 0.4667$ \\
$30^\circ$ & $3.0708 \pm 0.0849$ & $10.6594 \pm 0.1988$ & $23.7211 \pm 1.4052$ & $42.2121 \pm 2.0511$ & $64.2818 \pm 1.8082$ \\
\bottomrule
\end{tabular}
\end{table}

The error of \gns increases sharply with rotation angle and compounds over the rollout horizon, which is the behavior expected of a model that has no built-in rotational symmetry and was trained on trajectories captured in a single orientation. At $5^\circ$, a misalignment small enough to arise from the calibration of a capture volume, \gns already predicts less accurately than \model at the five-step horizon.

\section{Protein Dynamics in Solvent}
\label{si:protein}

\subsection{Case description}
\label{si:protein:desc}

\rev{We evaluate \model on a large collective conformational change of adenylate kinase: the driven transition ensemble of Seyler and Beckstein~\cite{seyler2017adk}, accessed through the MDAnalysis toolkit~\cite{mdanalysis2016}. The ensemble contains $200$ dynamic-importance-sampling paths, each following the $855$ backbone atoms through the conformational change over $90$ to $106$ frames. Along every path the radius of gyration and the inter-domain distances increase monotonically, so the trajectories are closed--open transitions in which the two mobile domains swing away from the rigid core. These are enhanced-sampling paths rather than unbiased kinetics: one step is a fixed fraction of the transition, not a physical time, and we read the benchmark as reproduction of the collective pathway.}

\rev{Unlike the equilibrium trajectory of the same protein (Section~\ref{si:protein:floor}), which samples confined fluctuation within a single basin and is bounded by a linear floor, this ensemble carries a large directed motion: a root-mean-square displacement of about $7\,\text{\AA}$ between
the closed and open states. Its predictable content is nonetheless dominated by a single canonical
route. A state-independent mean-trajectory baseline, which moves every test path along the same
average closed-open transition of the training paths regardless of its configuration, already
reaches a position error of $0.23$ to $0.37\,\text{\AA}^2$ across the five-step horizon; \model stays
only $4$ to $10\%$ below it while remaining bounded. The benchmark therefore tests bounded, stable
rollout of the collective transition and graph sparsity, not a large accuracy margin over this
trivial predictor.}

\rev{\textbf{Task and split.} We predict the configuration $\Delta = 15$ frames ahead over five autoregressive steps, a $75$-frame horizon spanning most of the transition. Because each path is a distinct trajectory, the split is by path: $140$ training, $30$ validation and $30$ test paths ($241$ five-step test windows). All models receive atomic positions and the causal backward-difference velocities of Equation~\eqref{eq:si:evaluation:velocity}. \model uses a one-hot atom-type feature, the $1708$ directed covalent bonds and one hub with $1710$ virtual edges ($3418$ directed edges), identical to its equilibrium configuration. \dgn is the controlled ablation with the same bonds and no hub. \eghn and \egno augment the bonds with their published $10\,\text{\AA}$ radius graph, adding approximately $55\,610$ directed contact edges.}

Atom type and partial charge are distinct physical descriptors and are reported explicitly rather than treated as equivalent inputs. Adding raw partial charge to \model prevented convergence in our implementation, so \model is evaluated with atom type alone. \rev{\eghn and \egno retain their published charge input.}

\begin{table}[h]
\centering
\caption{\textbf{Protein representation supplied to \model.} \rev{The input and contact graph used by \eghn and \egno are} described in the text.}
\label{tab:si:protein:rep}
\small
\begin{tabular}{ll}
\toprule
Property & Value \\
\midrule
Nodes & 855 backbone atoms \\
Physical edges & 1708 covalent backbone bonds \\
Hub edges & 1710 virtual edges to one hub (3418 directed in total) \\
Node features & position, causal backward-difference velocity, atom-type one-hot \\
Edge features & scalar attribute ($-1$ on hub edges) \\
Edge-local frame & yes (\dgn) \\
Global frame & yes (spectral, from the body-subgraph Laplacian) \\
External force & inferred; SO(3)-equivariant, translation-invariant \\
Seeds & \rev{$\{42\}$}, $n = 1$ \\
\bottomrule
\end{tabular}
\end{table}

\textbf{Metric.} Unscaled position mean squared error, per rollout step, over five autoregressive steps of $15$ frames each.

\subsection{Evaluated baselines}
\label{si:protein:baselines}

\rev{All four models are trained on this ensemble under the by-path split, one seed each. \model uses Equation~\eqref{eq:si:model:loss}; \dgn applies the same normalised-increment objective on the hub-free bond graph; \eghn uses the velocity head of Section~\ref{si:baselines:velhead} with its $\lambda_{\mathrm{link}}$-weighted linkage term; \egno uses its multi-frame objective. All are evaluated by the same unscaled position error under autoregressive rollout and retain the checkpoint with minimum validation loss. The contact-graph models carry the cost of their $10\,\text{\AA}$ graph: approximately $55\,610$ directed contact edges against the $1710$ hub edges of \model, so the hub supplies the global coupling required for the collective transition at a fraction of the graph density.}

\subsection{Hyperparameters}
\label{si:protein:hparams}

Settings for \model not listed in Table~\ref{tab:si:protein} follow Table~\ref{tab:si:model:shared}. This is the one case in which the latent width of \rev{\eghn and \egno} departs from ours: the reference protocol trains \eghn and \egno at $128$, and we retain their settings rather than reduce them.

\begin{table}[h]
\centering
\caption{\textbf{Protein dynamics: hyperparameters.} \rev{\model and \dgn are trained at $nf = 64$, as
on every other system; \eghn and \egno retain their published width of $128$.}}
\label{tab:si:protein}
\small
\begin{tabular}{lccccc}
\toprule
Model & latent $nf$ & capacity parameter & batch & learning rate & weight decay \\
\midrule
\model & 64  & $S = 4$ sub-steps & 8 & $5\times10^{-4}$ & $10^{-8}$ \\
\rev{\dgn} & \rev{64} & \rev{$5$ messages} & \rev{8} & \rev{$5\times10^{-4}$} & \rev{$10^{-8}$} \\
\eghn  & 128 & $20$ clusters     & 8 & $5\times10^{-4}$ & $10^{-8}$ \\
\egno  & 128 & $4$ layers        & 1 & $5\times10^{-4}$ & $10^{-8}$ \\
\bottomrule
\end{tabular}
\end{table}

\rev{All four models train with early stopping at a patience of $50$ validation checks. Prediction span $\Delta = 15$ frames throughout. Seed $42$.}

\subsection{Additional results}
\label{si:protein:additional}

\rev{Table~\ref{tab:si:protein:rollout} gives the full rollout for all evaluated models. \model remains bounded across the full horizon, rising from $0.221$ to a maximum of $0.346$ and settling near $0.28\,\text{\AA}^2$. \dgn, the hub-free ablation, is accurate at one step ($0.386$) but its rollout diverges to $6.12$; \eghn and \egno track to the third step and then diverge, reaching $361$ and non-finite values. The state-independent mean-trajectory baseline applies the training-averaged displacement at each step, regardless of the test path (last row of Table~\ref{tab:si:protein:rollout}).}

\begin{table}[h]
\centering
\caption{\rev{\textbf{Per-step rollout error on the transition ensemble.} Unscaled position mean squared
error (\AA$^2$). One step is $15$ frames, evaluated on the $30$ held-out test paths ($241$ five-step
windows, $855$ atoms) from each model's minimum-validation checkpoint. NaN entries mark a
diverged rollout.}}
\label{tab:si:protein:rollout}
\small
\begin{tabular}{lccccc}
\toprule
Method & 1 step & 2 steps & 3 steps & 4 steps & 5 steps \\
\midrule
\model & \rev{$0.221$} & \rev{$0.328$} & \rev{$0.346$} & \rev{$0.296$} & \rev{$0.277$} \\
\dgn   & \rev{$0.386$} & \rev{$1.164$} & \rev{$2.559$} & \rev{$4.499$} & \rev{$6.122$} \\
\eghn  & \rev{$0.214$} & \rev{$0.366$} & \rev{$0.679$} & \rev{$361.4$} & \rev{NaN} \\
\egno  & \rev{$0.228$} & \rev{$0.417$} & \rev{$1.018$} & \rev{$11.03$} & \rev{NaN} \\
\midrule
\rev{mean-trajectory} & \rev{$0.231$} & \rev{$0.341$} & \rev{$0.369$} & \rev{$0.330$} & \rev{$0.307$} \\
\bottomrule
\end{tabular}
\end{table}

\subsection{\rev{Why we do not use the equilibrium benchmark}}
\label{si:protein:floor}

\rev{Adenylate kinase also has an established prediction benchmark on its \emph{equilibrium} trajectory (apo, explicit solvent, $300\,$K and $1\,$bar, recorded every $240\,$ps over approximately $1\,\mu$s, the same $855$ backbone atoms, a $15$-frame horizon), introduced by \eghn~\cite{eghn} and adopted by \egno~\cite{egno}. We do not use it to rank
dynamical models: its single-step error is bounded below by a trivial predictor that no learned
model beats, so it scores thermal-fluctuation statistics rather than learnable, collective
dynamics. All numbers below are measured on the trajectory, using the reference benchmark split
and using the unscaled single-step position mean squared error.}

\begin{figure}[htbp]
  \centering
  \includegraphics[width=0.96\textwidth]{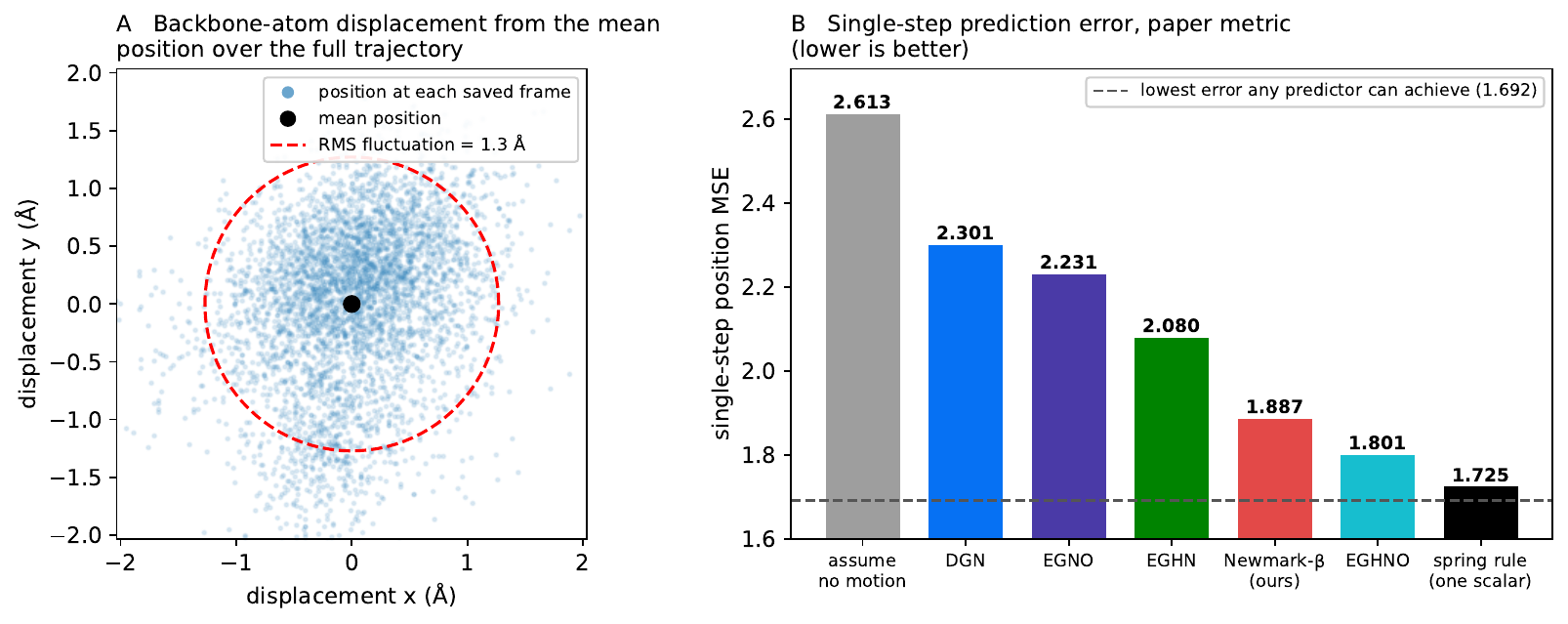}
  \caption{\rev{\textbf{The single-step benchmark is bounded by a data floor.} (a) One backbone atom's
  position at every recorded frame, relative to its mean; its root-mean-square fluctuation is $1.3\,\text{\AA}$.
  (b) Single-step position error for assume-no-motion, the published and reproduced learned models, and the
  one-coefficient restoring rule ($\Delta\mathbf{x} = -0.60(\mathbf{x}-\bar{\mathbf{x}})$, labelled the spring
  rule), against the lowest error any predictor of the current state can reach ($1.692$, dashed). Every
  learned model lies above the restoring rule and the floor. Lower is better.}}
  \label{fig:si:protein:floor}
\end{figure}

\rev{\textbf{The motion is a confined fluctuation.} Over the trajectory each backbone atom fluctuates about a fixed mean position with a root-mean-square amplitude of $1.9\,\text{\AA}$, from $0.8\,\text{\AA}$ in the rigid core to about $6\,\text{\AA}$ in the flexible loops (Fig.~\ref{fig:si:protein:floor}a), with no net translation or change of fold. The position autocorrelation at the $3.6\,$ns ($15$-frame) horizon is $C = 0.40$, a relaxation time of $\tau \approx 2.7\,$ns.}

\rev{\textbf{A linear restoring map captures the predictable component.}
The per-atom position increments are approximately Gaussian, with an excess
kurtosis of $0.71$. The larger kurtosis obtained by pooling all atoms results
from combining rigid-core and flexible-loop atoms with substantially different
fluctuation amplitudes. Under a stationary Gaussian approximation, the
minimum-mean-square predictor based on the present position is
\[
  \widehat{\mathbf{x}}_{t+1}
  =
  \bar{\mathbf{x}}
  +
  C\left(
    \mathbf{x}_t-\bar{\mathbf{x}}
  \right),
  \qquad C=0.40.
\]
Equivalently, the predicted displacement is
\[
  \Delta\widehat{\mathbf{x}}
  =
  (C-1)
  \left(
    \mathbf{x}_t-\bar{\mathbf{x}}
  \right)
  =
  -0.60
  \left(
    \mathbf{x}_t-\bar{\mathbf{x}}
  \right).
\]
This one-coefficient rule obtains a single-step mean squared error of
$1.7252\,\text{\AA}^2$, while fitting one coefficient per atom reduces the
error to $1.6915\,\text{\AA}^2$, the lowest value obtained by the predictors
examined here. Relative to the assume-no-motion error of
$2.6127\,\text{\AA}^2$, the per-atom map explains
$0.9212\,\text{\AA}^2$ of the displacement error. The remaining
$1.6915\,\text{\AA}^2$ shows no detectable correlation with the present
position, velocity, or neighbouring-atom coordinates.}

\rev{\textbf{No model beats the floor.} \model\ ($1.8868$), \eghn\ ($2.0801$) and the published~\cite{egno} \egno\ ($2.231$), \dgn\ ($2.301$) and \eghno\ ($1.801$) all lie above both the $1.7252$ of the one-coefficient restoring rule and the $1.6915$ floor (Fig.~\ref{fig:si:protein:floor}b). A linear map using all atom coordinates jointly reaches $29.9\%$ skill, no better than the per-atom rule, and after the restoring prediction is removed a regularised nonlinear model explains none of the remaining error out of sample. The ranking among the learned models therefore measures how closely each reproduces one linear coefficient, not the accuracy of a dynamical model.}

\rev{\textbf{Why the horizon leaves only statistics.} The atomic trajectory decorrelates on a Lyapunov timescale of about $0.2\,$ps, so the $3.6\,$ns horizon is roughly $18{,}000$ times longer and only the $2.7\,$ns statistical relaxation survives. Shortening the horizon lowers the attainable skill, to $21\%$ at a single frame, so no choice of horizon converts this benchmark into a test of dynamics. We therefore evaluate the collective dynamics on the driven transition ensemble of Section~\ref{si:protein:desc}, where the models are separated by their rollout stability rather than by a shared linear floor.}

\section{Joint-Moment Inference in Human Walking}
\label{si:biomech}

\subsection{Case description}
\label{si:biomech:desc}

We use a public dataset of healthy human walking recorded on an instrumented treadmill~\cite{vanderzee2022}, which provides marker trajectories together with joint moments and ground reaction forces obtained from an inverse-dynamics pipeline. We use subject p2, comprising $33$ trials and $18\,631$ frames recorded at $120\,$Hz. The measured moments are read only for evaluation and never enter Equation~\eqref{eq:si:model:loss}.

\textbf{Observed inputs and graph.} The $37$ markers are the nodes and carry positions and the causal velocities of Equation~\eqref{eq:si:evaluation:velocity}. The $63$ anatomical connections between markers are the physical edges, and \model adds $74$ directed virtual edges to one hub. The model receives no joint moment or ground reaction force. One prediction span is $\texttt{delta\_frame}=30$ frames, matching the motion-capture case.

\begin{table}[h]
\centering
\caption{\textbf{Graph representation of the biomechanics system.}}
\label{tab:si:biomech:rep}
\small
\begin{tabular}{ll}
\toprule
Property & Value \\
\midrule
Nodes & 37 markers \\
Physical edges & 63 anatomical connections \\
Hub edges & $2N = 74$ virtual edges to one hub \\
Node features & position, causal backward-difference velocity \\
Edge features & scalar attribute ($-1$ on hub edges) \\
Edge-local frame & yes (\dgn) \\
Global frame & yes (spectral, from the body-subgraph Laplacian) \\
External force & inferred; SO(3)-equivariant, translation-invariant \\
Seeds & single seed, $n = 1$ \\
\bottomrule
\end{tabular}
\end{table}

\textbf{Gait conditions and splits.} The split is by gait condition rather than by time. Training covers preferred walking at $0.7$, $1.1$, $1.25$, $1.4$, $1.6$ and $1.8\,\mathrm{m\,s^{-1}}$, together with two cadence variants at $1.25\,\mathrm{m\,s^{-1}}$ in which step frequency is raised while speed is held fixed. Validation covers preferred walking at $0.9$ and $1.25\,\mathrm{m\,s^{-1}}$ and the remainder of the $1.25\,\mathrm{m\,s^{-1}}$ cadence sweep. Held out entirely are the two constrained gait strategies, constant step length and constant step frequency, at every recorded speed, together with preferred walking at $2.0\,\mathrm{m\,s^{-1}}$.

Cadence and gait strategy are distinct axes. Varying step frequency at a fixed preferred speed is a condition on which the model is trained and validated. Holding step length or step frequency fixed while speed varies is not. The non-monotonic cadence response reported in the main text therefore concerns supervision rather than generalisation, whereas the constant-step-length, constant-step-frequency and $2.0\,\mathrm{m\,s^{-1}}$ conditions are the extrapolation result.


\textbf{Moment readout.} From the decoded internal forces the moment about each joint centre is formed by summing over the markers distal to that joint, $\mathbf{M} = -\sum_{i \in \text{distal}} \mathbf{r}_i \times \mathbf{F}_i$, with $\mathbf{r}_i$ measured from the joint centre. Only this contribution, the moment about the joint centre, is used. No term of the loss constrains the spin channel, so it is not an identifiable physical quantity and is not read out. The summation runs over thirteen markers at the hip, seven at the knee and two at the ankle.

\textbf{Evaluation sampling.} Trend metrics tile each trial with non-overlapping windows rather than sampling start frames at random. The two legs of a walking human are close to half a gait cycle out of phase, so a short randomly placed window can capture the stance-phase peak of one leg while clipping the other, and averaging over many such windows does not cancel the bias because the sampling is not phase-aware. Deterministic tiling covers every phase of every trial exactly once. The trial supplying the normalisation denominator is tiled at unit stride so that the divisor is low-noise. The measured moments are read at the native $120\,$Hz rate rather than at the sparse prediction-span rate: sampling the ground truth sparsely aliases the $\approx\!0.8\,$s stride period and returns a left--right ankle RMS ratio of $0.47$ where the densely sampled signal gives $0.99$.

\textbf{Normalisation.} Demand is a ratio rather than a calibrated moment. For each joint the root-mean-square of the moment profile is divided by the corresponding value in the preferred $1.25\,\mathrm{m\,s^{-1}}$ trial, at unit subject mass. Measured and inferred series are normalised separately. Within each series, both legs are divided by the mean of that series' left and right baseline values, so one divisor is used per source and the raw left--right asymmetry survives into the plotted curves. Dividing each leg by its own baseline would remove part of that asymmetry. Pearson correlations are invariant to such per-series rescaling; the normalised RMS errors and the asymmetry statistics are not, and all quoted values use the shared-divisor convention.

\textbf{Seeds.} A single trained model is used, on a single subject. The error bars in the figures of this section are the spread over evaluation tiles of a trial and not over seeds; the tiled evaluation is deterministic.

\subsection{Evaluated baselines}
\label{si:biomech:baselines}

No baseline is evaluated in this case. The purpose is to compare the joint-moment readout of \model with an independent inverse-dynamics reference, not to rank forward simulators. Forward prediction on articulated human motion is compared with the baselines in Section~\ref{si:mocap}.

\subsection{Hyperparameters}
\label{si:biomech:hparams}

Settings not listed in Table~\ref{tab:si:biomech} follow Table~\ref{tab:si:model:shared}.

\begin{table}[h]
\centering
\caption{\textbf{Biomechanics: hyperparameters.}}
\label{tab:si:biomech}
\small
\begin{tabular}{lccccc}
\toprule
Model & latent $nf$ & capacity parameter & batch & learning rate & weight decay \\
\midrule
\model & 64 & $S = 4$ sub-steps & 128 & $5\times10^{-4}$ & $10^{-10}$ \\
\bottomrule
\end{tabular}
\end{table}

At most $2000$ epochs with early stopping. Prediction span $\texttt{delta\_frame} = 30$ frames.

\subsection{Additional results}
\label{si:biomech:additional}

\begin{figure}[H]
  \centering
  \includegraphics[width=0.70\textwidth]{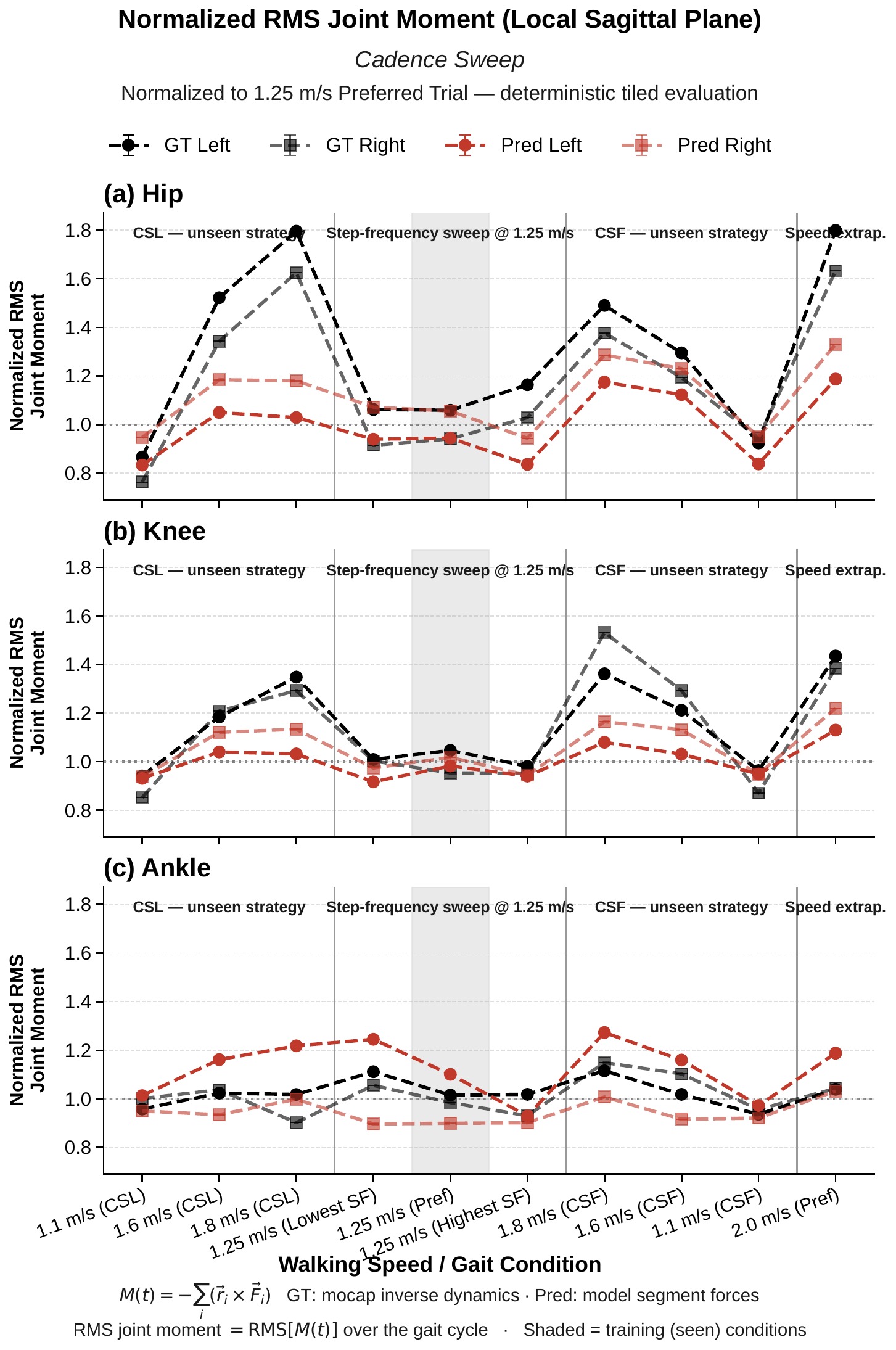}
  \caption{\textbf{Inferred demand at all three joints across the cadence sweep.} Hip, knee and ankle,
  measured against inferred, for the conditions of the main-text figure. The hip and knee track the
  measurement; the ankle does not, for the reason given above. Shaded groups lie within the training
  and validation data.}
  \label{fig:si:biomech:exp6}
\end{figure}

\begin{figure}[H]
  \centering
  \includegraphics[width=0.70\textwidth]{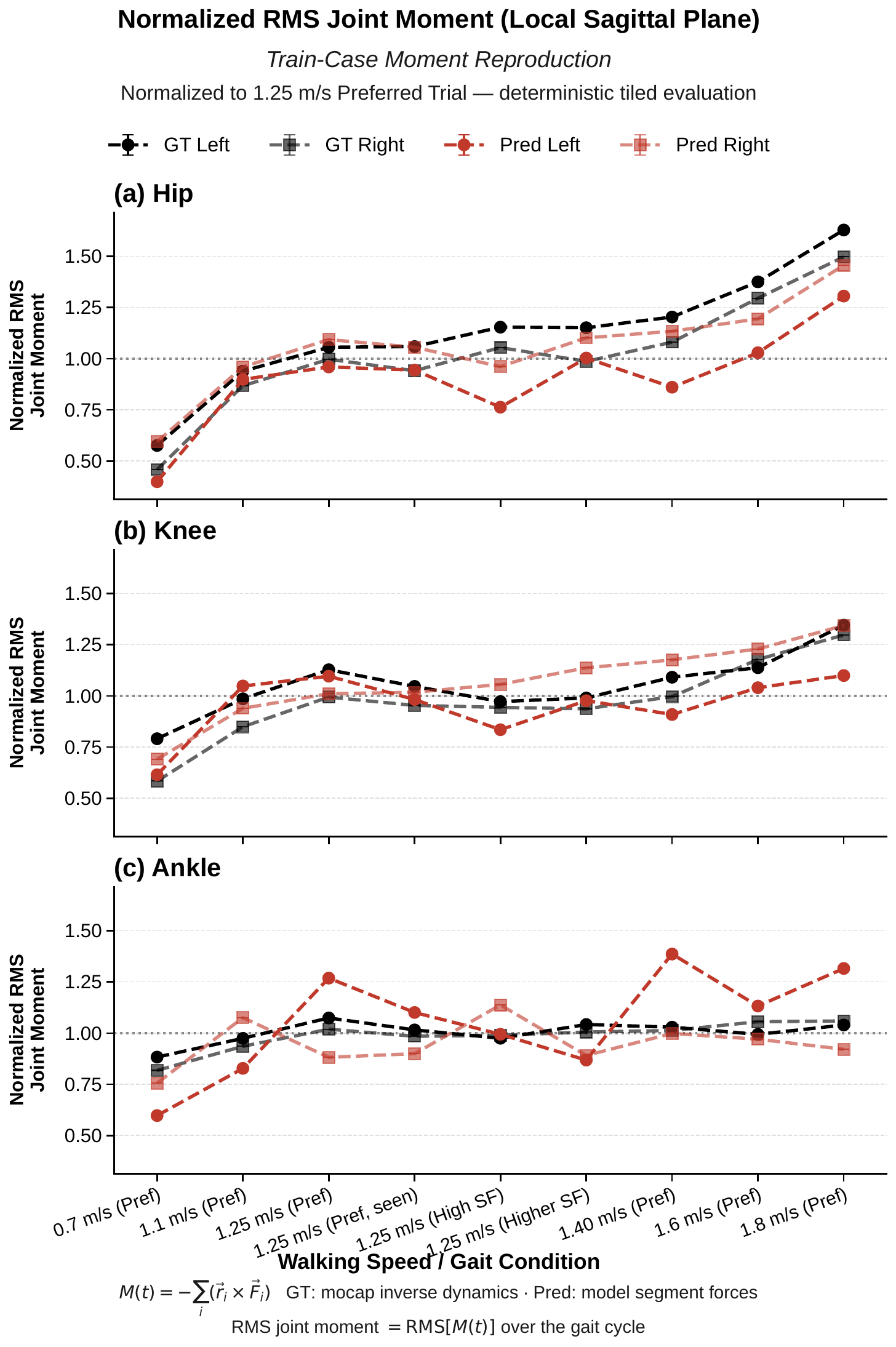}
  \caption{\textbf{In-distribution control.} The same three joints evaluated on conditions seen during
  training. Agreement at the knee ($r = 0.81$) is as good on these training conditions as on the
  withheld ones ($r = 0.88$). Readout accuracy is therefore limited by the moment model, not by
  distribution shift, and the inferred profiles are not artefacts of fitting.}
  \label{fig:si:biomech:exp4}
\end{figure}

The main text reports the knee. Figures~\ref{fig:si:biomech:exp6} and \ref{fig:si:biomech:exp4} give the hip and the ankle alongside it, for the cadence sweep of the main text and for the in-distribution control. The remaining sweeps behave comparably and are not reproduced here.

Pooled across five evaluation experiments and both legs, the inferred profile agrees with the measured one at a Pearson correlation of $0.943$ at the hip, $0.869$ at the knee and $0.492$ at the ankle, with normalised RMS errors of $0.171$, $0.213$ and $0.887$. The ankle is not recovered: it is anticorrelated with measurement in one condition, at $r = -0.322$, and its magnitude is low by roughly an order of magnitude. The left--right asymmetry of the gait is likewise absent. The measured imbalance ranges from $0.006$ to $0.170$ across conditions while the inferred imbalance remains at $0.048 \pm 0.017$, and the prediction error grows with the true asymmetry, at $r = +0.76$, reaching a mean error of $17.3\%$ in the condition where that asymmetry is greatest.

A contributing limitation is that the joint-moment readout is assembled from decoded internal segment forces and does not include the measured ground-reaction contribution; this is particularly restrictive at the ankle, where only two distal markers contribute to the readout. 

\section{\rev{Ablations}}
\label{si:ablations}

\rev{This section collects the ablations of \model's own components. Each removes or replaces one ingredient of the update and reports its effect. Four are established where they arise in the case sections and are summarised in Table~\ref{tab:si:ablations}; the angular-momentum channel, whose effect is on the inferred mechanics rather than on the rollout, is treated below.}

\begin{table}[htbp]
\centering
\caption{\textbf{Ablations of \model's components.} Each row removes or replaces one ingredient of the
update. The first four are established in the sections cited; the last is treated in Section~8.1.
Beam errors are whole-body, mean over the reported configurations, seed 42.}
\label{tab:si:ablations}
\small
\begin{tabular}{lll}
\toprule
Ablation & Effect & Established in \\
\midrule
Remove the operator-weighted hub \\(\& 12 substeps instead of 4)                    & beam error $0.54\%\to3.00\%$ & Table~5 \\
Set the implicit stiffness term $\beta=0$            & beam error $0.54\%\to0.70\%$ & Table~5 \\
\rev{Per-node update made explicit ($\mathbf{A}_i=\mathbf{I}$)} & \rev{beam error $0.54\%\to1.26\%$} & \rev{Table~5, Section~10.7} \\
Remove the angular-momentum channel                  & rollout tied; inferred joint moments degrade & Section~8.1 \\
\bottomrule
\end{tabular}
\end{table}

\subsection{\rev{The angular-momentum channel}}
\label{si:ablations:angular}

\rev{\model carries a per-node spin and a per-edge angular-momentum flux alongside the linear channel (the Cosserat channel of \dgn~\cite{dgn}, Section~\ref{si:newmark:angular}). To isolate its contribution we train a variant that decodes only linear momentum and translational response operators, with no spin state and no angular flux; the rest of the update is unchanged. The variant is SO(3)-equivariant to $3\times10^{-6}$ on motion capture and exactly equivariant in double precision on the beam.}

\begin{figure}[htbp]
  \centering
  \includegraphics[width=0.49\textwidth]{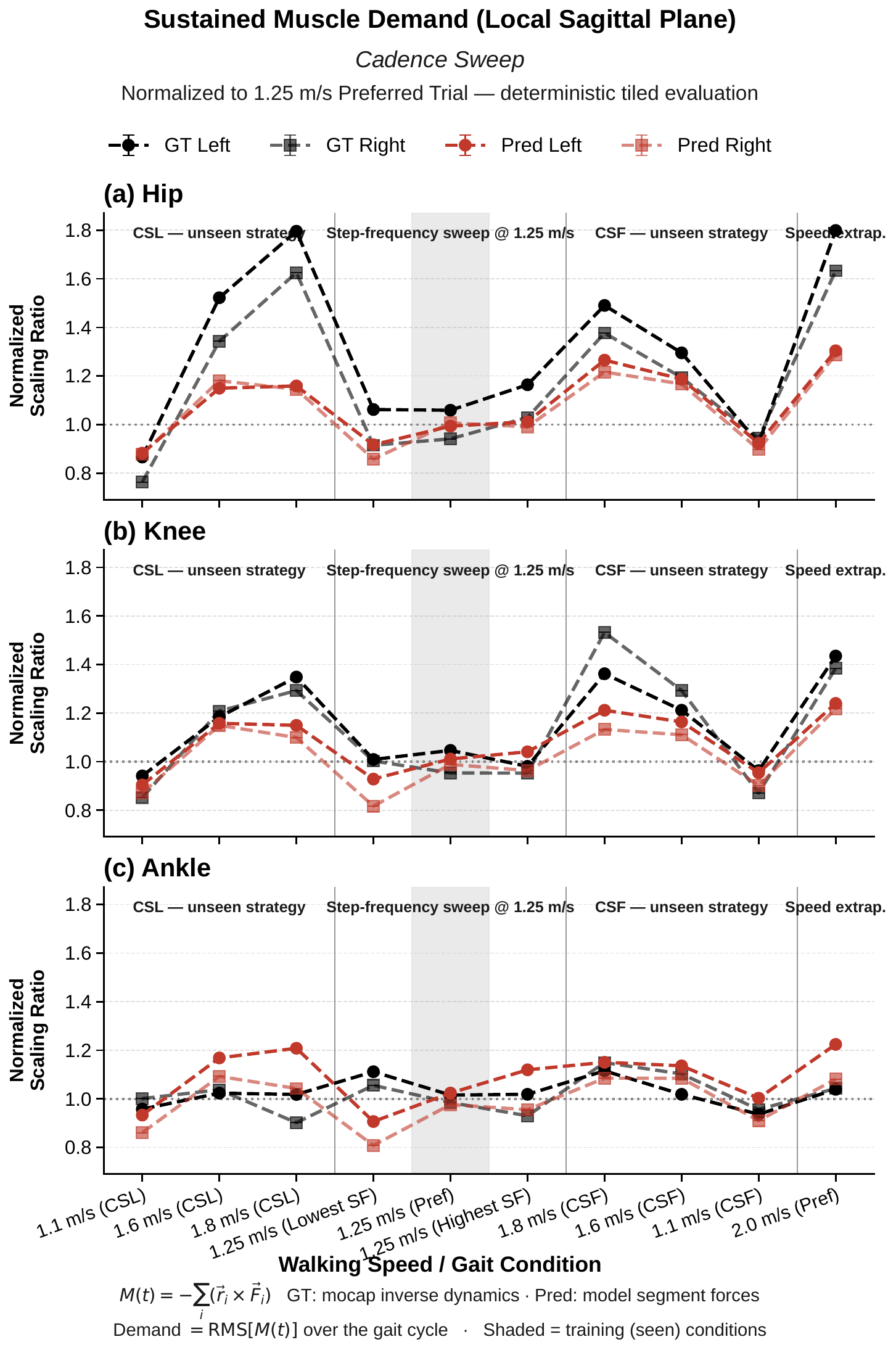}
  \hfill
  \includegraphics[width=0.49\textwidth]{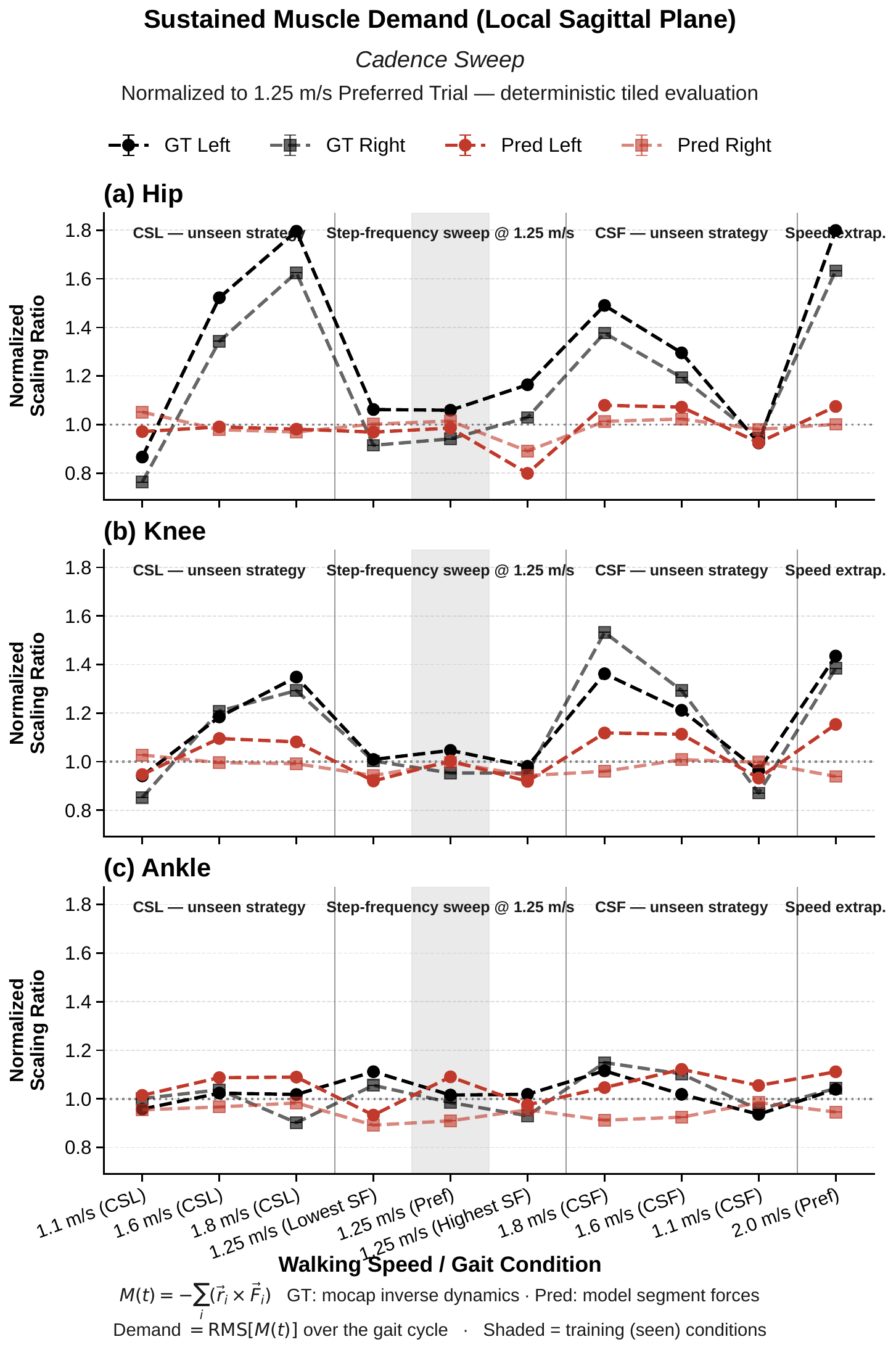}
  \caption{\rev{\textbf{The angular-momentum channel is needed for the inferred joint moments, not for the
  rollout.} Inferred against measured hip, knee and ankle demand across the cadence sweep, for the full
  model (left) and the variant with the angular-momentum channel removed (right). Measured demand is black
  and grey; inferred is red. With the channel the inferred hip and knee demand follows the measured trend;
  without it the inferred demand flattens toward unity and loses the trend. The moment construction
  $\mathbf{M} = -\sum_i \mathbf{r}_i\times\mathbf{F}_i$ is identical in both, so the difference is in the
  decoded force. The ankle is not recovered by either model, for the reason given in
  Section~\ref{si:biomech:additional}.}}
  \label{fig:si:ablation:moments}
\end{figure}

\rev{Removing the channel does not cost rollout accuracy. On motion capture the autoregressive error of the linear-only variant is slightly lower than the full model, by $6$ to $13\%$ across the five rollout steps (three seeds, with a widening spread), and on the biomechanics rollout the two are tied ($0.477$ against $0.480\,$mm). What changes is the inferred mechanics. The joint moments of Section~\ref{si:biomech} are assembled from the decoded internal forces by $\mathbf{M} = -\sum_i \mathbf{r}_i\times\mathbf{F}_i$, a construction identical in both models, yet without the angular channel the inferred demand flattens and no longer follows the measured trend across gait conditions (Fig.~\ref{fig:si:ablation:moments}). The channel therefore earns its place through the quality of the inferred mechanics, not through rollout error, and we retain it.}

\rev{A mechanism is plausible but not established here. The decoded angular flux is antisymmetric on each edge and sums to zero, so the channel exchanges angular momentum conservatively, whereas the linear-only variant places no such constraint on its decoded force. We hypothesise that this missing constraint is why the force-based moment readout degrades. We have not measured the angular-momentum residual of either rollout, so the link is stated as a hypothesis rather than a demonstrated cause.}

\section{From implicit coupling to an operator-weighted hub}
\label{si:hub}

The hub provides a tractable approximation of the non-local coupling produced
by a global implicit mechanical solve. The derivation separates the two sides
of the update. On the left-hand side, static condensation produces a dense
effective interaction operator whose rank-one approximation admits a star
representation. On the right-hand side, the same condensation transfers the
force drive from eliminated degrees of freedom to the retained degrees of
freedom. \model retains this star topology and approximates both effects
without eliminating physical nodes or assembling a global Schur complement.

\subsection{Implicit time stepping produces non-local coupling}
\label{si:hub:why}

Consider the semi-discrete equations of motion over one linearised substep,
\begin{equation}
  \mathbf{M}\ddot{\mathbf{x}}
  +\mathbf{D}_{\mathrm{glob}}\dot{\mathbf{x}}
  +\mathbf{K}_{\mathrm{glob}}\mathbf{x}
  =\mathbf{f}_{\mathrm{ext}}.
  \label{eq:si:hub:eom}
\end{equation}
The mass matrix $\mathbf{M}$ is block diagonal, whereas the assembled
stiffness $\mathbf{K}_{\mathrm{glob}}$ and damping
$\mathbf{D}_{\mathrm{glob}}$ couple connected nodes. Define
\begin{equation}
  \Delta\mathbf{v}
  :=\mathbf{v}_{t+1}-\mathbf{v}_t,
  \qquad
  \Delta\mathbf{x}
  :=\mathbf{x}_{t+1}-\mathbf{x}_t.
\end{equation}
For the average-acceleration Newmark parameters
$\gamma=\tfrac12$ and $\beta=\tfrac14$,
\begin{equation}
  \mathbf{M}\Delta\mathbf{v}
  =
  \left[
    \mathbf{f}_{\mathrm{ext}}
    +\tfrac12
    \left(
      \mathbf{f}_{\mathrm{int}}^t
      +\mathbf{f}_{\mathrm{int}}^{t+1}
    \right)
  \right]\subdt,
  \qquad
  \Delta\mathbf{x}
  =
  \subdt
  \left(
    \mathbf{v}_t+\tfrac12\Delta\mathbf{v}
  \right).
  \label{eq:si:hub:newmark}
\end{equation}
Within the substep, the linearised internal-force change is
\begin{equation}
  \mathbf{f}_{\mathrm{int}}^{t+1}
  =
  \mathbf{f}_{\mathrm{int}}^t
  -\mathbf{D}_{\mathrm{glob}}\Delta\mathbf{v}
  -\mathbf{K}_{\mathrm{glob}}\Delta\mathbf{x}.
  \label{eq:si:hub:forceincrement}
\end{equation}
Using the Newmark position increment in
Equation~\eqref{eq:si:hub:forceincrement} gives
\begin{align}
  \mathbf{f}_{\mathrm{int}}^{t+1}
  &=
  \mathbf{f}_{\mathrm{int}}^t
  -\mathbf{D}_{\mathrm{glob}}\Delta\mathbf{v}
  -\subdt\,\mathbf{K}_{\mathrm{glob}}\mathbf{v}_t
  -\tfrac12\subdt\,
   \mathbf{K}_{\mathrm{glob}}\Delta\mathbf{v},
  \nonumber\\
  \tfrac12
  \left(
    \mathbf{f}_{\mathrm{int}}^t
    +\mathbf{f}_{\mathrm{int}}^{t+1}
  \right)
  &=
  \mathbf{f}_{\mathrm{int}}^t
  -\tfrac12\mathbf{D}_{\mathrm{glob}}\Delta\mathbf{v}
  -\tfrac12\subdt\,
   \mathbf{K}_{\mathrm{glob}}\mathbf{v}_t
  -\tfrac14\subdt\,
   \mathbf{K}_{\mathrm{glob}}\Delta\mathbf{v}.
  \label{eq:si:hub:average-force}
\end{align}
Multiplying the second line by $\subdt$ and collecting all terms proportional
to $\Delta\mathbf{v}$ on the left gives
\begin{equation}
  \underbrace{
  \left(
    \mathbf{M}+\mathbf{C}_{\mathrm{glob}}
  \right)\Delta\mathbf{v}
  }_{\text{left-hand side: implicit mechanical response}}
  =
  \underbrace{
  \mathbf{r}
  }_{\text{right-hand side: known force drive}},
  \label{eq:si:hub:assembled}
\end{equation}
where
\begin{equation}
  \mathbf{C}_{\mathrm{glob}}
  :=
  \tfrac12\subdt\,\mathbf{D}_{\mathrm{glob}}
  +\tfrac14\subdt^2\mathbf{K}_{\mathrm{glob}},
  \label{eq:si:hub:coupling-operator}
\end{equation}
and
\begin{equation}
  \mathbf{r}
  :=
  \subdt
  \left(
    \mathbf{f}_{\mathrm{ext}}
    +\mathbf{f}_{\mathrm{int}}^t
  \right)
  -\tfrac12\subdt^2
   \mathbf{K}_{\mathrm{glob}}\mathbf{v}_t.
  \label{eq:si:hub:known-rhs}
\end{equation}
Section~\ref{si:newmark} derives the complete update used by \model.

The mass matrix is local, while $\mathbf{C}_{\mathrm{glob}}$ contains the
inter-node coupling. Although the assembled system is sparse, eliminating
some degrees of freedom generally produces dense effective coupling on the
left-hand side and dense force transfer on the right-hand side.

\subsection{Static condensation of the implicit update}
\label{si:hub:schur}

\paragraph{Retained and internal degrees of freedom.}
Static condensation divides the degrees of freedom into a retained set $R$
and an internal set $I$. The retained set contains the nodes whose increments
remain explicit in the reduced problem. The internal set contains intermediate
nodes that are eliminated algebraically.

Eliminating an internal node does not discard its mechanical effect. Its
increment is solved in terms of the retained increments and the applied drive,
and the result is substituted into the retained equations. A path between two
retained nodes can therefore be replaced by a direct effective coupling after
the intermediate internal nodes have been eliminated. This is how a sparse
local operator can produce a dense reduced operator.

The retained/internal partition is used here only as a thought experiment to
reveal the required interaction topology. \model does not eliminate physical
nodes.

\paragraph{The Schur complement.}
Consider the generic partitioned system
\begin{equation}
  \begin{bmatrix}
    \mathbf{A} & \mathbf{B}\\
    \mathbf{E} & \mathbf{D}
  \end{bmatrix}
  \begin{bmatrix}
    \mathbf{x}\\
    \mathbf{y}
  \end{bmatrix}
  =
  \begin{bmatrix}
    \mathbf{f}\\
    \mathbf{g}
  \end{bmatrix},
  \label{eq:si:hub:generic-block-system}
\end{equation}
where $\mathbf{x}$ contains the retained unknowns and $\mathbf{y}$ contains
the internal unknowns. The internal block row gives
\begin{equation}
  \mathbf{y}
  =
  \mathbf{D}^{-1}
  \left(
    \mathbf{g}-\mathbf{E}\mathbf{x}
  \right),
  \label{eq:si:hub:generic-internal-state}
\end{equation}
provided that $\mathbf{D}$ is invertible. Substituting this result into the
retained block row gives
\begin{equation}
  \underbrace{
  \left(
    \mathbf{A}-\mathbf{B}\mathbf{D}^{-1}\mathbf{E}
  \right)\mathbf{x}
  }_{\text{condensed left-hand side}}
  =
  \underbrace{
  \mathbf{f}-\mathbf{B}\mathbf{D}^{-1}\mathbf{g}
  }_{\text{condensed right-hand side}}.
  \label{eq:si:hub:generic-condensation}
\end{equation}
The matrix
\begin{equation}
  \mathbf{A}-\mathbf{B}\mathbf{D}^{-1}\mathbf{E}
\end{equation}
is the Schur complement of the internal block $\mathbf{D}$. It is the exact
operator acting on the retained unknowns after the internal unknowns have
been eliminated. The same elimination transfers the internal drive
$\mathbf{g}$ to the retained right-hand side through
$-\mathbf{B}\mathbf{D}^{-1}\mathbf{g}$.

Thus, static condensation has two simultaneous effects: it modifies the
retained operator on the left-hand side and transfers the internal drive to
the retained right-hand side.

\paragraph{Application to the Newmark system.}
Partition the velocity increments and the known force drive in
Equation~\eqref{eq:si:hub:assembled} as
\begin{equation}
  \Delta\mathbf{v}
  =
  \begin{bmatrix}
    \Delta\mathbf{v}_R\\
    \Delta\mathbf{v}_I
  \end{bmatrix},
  \qquad
  \mathbf{r}
  =
  \begin{bmatrix}
    \mathbf{r}_R\\
    \mathbf{r}_I
  \end{bmatrix}.
  \label{eq:si:hub:partitioned-state}
\end{equation}
Under the same partition,
\begin{equation}
  \mathbf{M}
  =
  \begin{bmatrix}
    \mathbf{M}_R & \mathbf{0}\\
    \mathbf{0} & \mathbf{M}_I
  \end{bmatrix},
  \qquad
  \mathbf{C}_{\mathrm{glob}}
  =
  \begin{bmatrix}
    \mathbf{C}_{RR} & \mathbf{C}_{RI}\\
    \mathbf{C}_{IR} & \mathbf{C}_{II}
  \end{bmatrix}.
  \label{eq:si:hub:partitioned-operators}
\end{equation}
Here $\mathbf{C}_{RR}$ maps retained increments into the retained equations,
$\mathbf{C}_{RI}$ maps internal increments into the retained equations,
$\mathbf{C}_{IR}$ maps retained increments into the internal equations, and
$\mathbf{C}_{II}$ maps internal increments into the internal equations.
These are precisely the four blocks of $\mathbf{C}_{\mathrm{glob}}$ defined
in Equation~\eqref{eq:si:hub:coupling-operator}.

Using these blocks, Equation~\eqref{eq:si:hub:assembled} becomes
\begin{equation}
  \begin{bmatrix}
    \mathbf{M}_R+\mathbf{C}_{RR}
    & \mathbf{C}_{RI}\\
    \mathbf{C}_{IR}
    & \mathbf{M}_I+\mathbf{C}_{II}
  \end{bmatrix}
  \begin{bmatrix}
    \Delta\mathbf{v}_R\\
    \Delta\mathbf{v}_I
  \end{bmatrix}
  =
  \begin{bmatrix}
    \mathbf{r}_R\\
    \mathbf{r}_I
  \end{bmatrix}.
  \label{eq:si:hub:partitioned-newmark}
\end{equation}
The correspondence with
Equation~\eqref{eq:si:hub:generic-block-system} is
\begin{equation}
  \mathbf{A}
  =\mathbf{M}_R+\mathbf{C}_{RR},
  \qquad
  \mathbf{B}
  =\mathbf{C}_{RI},
  \qquad
  \mathbf{E}
  =\mathbf{C}_{IR},
  \qquad
  \mathbf{D}
  =\mathbf{M}_I+\mathbf{C}_{II}.
\end{equation}
The internal block row therefore gives
\begin{equation}
  \Delta\mathbf{v}_I
  =
  \left(
    \mathbf{M}_I+\mathbf{C}_{II}
  \right)^{-1}
  \left(
    \mathbf{r}_I
    -\mathbf{C}_{IR}\Delta\mathbf{v}_R
  \right).
  \label{eq:si:hub:eliminated-full-state}
\end{equation}
Substituting this result into the retained block row gives
\begin{equation}
  \underbrace{
  \left[
    \mathbf{M}_R+\mathbf{C}_{RR}
    -\mathbf{C}_{RI}
     \left(
       \mathbf{M}_I+\mathbf{C}_{II}
     \right)^{-1}
     \mathbf{C}_{IR}
  \right]\Delta\mathbf{v}_R
  }_{\text{condensed left-hand side}}
  =
  \underbrace{
    \mathbf{r}_R
    -\mathbf{C}_{RI}
     \left(
       \mathbf{M}_I+\mathbf{C}_{II}
     \right)^{-1}
     \mathbf{r}_I
  }_{\text{condensed right-hand side}}.
  \label{eq:si:hub:condensed-newmark}
\end{equation}
Equation~\eqref{eq:si:hub:condensed-newmark} is the exact condensed Newmark
equation. Its left- and right-hand-side consequences are considered
separately below.

\subsubsection{Left-hand side: Schur-complement coupling and the rank-one virtual-hub topology}
\label{si:hub:translation}

To isolate the topology produced specifically by the stiffness--damping
interaction, consider the static-condensation limit
\begin{equation}
  \mathbf{M}_I=\mathbf{0}.
\end{equation}
The matrix $\mathbf{M}_I$ is the mass block associated with the internal
degrees of freedom. Setting $\mathbf{M}_I=\mathbf{0}$ means that the internal
nodes are treated as massless and equilibrate without an inertial
contribution. This auxiliary limit is used only to expose the topology of
$\mathbf{C}_{\mathrm{glob}}$. It does not mean that physical nodes are
massless in \model.

For this static interaction problem, the blocks in
Equation~\eqref{eq:si:hub:generic-block-system} are
\begin{equation}
  \mathbf{A}=\mathbf{C}_{RR},
  \qquad
  \mathbf{B}=\mathbf{C}_{RI},
  \qquad
  \mathbf{E}=\mathbf{C}_{IR},
  \qquad
  \mathbf{D}=\mathbf{C}_{II}.
\end{equation}
The condensed interaction operator is therefore
\begin{equation}
  \mathbf{C}_{\mathrm{cond}}
  :=
  \mathbf{C}_{RR}
  -\mathbf{C}_{RI}\mathbf{C}_{II}^{-1}\mathbf{C}_{IR}.
  \label{eq:si:hub:schurdef}
\end{equation}
The Schur complement $\mathbf{C}_{\mathrm{cond}}$ is the exact interaction
operator acting on the retained increments after the internal increments have
been eliminated. It combines the direct retained-to-retained coupling
$\mathbf{C}_{RR}$ with the indirect coupling
\begin{equation}
  -\mathbf{C}_{RI}\mathbf{C}_{II}^{-1}\mathbf{C}_{IR}
\end{equation}
transmitted through the internal nodes. Because
$\mathbf{C}_{II}^{-1}$ is generally dense,
$\mathbf{C}_{\mathrm{cond}}$ can couple retained nodes that were not
neighbours in the original graph.

A uniform velocity increment,
$\Delta\mathbf{v}_i=\mathbf{a}$ at every node, produces no relative motion and
therefore no stiffness or damping contribution. Static condensation preserves
this property, so each block row of the condensed interaction operator
satisfies
\begin{equation}
  (\mathbf{C}_{\mathrm{cond}})_{ii}
  +\sum_{\substack{j\in R\\j\ne i}}
   (\mathbf{C}_{\mathrm{cond}})_{ij}
  =\mathbf{0},
  \qquad
  (\mathbf{C}_{\mathrm{cond}})_{ii}
  =
  -\sum_{\substack{j\in R\\j\ne i}}
   (\mathbf{C}_{\mathrm{cond}})_{ij}.
  \label{eq:si:hub:condensed-rowsum}
\end{equation}
Consequently,
\begin{equation}
  (\mathbf{C}_{\mathrm{cond}}\Delta\mathbf{v}_R)_i
  =
  \sum_{\substack{j\in R\\j\ne i}}
  \left[
    -(\mathbf{C}_{\mathrm{cond}})_{ij}
  \right]
  \left(
    \Delta\mathbf{v}_i-\Delta\mathbf{v}_j
  \right).
  \label{eq:si:hub:pairwise}
\end{equation}

The collection of pairwise coefficients
$-(\mathbf{C}_{\mathrm{cond}})_{ij}$ can be dense. In the isotropic scalar
case, these coefficients can be approximated using one separable factor per
retained node:
\begin{equation}
  -(\mathbf{C}_{\mathrm{cond}})_{ij}
  \approx
  w_iw_j\mathbf{I},
  \qquad
  w_i\ge0.
  \label{eq:si:hub:rank1}
\end{equation}
The scalar coefficient matrix with entries $w_iw_j$ is the rank-one outer
product $\mathbf{w}\mathbf{w}^{\top}$. Thus
Equation~\eqref{eq:si:hub:rank1} is a rank-one approximation of the dense
pairwise coupling coefficients. It is not an assumption that the exact Schur
complement $\mathbf{C}_{\mathrm{cond}}$ is rank one.

Substituting Equation~\eqref{eq:si:hub:rank1} into
Equation~\eqref{eq:si:hub:pairwise} gives
\begin{align}
  (\mathbf{C}_{\mathrm{cond}}\Delta\mathbf{v}_R)^{(1)}_i
  &=
  \sum_{j\ne i}
  w_iw_j
  \left(
    \Delta\mathbf{v}_i-\Delta\mathbf{v}_j
  \right)
  \nonumber\\
  &=
  w_i
  \left[
    \left(
      \sum_jw_j
    \right)\Delta\mathbf{v}_i
    -
    \sum_jw_j\Delta\mathbf{v}_j
  \right].
  \label{eq:si:hub:rank1-factor}
\end{align}
Define the shared weighted increment
\begin{equation}
  \overline{\Delta\mathbf{v}}
  :=
  \frac{
    \sum_jw_j\Delta\mathbf{v}_j
  }{
    \sum_jw_j
  }.
  \label{eq:si:hub:shared-increment}
\end{equation}
Equation~\eqref{eq:si:hub:rank1-factor} becomes
\begin{equation}
  (\mathbf{C}_{\mathrm{cond}}\Delta\mathbf{v}_R)^{(1)}_i
  =
  c_{iH}
  \left(
    \Delta\mathbf{v}_i
    -\overline{\Delta\mathbf{v}}
  \right),
  \qquad
  c_{iH}
  :=
  w_i\sum_jw_j.
  \label{eq:si:hub:star}
\end{equation}
This is the interaction form of a star. Each retained node couples only to the
same shared quantity $\overline{\Delta\mathbf{v}}$, represented by a virtual
hub. This establishes the rank-one virtual-hub topology on the left-hand side.

\subsubsection{Right-hand side: Transfer of the internal force drive}
\label{si:hub:rhs}

The exact condensed right-hand side in
Equation~\eqref{eq:si:hub:condensed-newmark} is
\begin{equation}
  \mathbf{r}_{\mathrm{cond}}
  :=
  \mathbf{r}_R
  -\mathbf{C}_{RI}
   \left(
     \mathbf{M}_I+\mathbf{C}_{II}
   \right)^{-1}
   \mathbf{r}_I.
  \label{eq:si:hub:condensed-rhs-full}
\end{equation}
In the static-condensation limit, this becomes
\begin{equation}
  \mathbf{r}_{\mathrm{cond}}
  =
  \mathbf{r}_R
  -\mathbf{C}_{RI}\mathbf{C}_{II}^{-1}\mathbf{r}_I.
  \label{eq:si:hub:condensed-rhs}
\end{equation}
The term $\mathbf{r}_R$ is the force drive already acting on the retained
degrees of freedom. The second term transfers the force drive from the
eliminated internal degrees of freedom into the retained equations. Because
$\mathbf{C}_{II}^{-1}$ is generally dense, a force acting at one internal node
can contribute to the effective right-hand side of many retained nodes.

The corresponding transfer can be written explicitly for the rank-one star.
Let $c_{iH}$ denote the coupling between retained node $i$ and the hub, and let
$\mathbf{r}_H$ denote the force drive associated with the condensed internal
degrees of freedom. The hub equation is
\begin{equation}
  \sum_i
  c_{iH}
  \left(
    \Delta\mathbf{v}_H-\Delta\mathbf{v}_i
  \right)
  =
  \mathbf{r}_H.
  \label{eq:si:hub:forced-hub-balance}
\end{equation}
Solving for the hub increment gives
\begin{equation}
  \Delta\mathbf{v}_H
  =
  \frac{
    \sum_i c_{iH}\Delta\mathbf{v}_i+\mathbf{r}_H
  }{
    \sum_i c_{iH}
  }.
  \label{eq:si:hub:forced-hub-state}
\end{equation}
Substitution into the retained-node equation gives
\begin{equation}
  c_{iH}
  \left[
    \Delta\mathbf{v}_i
    -
    \frac{
      \sum_jc_{jH}\Delta\mathbf{v}_j
    }{
      \sum_jc_{jH}
    }
  \right]
  =
  \mathbf{r}_i
  +
  \frac{
    c_{iH}
  }{
    \sum_jc_{jH}
  }
  \mathbf{r}_H.
  \label{eq:si:hub:forced-star}
\end{equation}
The left-hand side is the star interaction derived previously. The
right-hand side shows that the force drive associated with the condensed
internal degrees of freedom is distributed among the retained nodes according
to their hub couplings. Node $i$ receives the fraction
\begin{equation}
  \frac{
    c_{iH}
  }{
    \sum_jc_{jH}
  }
  \mathbf{r}_H.
\end{equation}
These fractions preserve the complete internal drive:
\begin{equation}
  \sum_i
  \frac{
    c_{iH}
  }{
    \sum_jc_{jH}
  }
  \mathbf{r}_H
  =
  \mathbf{r}_H.
  \label{eq:si:hub:force-transfer-sum}
\end{equation}

If the hub represents only internal redistribution and carries no external
drive, then $\mathbf{r}_H=\mathbf{0}$. In this case, the hub introduces no
additional net force or linear impulse into the retained system. This
zero-resultant condition does not by itself establish an energy balance,
which would additionally require a condition on the total power of the
transferred forces.

The left-hand-side derivation therefore establishes the effective star
operator, while the right-hand-side derivation shows how the same star
transfers a force drive associated with condensed internal degrees of freedom.

\subsection{Implementation of the Virtual Hub in \model: A Rank-One-Motivated Approximation of Schur-Complement-Induced Non-Local Coupling}
\label{si:hub:implementation}

The preceding derivation is a condensation thought experiment. \model does
not eliminate physical nodes, assemble $\mathbf{C}_{\mathrm{cond}}$, compute
$\mathbf{r}_{\mathrm{cond}}$ or recover the scalar factors $w_i$. Instead, it
implements the two conclusions of the derivation directly.

On the left-hand side, the scalar star interaction
\begin{equation}
  c_{iH}
  \left(
    \Delta\mathbf{v}_i-\Delta\mathbf{v}_H
  \right)
\end{equation}
motivates an operator-weighted virtual hub. \model replaces the scalar
coupling $c_{iH}$ with learned stiffness and damping operators and computes
the shared hub state from operator-weighted equilibrium.

On the right-hand side, the force-transfer result motivates learned force and
angular-flux messages carried by the virtual edges. These messages make the
hub an internal relay rather than an external actuator.

\subsubsection{Left-hand-side implementation: Operator-weighted hub coordinates}
\label{si:hub:state}

The stiffness and damping channels act on different mechanical quantities. At
the level of the linearised edge contribution,
\begin{equation}
  \delta\mathbf{f}^{K}_i
  =
  -\sum_{j\in\mathcal{N}(i)}
  \Kt_{ij}
  \left(
    \delta\mathbf{x}_i-\delta\mathbf{x}_j
  \right),
  \qquad
  \delta\mathbf{f}^{D}_i
  =
  -\sum_{j\in\mathcal{N}(i)}
  \Dt_{ij}
  \left(
    \delta\mathbf{v}_i-\delta\mathbf{v}_j
  \right).
  \label{eq:si:hub:mechanical-channels}
\end{equation}
Define the nodal operators
\begin{equation}
  \Kt_i
  =
  \sum_{j\in\mathcal{N}(i)}\Kt_{ij},
  \qquad
  \Dt_i
  =
  \sum_{j\in\mathcal{N}(i)}\Dt_{ij},
  \qquad
  \Dt_{\mathrm{rot},i}
  =
  \sum_{j\in\mathcal{N}(i)}
  \Dt_{\mathrm{rot},ij}.
  \label{eq:si:hub:nodal-operators}
\end{equation}

The virtual hub is massless and has no independently advanced state. Its
position, velocity and spin are obtained from equilibrium in the corresponding
operator channels:
\begin{align}
  \mathbf{x}_H
  &=
  \left(
    \sum_i\Kt_i
  \right)^{-1}
  \sum_i\Kt_i\mathbf{x}_i,
  \nonumber\\
  \mathbf{v}_H
  &=
  \left(
    \sum_i\Dt_i
  \right)^{-1}
  \sum_i\Dt_i\mathbf{v}_i,
  \nonumber\\
  \bm{\omega}_H
  &=
  \left(
    \sum_i\Dt_{\mathrm{rot},i}
  \right)^{-1}
  \sum_i
  \Dt_{\mathrm{rot},i}\bm{\omega}_i.
  \label{eq:si:hub:equilibrium}
\end{align}
These expressions are the operator-valued counterparts of the shared weighted
quantity derived for the rank-one star.

For example, if $\Kt_i=k_i\mathbf{I}$, then
\begin{equation}
  \mathbf{x}_H
  =
  \frac{
    \sum_i k_i\mathbf{x}_i
  }{
    \sum_i k_i
  }.
  \label{eq:si:hub:scalar-mean}
\end{equation}
Only when all $k_i$ are equal does this expression reduce to the arithmetic
centroid. The hub is therefore an operator-weighted mechanical state, not a
geometric centroid.

The Newmark factors multiplying the stiffness and damping channels cancel
from their corresponding equilibrium equations. The hub position, velocity
and spin are recomputed at every substep and carry no state between substeps.
The implementation symmetrises each summed operator and adds
$10^{-6}\mathbf{I}$ before the three $3\times3$ solves.

\subsubsection{Right-hand-side implementation: Balanced virtual-edge interactions}
\label{si:hub:recoil}

The right-hand-side derivation shows that the star acts as an internal relay:
it transfers a force drive associated with condensed internal degrees of
freedom to the retained degrees of freedom. The virtual hub in \model plays
the same relay role. It is not an external actuator and carries no prescribed
external load.

The model does not explicitly compute $\mathbf{r}_H$ or the exact
Schur-complement force transfer. Instead, it decodes force and angular-flux
contributions on the virtual edges. Because these quantities represent
internal transfer through an unloaded hub, they are projected to have zero
sum:
\begin{equation}
  \widetilde{\mathbf{f}}_{H\to j}
  =
  \mathbf{f}_{H\to j}
  -\frac{1}{n_H}
   \sum_k\mathbf{f}_{H\to k},
  \qquad
  \widetilde{\mathbf{A}}_{H\to j}
  =
  \mathbf{A}_{H\to j}
  -\frac{1}{n_H}
   \sum_k\mathbf{A}_{H\to k}.
  \label{eq:si:hub:zerosum}
\end{equation}
Therefore,
\begin{equation}
  \sum_j\widetilde{\mathbf{f}}_{H\to j}
  =
  \mathbf{0},
  \qquad
  \sum_j\widetilde{\mathbf{A}}_{H\to j}
  =
  \mathbf{0}.
  \label{eq:si:hub:zeroflux}
\end{equation}
The first identity ensures that the hub redistributes force among the physical
nodes without adding a net linear resultant. This is the model counterpart of
the unloaded theoretical hub, $\mathbf{r}_H=\mathbf{0}$. The second identity
imposes the corresponding zero-sum condition on the angular-flux coordinates.

Because different virtual edges can use different reference points, zero net
force does not necessarily imply zero net moment. The virtual-edge forces can
retain the residual couple
\begin{equation}
  \mathbf{T}_H
  =
  \sum_j
  \mathbf{r}^{0}_{Hj}
  \times
  \widetilde{\mathbf{f}}_{H\to j}.
  \label{eq:si:hub:couple}
\end{equation}
The corresponding linear-momentum, angular-momentum and energy properties must
therefore be stated separately.

\subsubsection{Computational cost}
\label{si:hub:cost}

The hub adds $2N$ directed virtual edges and three $3\times3$
operator-weighted solves per substep. Its additional cost is linear in the
number of physical nodes. The augmented graph has diameter two, while the
physical-edge message passing and independent nodal Newmark solves remain
unchanged.
\section{Virtual-Hub Coupling and the Semi-Implicit Nodal Newmark Update in \model}
\label{si:newmark}

The preceding section established the virtual hub as a tractable
representation of non-local operator coupling and internal force transfer.
We now describe how \model combines physical-edge and virtual-edge messages
with independent nodal Newmark solves to advance the physical state. The
connection to classical Newmark integration identifies the numerical template;
it does not make the learned update a classical global implicit solve.

\subsection{One learned mechanical substep}
\label{si:newmark:setting}

One observed transition of duration $\dt$ is divided into $S$ substeps of size
$\subdt=\dt/S$. At the beginning of each substep, the network decodes the net
force acting on physical node $i$,
\begin{equation}
  \mathbf{b}_i
  =
  \mathbf{f}^{\mathrm{ext}}_i
  +\sum_{j\in\mathcal{N}(i)}\mathbf{f}_{ij},
  \label{eq:si:eom}
\end{equation}
where $\mathcal{N}(i)$ includes both physical neighbours and the virtual hub.
Thus physical-edge and virtual-edge forces contribute directly to the
right-hand side of the nodal update.

Each edge also carries a position-response operator $\Kt_{ij}$ and a
velocity-response operator $\Dt_{ij}$. Their nodal sums are
\begin{equation}
  \Kt_i
  =
  \sum_{j\in\mathcal{N}(i)}\Kt_{ij},
  \qquad
  \Dt_i
  =
  \sum_{j\in\mathcal{N}(i)}\Dt_{ij}.
  \label{eq:si:tangents}
\end{equation}
The virtual-edge operators therefore contribute to the implicit mechanical
response on the left-hand side, while the virtual-edge forces contribute to
the drive on the right-hand side. This mirrors the operator and force-transfer
roles derived for the hub in Section~\ref{si:hub}.

The matrices $\Kt_i$ and $\Dt_i$ are learned, state-dependent response
coefficients. They are not obtained by differentiating the decoded forces and
are not assumed to be physical material tangents. The inverse mass is likewise
learned and parameterised as the positive isotropic operator
\begin{equation}
  \mathbf{M}_i^{-1}=m_i^{-1}\mathbf{I},
  \qquad
  m_i>0.
\end{equation}

Define the nodal increments
\begin{equation}
  \Delta\mathbf{v}_i
  :=
  \mathbf{v}_{i,t+1}-\mathbf{v}_{i,t},
  \qquad
  \Delta\mathbf{x}_i
  :=
  \mathbf{x}_{i,t+1}-\mathbf{x}_{i,t}.
\end{equation}
The decoded operators represent the change in mechanical response over the
substep as
\begin{equation}
  \Delta\mathbf{f}^{\mathrm{resp}}_i
  =
  -\Dt_i\Delta\mathbf{v}_i
  -\Kt_i\Delta\mathbf{x}_i.
  \label{eq:si:response-change}
\end{equation}
Equation~\eqref{eq:si:response-change} defines how the learned operators enter
the update. It does not identify them as Jacobians of $\mathbf{b}_i$.

The velocity increment is determined from the trapezoidal response balance
\begin{equation}
  \mathbf{M}_i\Delta\mathbf{v}_i
  =
  \mathbf{b}_i\subdt
  -\tfrac12\subdt\,\Dt_i\Delta\mathbf{v}_i
  -\tfrac12\subdt\,\Kt_i\Delta\mathbf{x}_i.
  \label{eq:si:response-balance}
\end{equation}
The position increment is evaluated from the mean velocity,
\begin{equation}
  \Delta\mathbf{x}_i
  =
  \tfrac12\subdt
  \left(
    \mathbf{v}_{i,t}+\mathbf{v}_{i,t+1}
  \right)
  =
  \subdt
  \left(
    \mathbf{v}_{i,t}
    +\tfrac12\Delta\mathbf{v}_i
  \right).
  \label{eq:si:dx}
\end{equation}
Substituting Equation~\eqref{eq:si:dx} into
Equation~\eqref{eq:si:response-balance}, collecting the terms in
$\Delta\mathbf{v}_i$, and multiplying by $\mathbf{M}_i^{-1}$ gives the update
implemented by \model:
\begin{equation}
  \boxed{
  \left[
    \mathbf{I}
    +\tfrac12\subdt\,\mathbf{M}_i^{-1}\Dt_i
    +\tfrac14\subdt^2\mathbf{M}_i^{-1}\Kt_i
  \right]\Delta\mathbf{v}_i
  =
  \mathbf{M}_i^{-1}\mathbf{b}_i\subdt
  -\tfrac12\subdt^2
   \mathbf{M}_i^{-1}\Kt_i\mathbf{v}_{i,t}
  }.
  \label{eq:si:solve}
\end{equation}
The state is then advanced by
\begin{equation}
  \mathbf{v}_{i,t+1}
  =
  \mathbf{v}_{i,t}+\Delta\mathbf{v}_i,
  \qquad
  \mathbf{x}_{i,t+1}
  =
  \mathbf{x}_{i,t}+\Delta\mathbf{x}_i.
  \label{eq:si:state-update}
\end{equation}

The physical and virtual messages make $\mathbf{b}_i$, $\Kt_i$ and $\Dt_i$
dependent on the current state of the complete graph. Once these quantities
have been decoded, Equation~\eqref{eq:si:solve} is solved independently at
each free physical node as a $3\times3$ system.

\subsection{Connection to average-acceleration Newmark integration and the meaning of semi-implicit}
\label{si:newmark:family}
\label{si:newmark:deviations}

For the average-acceleration Newmark parameters
$\gamma=\tfrac12$ and $\beta=\tfrac14$, the classical velocity and position
increments satisfy
\begin{equation}
  \Delta\mathbf{v}
  =
  \tfrac12\subdt
  \left(
    \mathbf{a}_t+\mathbf{a}_{t+1}
  \right),
  \qquad
  \Delta\mathbf{x}
  =
  \subdt\mathbf{v}_t
  +\tfrac12\subdt\Delta\mathbf{v}.
  \label{eq:si:newmark-average}
\end{equation}
The second relation is exactly the mean-velocity position rule in
Equation~\eqref{eq:si:dx}.

For the classical linear system
\begin{equation}
  \mathbf{M}\ddot{\mathbf{x}}
  +\mathbf{D}\dot{\mathbf{x}}
  +\mathbf{K}\mathbf{x}
  =
  \mathbf{f}_{\mathrm{ext}},
  \label{eq:si:linear-system}
\end{equation}
the internal-force change over one step is
\begin{equation}
  -\mathbf{D}\Delta\mathbf{v}
  -\mathbf{K}\Delta\mathbf{x}.
\end{equation}
The average-acceleration rule therefore produces the coefficients
$\subdt/2$ and $\subdt^2/4$ that appear in
Equation~\eqref{eq:si:solve}. This correspondence explains the numerical form
of the learned update. It does not identify the learned operators $\Kt_i$ and
$\Dt_i$ with the physical matrices $\mathbf{K}$ and $\mathbf{D}$.

The learned update is implicit in each nodal increment because
$\Delta\mathbf{v}_i$ and the associated position increment enter the response
being solved. It is nevertheless not a classical global implicit Newmark
step. In \model,

\begin{enumerate}
  \item the response operators are decoded from the current graph state rather
  than obtained by differentiating a prescribed mechanical residual;
  \item one learned solve is performed per substep, without an inner Newton
  iteration; and
  \item the global $3N\times3N$ solve is replaced by independent nodal solves,
  while message passing and the operator-weighted hub supply non-local
  information.
\end{enumerate}

The term \emph{semi-implicit} refers to this separation: each nodal response
is treated implicitly, but the complete off-diagonal mechanical system is not
assembled or solved exactly.

\subsection{Response-operator construction and nodal solvability}
\label{si:newmark:spd}
\label{si:newmark:wellposed}

Each edge decoder emits six scalars and forms the lower-triangular matrix
\begin{equation}
  \mathbf{L}
  =
  \begin{bmatrix}
    d_0&0&0\\
    s_1&d_1&0\\
    s_3&s_4&d_2
  \end{bmatrix},
  \qquad
  (d_0,d_1,d_2)
  =
  \operatorname{softplus}(s_0,s_2,s_5)+10^{-4}.
  \label{eq:si:cholesky}
\end{equation}
If $\mathbf{R}_{ij}$ is the edge-local orthonormal frame, the corresponding
operator in global coordinates is
\begin{equation}
  \mathbf{W}_{ij}
  =
  \mathbf{R}_{ij}
  \mathbf{L}\mathbf{L}^{\top}
  \mathbf{R}_{ij}^{\top}.
  \label{eq:si:global-operator}
\end{equation}
The positive diagonal of $\mathbf{L}$ makes
$\mathbf{L}\mathbf{L}^{\top}$ positive definite. Under a rotation
$\mathbf{Q}$,
\begin{equation}
  \mathbf{R}_{ij}
  \mapsto
  \mathbf{Q}\mathbf{R}_{ij},
  \qquad
  \mathbf{W}_{ij}
  \mapsto
  \mathbf{Q}\mathbf{W}_{ij}\mathbf{Q}^{\top}.
\end{equation}
The edge operators, their nodal sums and the solution of
Equation~\eqref{eq:si:solve} are therefore rotation-equivariant.

Positive definiteness alone does not imply symmetry between the two directed
traversals of an edge. The equality
$\mathbf{W}_{ji}=\mathbf{W}_{ij}$ follows from the interaction construction:
the decoder coefficients are shared between the two traversals, and reversal
of the local frame leaves the quadratic operator unchanged.

For the isotropic inverse mass
$\mathbf{M}_i^{-1}=m_i^{-1}\mathbf{I}$, the coefficient matrix in the nodal
solve is
\begin{equation}
  \mathbf{A}_i
  =
  \mathbf{I}
  +\tfrac12\subdt\,m_i^{-1}\Dt_i
  +\tfrac14\subdt^2m_i^{-1}\Kt_i.
  \label{eq:si:nodal-coefficient}
\end{equation}
Because $\Kt_i$ and $\Dt_i$ are symmetric positive semidefinite,
$\mathbf{A}_i$ is symmetric positive definite for every $\subdt>0$, with
\begin{equation}
  \lambda_{\min}(\mathbf{A}_i)\ge1,
  \qquad
  \left\lVert\mathbf{A}_i^{-1}\right\rVert_2\le1.
  \label{eq:si:nodal-solvability}
\end{equation}
Thus every nodal $3\times3$ solve remains invertible as the step size or
operator magnitude increases. This guarantees solvability of an individual
nodal system, not stability of the complete learned rollout. The
implementation adds $10^{-8}\mathbf{I}$ as a floating-point safeguard.

The update is invariant to a uniform translation of the positions because
the graph construction uses relative coordinates. It is not invariant to a
Galilean velocity shift: the term
\begin{equation}
  -\tfrac12\subdt^2
  \mathbf{M}_i^{-1}\Kt_i\mathbf{v}_{i,t}
\end{equation}
changes under
$\mathbf{v}_{i,t}\mapsto\mathbf{v}_{i,t}+\mathbf{c}$.

\subsection{Classical stability correspondence and its scope}
\label{si:newmark:properties}
\label{si:newmark:scope}

The stability statement concerns only the classical linear test problem. Set
$m=1$, $D=0$ and $K=\omega^2$ in
Equation~\eqref{eq:si:linear-system}, and set the learned coefficients in
Equation~\eqref{eq:si:solve} equal to these exact linear coefficients. Under
this controlled substitution, the nodal update reduces to the classical
average-acceleration Newmark method.

Let
\begin{equation}
  \Omega=\subdt\omega,
  \qquad
  a=\frac{4\Omega^2}{4+\Omega^2}.
\end{equation}
The amplification matrix carrying
$(x_t,\subdt v_t)$ to $(x_{t+1},\subdt v_{t+1})$ is
\begin{equation}
  \mathbf{G}(\Omega)
  =
  \begin{bmatrix}
    1-a/2&1-a/4\\
    -a&1-a/2
  \end{bmatrix}.
  \label{eq:si:amplification}
\end{equation}
For every finite $\Omega>0$,
\begin{equation}
  \det\mathbf{G}(\Omega)=1,
  \qquad
  \rho\!\left(\mathbf{G}(\Omega)\right)=1.
  \label{eq:si:linear-stability}
\end{equation}
The classical undamped update is therefore unconditionally stable and
introduces no algorithmic decay. It remains subject to phase error:
\begin{equation}
  \theta
  =
  \Omega
  \left(
    1-\tfrac1{12}\Omega^2+\mathcal{O}(\Omega^4)
  \right),
  \qquad
  \frac{T_{\mathrm{num}}}{T}
  =
  1+\tfrac1{12}\Omega^2+\mathcal{O}(\Omega^4).
  \label{eq:si:period}
\end{equation}
Thus a large step can remain bounded while accumulating phase error. The
classical method is second-order accurate for this linear problem.

These standard properties do not establish unconditional stability of
\model. In the trained model, the forces and response operators depend on the
predicted state, the operators are not force derivatives, and the global
off-diagonal system is replaced by local solves, message passing and the
virtual hub. Stability of the complete rollout is therefore evaluated
empirically through the beam rollouts, the explicit ablation and the
stiffness-dependent comparisons in Section~\ref{si:beam:stiffness}.

\subsection{Angular update}
\label{si:newmark:angular}

The angular state uses the same algebraic update:
\begin{equation}
  \left[
    \mathbf{I}
    +\tfrac12\subdt\,
     \mathbf{I}_i^{-1}\Dt_{\mathrm{rot},i}
    +\tfrac14\subdt^2
     \mathbf{I}_i^{-1}\Kt_{\mathrm{rot},i}
  \right]\Delta\bm{\omega}_i
  =
  \mathbf{I}_i^{-1}\bm{\tau}_i\subdt
  -\tfrac12\subdt^2
   \mathbf{I}_i^{-1}
   \Kt_{\mathrm{rot},i}\bm{\omega}_{i,t}.
  \label{eq:si:solve_rot}
\end{equation}
Here $\mathbf{I}_i^{-1}$ is a learned positive isotropic inverse inertia,
$\bm{\tau}_i$ is the aggregated decoded torque, and
$\Kt_{\mathrm{rot},i}$ and $\Dt_{\mathrm{rot},i}$ are learned rotational
response operators. The angular state is latent and receives no direct
supervision.

\subsection{External forcing and constrained nodes}
\label{si:newmark:substep}

For the beam experiments, the applied load is observed. Its known values
across the outer interval are interpolated over the $S$ substeps and held
constant within each substep. For motion capture, protein dynamics and walking
biomechanics, external forcing is unobserved and decoded from the current
predicted state at the beginning of each substep. The decoded value is held
constant during the corresponding nodal solve, and no future forcing is
supplied.

Equation~\eqref{eq:si:solve} is applied only to free physical nodes. Clamped
degrees of freedom are prescribed. The virtual hub is not integrated as a
physical degree of freedom; its position, velocity and spin are recomputed
from the current physical state and learned operators at every substep.

\subsection{Explicit limit}
\label{si:newmark:limit}

As
\begin{equation}
  \Kt_i,\Dt_i\to\mathbf{0},
\end{equation}
Equation~\eqref{eq:si:solve} reduces to
\begin{equation}
  \Delta\mathbf{v}_i
  =
  \mathbf{M}_i^{-1}\mathbf{b}_i\subdt,
  \qquad
  \Delta\mathbf{x}_i
  =
  \tfrac12\subdt
  \left(
    \mathbf{v}_{i,t}+\mathbf{v}_{i,t+1}
  \right).
  \label{eq:si:explicit_limit}
\end{equation}
This limit integrates the decoded force explicitly while retaining the
trapezoidal position rule. Its momentum properties depend on whether the
contributing fluxes arise from antisymmetric physical-edge exchanges or from
the projected virtual-hub interactions. Section~\ref{si:conservation} treats
these cases separately.
\section{Momentum Balance of the Decoded Interactions and the Residual of the Independent Nodal Update}
\label{si:conservation}

The decoded interactions and the numerical update have distinct conservation
properties. Physical-edge interactions exchange linear and angular momentum
pairwise, while the virtual hub enforces collective balance over its incident
virtual edges. These properties hold before time integration.

The independent semi-implicit nodal update preserves this interaction-level
balance only up to a finite-substep linear-momentum residual. The residual is
not an additional force produced by the decoder or the virtual hub. It arises
when the balanced decoded drive is processed by the learned response operators
independently at each physical node. Under the boundedness assumptions stated
below, the absolute residual over one substep is
$\mathcal{O}(\subdt^2)$.

The interaction-level analysis concerns linear and angular momentum. The
numerical-update analysis concerns total linear momentum. No
energy-conservation claim is made, because zero net force does not by itself
imply zero mechanical power.

For the dynamically integrated physical nodes
$\mathcal{V}_{\mathrm{dyn}}$, define the linear-momentum increment
\begin{equation}
  \Delta\mathbf{P}
  :=
  \mathbf{P}_{t+1}-\mathbf{P}_t
  =
  \sum_{i\in\mathcal{V}_{\mathrm{dyn}}}
  \mathbf{M}_i\Delta\mathbf{v}_i.
  \label{eq:si:total-momentum}
\end{equation}
The virtual hub is massless and is not included in this sum. The total
external force acting on the integrated nodes is
\begin{equation}
  \mathbf{F}_{\mathrm{ext}}
  :=
  \sum_{i\in\mathcal{V}_{\mathrm{dyn}}}
  \mathbf{f}^{\mathrm{ext}}_i.
  \label{eq:si:total-external-force}
\end{equation}
Reactions associated with prescribed degrees of freedom must be included in
the external impulse.

\subsection{Balance of the Decoded Momentum Exchanges}

\subsubsection{Pairwise Balance of Physical-Edge Interactions}
\label{si:conservation:fluxes}

Consider a physical edge $(i,j)$. Its two directed traversals satisfy
\begin{equation}
  \mathbf{f}_{ij}
  =
  -\mathbf{f}_{ji},
  \qquad
  \mathbf{A}_{ij}
  =
  -\mathbf{A}_{ji},
  \qquad
  \mathbf{r}^{0}_{ij}
  =
  \mathbf{r}^{0}_{ji}.
  \label{eq:si:physical-antisymmetry}
\end{equation}
Here $\mathbf{f}_{ij}$ is the force transmitted from $j$ to $i$,
$\mathbf{A}_{ij}$ is the decoded angular-momentum flux, and
$\mathbf{r}^{0}_{ij}$ is the application point shared by the two traversals.
The spin torque delivered to node $i$ is
\begin{equation}
  \bm{\tau}_{ij}
  =
  \mathbf{A}_{ij}
  -
  \left(
    \mathbf{r}_i-\mathbf{r}^{0}_{ij}
  \right)
  \times\mathbf{f}_{ij}.
  \label{eq:si:physical-spin-torque}
\end{equation}

The two forces have zero resultant:
\begin{equation}
  \mathbf{f}_{ij}
  +
  \mathbf{f}_{ji}
  =
  \mathbf{0}.
  \label{eq:si:physical-linear-balance}
\end{equation}

Angular momentum contains an orbital contribution from the force and an
intrinsic contribution from the spin torque. About an arbitrary origin
$\mathbf{o}$, the combined contribution of the two traversals is
\begin{align}
  &
  \left(
    \mathbf{r}_i-\mathbf{o}
  \right)\times\mathbf{f}_{ij}
  +\bm{\tau}_{ij}
  +
  \left(
    \mathbf{r}_j-\mathbf{o}
  \right)\times\mathbf{f}_{ji}
  +\bm{\tau}_{ji}
  \nonumber\\
  &\qquad=
  \left(
    \mathbf{r}^{0}_{ij}-\mathbf{o}
  \right)
  \times
  \left(
    \mathbf{f}_{ij}+\mathbf{f}_{ji}
  \right)
  +
  \mathbf{A}_{ij}+\mathbf{A}_{ji}
  =
  \mathbf{0}.
  \label{eq:si:physical-angular-balance}
\end{align}
Each physical edge therefore exchanges linear and angular momentum without
creating either.

\subsubsection{Collective Balance of Virtual-Hub Interactions}
\label{si:conservation:hub}

The virtual hub is a massless internal relay. It redistributes interactions
among the physical nodes, but it is not an independently integrated
mechanical body and carries no prescribed external load.

Physical-edge balance is imposed pairwise. Virtual-hub balance is imposed
collectively over all hub-to-body edges. The projection derived in
Section~\ref{si:hub:recoil} gives
\begin{equation}
  \sum_j
  \widetilde{\mathbf{f}}_{H\to j}
  =
  \mathbf{0},
  \qquad
  \sum_j
  \widetilde{\mathbf{A}}_{H\to j}
  =
  \mathbf{0}.
  \label{eq:si:conservation:hub-zero-sum}
\end{equation}
Thus the hub adds no net linear resultant, and the decoded
angular-momentum-flux channel is also collectively balanced.

On each virtual edge, the force and spin torque are referred to the same
learned application point:
\begin{equation}
  \widetilde{\bm{\tau}}_{H\to j}
  =
  \widetilde{\mathbf{A}}_{H\to j}
  -
  \left(
    \mathbf{r}_j-\mathbf{r}^{0}_{Hj}
  \right)
  \times
  \widetilde{\mathbf{f}}_{H\to j}.
  \label{eq:si:hub-spin-torque}
\end{equation}
The total angular contribution delivered to the physical nodes about an
arbitrary origin $\mathbf{o}$ is therefore
\begin{align}
  \mathcal{M}_H(\mathbf{o})
  &:=
  \sum_j
  \left[
    \left(
      \mathbf{r}_j-\mathbf{o}
    \right)
    \times
    \widetilde{\mathbf{f}}_{H\to j}
    +
    \widetilde{\bm{\tau}}_{H\to j}
  \right]
  \nonumber\\
  &=
  \sum_j
  \left[
    \left(
      \mathbf{r}^{0}_{Hj}-\mathbf{o}
    \right)
    \times
    \widetilde{\mathbf{f}}_{H\to j}
    +
    \widetilde{\mathbf{A}}_{H\to j}
  \right].
  \label{eq:si:hub-total-moment}
\end{align}
Using Equation~\eqref{eq:si:conservation:hub-zero-sum}, this can be expressed
relative to any common hub point $\mathbf{r}_H$ as
\begin{equation}
  \mathcal{M}_H(\mathbf{o})
  =
  \sum_j
  \left(
    \mathbf{r}^{0}_{Hj}-\mathbf{r}_H
  \right)
  \times
  \widetilde{\mathbf{f}}_{H\to j}.
  \label{eq:si:hub-reference-point-term}
\end{equation}
The arbitrary origin has disappeared because the projected forces have zero
resultant.

Equation~\eqref{eq:si:hub-reference-point-term} vanishes when all virtual-edge
application points coincide at $\mathbf{r}_H$, or more generally when their
force-weighted moment about $\mathbf{r}_H$ is zero. The virtual hub therefore
enforces exact collective balance of the decoded force and
angular-momentum-flux channels. Exact total angular-momentum balance about a
common spatial origin additionally depends on the learned application
points. Their locations relative to the operator-weighted hub have not been
measured, so this geometric contribution is not quantified here.

\subsection{Linear-Momentum Residual of the Independent Nodal Update}
\label{si:conservation:violation}

For clarity, we reproduce the quantities entering the nodal update from
Section~\ref{si:newmark:setting}. At physical node $i$, the decoded drive and
the nodal response operators are
\begin{equation}
  \mathbf{b}_i
  =
  \mathbf{f}^{\mathrm{ext}}_i
  +
  \sum_{j\in\mathcal{N}(i)}
  \mathbf{f}_{ij},
  \qquad
  \Kt_i
  =
  \sum_{j\in\mathcal{N}(i)}
  \Kt_{ij},
  \qquad
  \Dt_i
  =
  \sum_{j\in\mathcal{N}(i)}
  \Dt_{ij}.
  \label{eq:si:conservation:nodal-quantities}
\end{equation}
The physical-edge antisymmetry and virtual-hub projection established above
give
\begin{equation}
  \sum_{i\in\mathcal{V}_{\mathrm{dyn}}}
  \mathbf{b}_i
  =
  \mathbf{F}_{\mathrm{ext}}.
  \label{eq:si:balanced-drive}
\end{equation}

For the isotropic mass
$\mathbf{M}_i=m_i\mathbf{I}$, the coefficient matrix is
\begin{equation}
  \mathbf{A}_i
  :=
  \mathbf{I}
  +\tfrac12\subdt\,m_i^{-1}\Dt_i
  +\tfrac14\subdt^2m_i^{-1}\Kt_i.
  \label{eq:si:conservation:nodal-coefficient}
\end{equation}
The independent nodal update from
Equation~\eqref{eq:si:solve} is reproduced here as
\begin{equation}
  \mathbf{A}_i\Delta\mathbf{v}_i
  =
  m_i^{-1}\mathbf{b}_i\subdt
  -
  \tfrac12\subdt^2
  m_i^{-1}\Kt_i\mathbf{v}_{i,t}.
  \label{eq:si:conservation:nodal-solve}
\end{equation}
Multiplying by $m_i$ and solving for the momentum increment gives
\begin{equation}
  \Delta\mathbf{p}_i
  =
  \mathbf{A}_i^{-1}
  \left[
    \mathbf{b}_i\subdt
    -
    \tfrac12\subdt^2
    \Kt_i\mathbf{v}_{i,t}
  \right].
  \label{eq:si:dp}
\end{equation}
Equation~\eqref{eq:si:dp} contains two response-dependent contributions to
the total momentum increment.

\paragraph{Response of a balanced force pair.}

Consider an antisymmetric physical-edge force
$\mathbf{f}_{ji}=-\mathbf{f}_{ij}$. Its direct contribution to the total
momentum increment is
\begin{equation}
  \Delta\mathbf{p}_{\mathrm{pair}}(ij)
  =
  \left(
    \mathbf{A}_i^{-1}
    -
    \mathbf{A}_j^{-1}
  \right)
  \mathbf{f}_{ij}\subdt.
  \label{eq:si:mech1}
\end{equation}
The decoded forces remain equal and opposite, but their momentum increments
need not cancel when the two nodal coefficient matrices differ. These
matrices depend on the learned masses and on the nodal position- and
velocity-response operators.

The same reasoning applies to the collectively balanced virtual-hub forces.
Their sum is zero before integration, but multiplication by different
$\mathbf{A}_i^{-1}$ matrices need not preserve that cancellation. The
virtual-hub projection remains balanced; the finite-substep residual is
introduced afterwards by the independent nodal response.

\paragraph{Known-velocity position response.}

Equation~\eqref{eq:si:dp} also contains the total contribution
\begin{equation}
  \Delta\mathbf{P}^{\,\mathrm{pos}}
  =
  -
  \tfrac12\subdt^2
  \sum_i
  \mathbf{A}_i^{-1}
  \Kt_i\mathbf{v}_{i,t}.
  \label{eq:si:known-velocity-residual}
\end{equation}
This term is the known-velocity position response on the right-hand side of
Equation~\eqref{eq:si:conservation:nodal-solve}. Because it is evaluated
independently at each physical node, its global sum is not constrained to
vanish.

Summing Equation~\eqref{eq:si:dp}, using
Equation~\eqref{eq:si:balanced-drive}, and subtracting the external impulse
gives the complete one-substep residual:
\begin{equation}
  \boxed{
  \mathbf{R}_{P}
  :=
  \Delta\mathbf{P}
  -
  \subdt\mathbf{F}_{\mathrm{ext}}
  =
  \subdt
  \sum_i
  \left(
    \mathbf{A}_i^{-1}-\mathbf{I}
  \right)
  \mathbf{b}_i
  -
  \tfrac12\subdt^2
  \sum_i
  \mathbf{A}_i^{-1}
  \Kt_i\mathbf{v}_{i,t}
  }.
  \label{eq:si:residual}
\end{equation}
The first term appears because each component of the balanced drive is acted
on by a different nodal matrix $\mathbf{A}_i^{-1}$. The second is the
known-velocity position-response contribution. Both terms are specific to the
independent nodal solves.

To show why the same residual is absent from a fully assembled Newmark solve,
we reproduce the corresponding global update:
\begin{equation}
  \left[
    \mathbf{M}
    +\tfrac12\subdt\,\mathbf{D}_{\mathrm{glob}}
    +\tfrac14\subdt^2\mathbf{K}_{\mathrm{glob}}
  \right]\Delta\mathbf{v}
  =
  \mathbf{b}\subdt
  -
  \tfrac12\subdt^2
  \mathbf{K}_{\mathrm{glob}}\mathbf{v}_t.
  \label{eq:si:conservation:global-newmark}
\end{equation}
Let
\begin{equation}
  \mathbf{e}^{\top}
  :=
  \begin{bmatrix}
    \mathbf{I}&\cdots&\mathbf{I}
  \end{bmatrix}
\end{equation}
denote the operator that sums the nodal blocks of a stacked vector. The
assembled stiffness and damping operators contain both nodal and cross-node
response blocks. Their internal block sums vanish:
\begin{equation}
  \mathbf{e}^{\top}\mathbf{K}_{\mathrm{glob}}
  =
  \mathbf{0},
  \qquad
  \mathbf{e}^{\top}\mathbf{D}_{\mathrm{glob}}
  =
  \mathbf{0}.
  \label{eq:si:conservation:global-zero-sum}
\end{equation}
Left-multiplying
Equation~\eqref{eq:si:conservation:global-newmark} by
$\mathbf{e}^{\top}$ therefore eliminates the stiffness and damping terms on
both sides:
\begin{equation}
  \mathbf{e}^{\top}\mathbf{M}\Delta\mathbf{v}
  =
  \subdt\,\mathbf{e}^{\top}\mathbf{b}.
\end{equation}
Using
\begin{equation}
  \mathbf{e}^{\top}\mathbf{M}\Delta\mathbf{v}
  =
  \Delta\mathbf{P},
  \qquad
  \mathbf{e}^{\top}\mathbf{b}
  =
  \mathbf{F}_{\mathrm{ext}},
\end{equation}
gives
\begin{equation}
  \Delta\mathbf{P}
  =
  \subdt\mathbf{F}_{\mathrm{ext}},
  \qquad
  \mathbf{R}_{P}
  =
  \mathbf{0}.
  \label{eq:si:conservation:global-balance}
\end{equation}

The distinction is structural. In the assembled system, the internal
stiffness and damping contributions cancel when the coupled equations are
summed, before the global system is inverted. In \model, the global system is
replaced by independent nodal solves, so the balanced drives are acted on
separately by the matrices $\mathbf{A}_i^{-1}$ and the cross-node cancellation
is not imposed. The virtual hub and message passing provide a tractable
approximation of the resulting non-local coupling, while
Equation~\eqref{eq:si:residual} quantifies the remaining one-substep
linear-momentum residual.

\subsection{Asymptotic Order of the Residual and Conservative Explicit-Response Limit}

For bounded masses and response operators,
Equation~\eqref{eq:si:conservation:nodal-coefficient} gives
\begin{equation}
  \mathbf{A}_i^{-1}
  =
  \mathbf{I}
  +
  \mathcal{O}(\subdt).
  \label{eq:si:coefficient-order}
\end{equation}
The first term in Equation~\eqref{eq:si:residual} is therefore
$\mathcal{O}(\subdt^2)$. The second already contains the factor
$\subdt^2$ and has the same order for bounded states and response operators.
Consequently,
\begin{equation}
  \mathbf{R}_{P}
  =
  \mathcal{O}(\subdt^2).
  \label{eq:si:residual-order}
\end{equation}
The absolute one-substep residual is second order in $\subdt$. Relative to a
non-zero impulse of order $\subdt$, it is first order in $\subdt$. This is a
local asymptotic result, not a bound on accumulated conservation drift.

The explicit-response limit is reproduced here for completeness. As
$\Kt_i,\Dt_i\to\mathbf{0}$,
\begin{equation}
  \mathbf{A}_i
  \longrightarrow
  \mathbf{I},
  \qquad
  \Delta\mathbf{p}_i
  \longrightarrow
  \mathbf{b}_i\subdt.
  \label{eq:si:conservation:explicit-limit}
\end{equation}
Using Equation~\eqref{eq:si:balanced-drive} then gives
\begin{equation}
  \Delta\mathbf{P}
  \longrightarrow
  \subdt\mathbf{F}_{\mathrm{ext}},
  \qquad
  \mathbf{R}_{P}
  \longrightarrow
  \mathbf{0}.
  \label{eq:si:conservation:explicit-balance}
\end{equation}
Thus balanced decoded forces and heterogeneous masses do not themselves
produce the residual. It enters through the learned semi-implicit response
terms.

\subsection{Numerical Verification}
\label{si:conservation:numerics}

We verify the two contributions in
Equation~\eqref{eq:si:residual} using a $12$-node system with random
positive-definite position- and velocity-response operators, non-uniform
masses, exactly antisymmetric physical-edge forces, and no external force.
The relative residual is
\begin{equation}
  R(\subdt)
  :=
  \frac{
    \left\lVert
      \sum_i m_i\Delta\mathbf{v}_i
    \right\rVert
  }{
    \sum_i
    \left\lVert
      m_i\Delta\mathbf{v}_i
    \right\rVert
  }.
  \label{eq:si:relative-residual}
\end{equation}

Two controlled modifications isolate the two terms in
Equation~\eqref{eq:si:residual}. Replacing the node-dependent
$\mathbf{A}_i^{-1}$ matrices by one common matrix removes the first term
because $\sum_i\mathbf{b}_i=\mathbf{0}$ in this test. Suppressing
$-\tfrac12\subdt^2\Kt_i\mathbf{v}_{i,t}$ removes the second term. These are
diagnostic checks, not alternative implementations of \model.

The observed order between the two reported substep sizes is
\begin{equation}
  p
  :=
  \frac{
    \log
    \left[
      R(0.2)/R(0.0125)
    \right]
  }{
    \log(16)
  }.
  \label{eq:si:observed-residual-order}
\end{equation}
A value near one corresponds to a first-order relative residual and,
equivalently, a second-order absolute residual.

\begin{table}[htbp]
  \centering
  \caption{\textbf{Linear-momentum residual of the independent nodal solve.}
  A common coefficient matrix removes the first term in
  Equation~\eqref{eq:si:residual}. Suppressing the known-velocity
  position-response term removes the second.}
  \label{tab:si:conservation}
  \small
  \begin{tabular}{llccc}
    \toprule
    & Configuration
    & $\subdt=0.2$
    & $\subdt=0.0125$
    & Observed $p$ \\
    \midrule
    A
    & Original nodal update
    & $4.24\times10^{-2}$
    & $4.10\times10^{-3}$
    & $0.84$ \\
    B
    & Common coefficient matrix
    & $3.23\times10^{-2}$
    & $2.21\times10^{-3}$
    & $0.97$ \\
    D
    & Known-velocity response term suppressed
    & $2.83\times10^{-2}$
    & $2.97\times10^{-3}$
    & $0.81$ \\
    C
    & Both controlled modifications
    & $1.70\times10^{-8}$
    & $1.67\times10^{-8}$
    & $\mathrm{n/a}$ \\
    \bottomrule
  \end{tabular}
\end{table}

The original nodal update in row~A exhibits the predicted approximately
first-order relative residual. Retaining either contribution alone,
rows~B and~D, gives the same asymptotic order. Removing both contributions,
row~C, reduces the residual to single-precision round-off. The pairwise
expression in Equation~\eqref{eq:si:mech1} agrees component-wise with the
measured contribution to $2.4\times10^{-7}$.

These checks verify the derived algebra for prescribed response operators.
They do not measure accumulated conservation drift in a trained rollout.

\subsection{Scope of the Conservation Statements}
\label{si:conservation:scope}

The physical-edge decoder conserves linear and angular momentum exactly under
the antisymmetry and shared-application-point conditions in
Equation~\eqref{eq:si:physical-antisymmetry}. The virtual hub enforces exact
collective balance of its decoded force and angular-momentum-flux channels.
Its total moment about a common spatial origin additionally depends on the
learned application points through
Equation~\eqref{eq:si:hub-reference-point-term}; this geometric contribution
has not been measured.

The decoded interactions therefore retain the intended conservation
structure before integration. The independent semi-implicit nodal update
introduces the linear-momentum residual in
Equation~\eqref{eq:si:residual}. Under the stated boundedness assumptions,
the residual is $\mathcal{O}(\subdt^2)$ over one substep and vanishes in the
explicit-response limit.

The numerical derivation concerns total linear momentum. It does not
establish a corresponding finite-substep result for the complete angular
update. Conservation drift has also not been measured in the trained
rollouts. The walking systems exchange momentum with the ground, the beam
with its clamp, and the protein with its represented environment. Quantifying
physical drift in these systems would require a complete accounting of
external impulses and boundary reactions.

\end{document}